\documentclass[lettersize,journal]{IEEEtran}
\usepackage{xcolor,soul,framed} 
\colorlet{shadecolor}{yellow}
\usepackage[pdftex]{graphicx}
\graphicspath{{../pdf/}{../jpeg/}}
\DeclareGraphicsExtensions{.pdf,.jpeg,.png}
\usepackage{soul}
\usepackage{makecell}
\usepackage{multirow}
\usepackage{multicol}
\usepackage{enumitem}
\usepackage{rotating}
\usepackage{array}
\usepackage{psfrag}
\usepackage{subfigure}
\usepackage{supertabular}
\usepackage[graphicx]{realboxes}
\usepackage{adjustbox}
\usepackage{threeparttable}
\usepackage{tikz}
\usepackage[top=0.7in,bottom=0.4in,left=0.4in,textwidth=7.7in]{geometry}
\usepackage{cite}
\usepackage{amsmath}
\usepackage{amsfonts}
\usepackage{bm}
\usepackage{booktabs}
\usepackage{lscape}
\usepackage{verbatim}
\usepackage{longtable}
\usepackage{stfloats}
\usepackage[misc]{ifsym}

\newcommand{\ie}{\emph{i.e., }}

\newcommand{\etc}{\emph{etc. }}
\usepackage[pagebackref=false,breaklinks=true,linkcolor=red,anchorcolor=blue, urlcolor=black, citecolor=green,letterpaper=true,colorlinks,bookmarks=false]{hyperref}
\usepackage{url}

\usepackage{tabularx}
\usepackage{multirow} 
\usepackage{bbding}
\usepackage{mathtools}
\usepackage{pifont}
\usepackage{bbm}
\usepackage{tablefootnote}

\newcommand{\C}{{\bf C}}
\newcommand{\Pbf}{{\bf P}}

\newcommand{\Z}{{\bf Z}}

\newcommand{\Ybf}{{\bf Y}}

\newcommand{\Bbf}{{\bf B}}

\newcommand{\X}{{\bf X}}
\newcommand{\Xbf}{{\bf X}}
\newcommand{\W}{{\bf W}}
\newcommand{\wbm}{{\bm w}}

\newcommand{\ybm}{{\bm y}}
\newcommand{\vbm}{{\bm v}}
\newcommand{\zbm}{{\bm z}}
\newcommand{\qbm}{{\bm q}}

\newcommand{\obm}{{\bm o}}
\newcommand{\tbm}{{\bm t}}
\newcommand*{\circled}[1]{\lower.7ex\hbox{\tikz\draw (0pt, 0pt)%
    circle (.4em) node {\makebox[0.6em][c]{\footnotesize #1}};}}
\newcommand{\cbm}{{\bm c}}

\newcommand{\bbm}{{\bm b}}
\newcommand{\pbm}{{\bm p}}
\newcommand{\hbm}{{\bm h}}
\newcommand{\deltabm}{{\bm \delta}}
\newcommand{\Abm}{{\bm A}}
\newcommand{\abm}{{\bm a}}
\newcommand{\ubm}{{\bm u}}
\newcommand{\gammabm}{{\bm \gamma}}
\newcommand{\omegabm}{{\bm \omega}}

\newcommand{\Mbf}{{\bf M}}
\newcommand{\Kbf}{{\bf K}}

\newcommand{\xbm}{{\bm x}}
\newcommand{\mubm}{{\bm \mu}}
\usepackage{microtype}      

\usepackage{nicefrac}       

\newcommand{\Pcal}{{\mathcal{P}}}
\newcommand{\Rbb}{{\mathbb{R}}}
\newcommand{\Ebb}{{\mathbb{E}}}
\newcommand{\thetabm}{{\bm \theta}}

\newcommand{\Lcal}{{\mathcal{L}}}

\newcommand{\Scal}{{\mathcal{S}}}
\newcommand{\Gcal}{{\mathcal{G}}}

\newcommand{\Ecal}{{\mathcal{E}}}
\newcommand{\Dcal}{{\mathcal{D}}}
\newcommand{\Acal}{{\mathcal{A}}}
\newcommand{\Kcal}{{\mathcal{K}}}

\newcommand{\Xcal}{{\mathcal{X}}}
\newcommand{\Vcal}{{\mathcal{V}}}
\newcommand{\Ycal}{{\mathcal{Y}}}
\newcommand{\Ccal}{{\mathcal{C}}}

\newcommand{\Mcalhat}{{\hat{\mathcal{M}}}}
\newcommand{\Mcal}{{\mathcal{M}}}
\newcommand{\Wcal}{{\mathcal{W}}}

\usepackage{rotating}
\usepackage{xcolor}

\newcommand{\nop}[1]{}
\newcommand{\myetal}{\textit{et al }}
 
\begin{document}

\title{Deep Intellectual Property Protection: A Survey}

\author{\IEEEauthorblockN{Yuchen Sun, Tianpeng Liu, Panhe Hu, Qing Liao, Shaojing Fu,
Nenghai Yu, Deke Guo, Yongxiang Liu${}^*$, 
Li Liu${}^*$ \\[-10pt]\thanks{
Y. Sun (sunyuchen18@nudt.edu.cn) and D. Guo (guodeke@gmail.com) are with Science and Technology Laboratory on Information Systems Engineering, the College of Systems Engineering, National University of Defense Technology (NUDT), Changsha, China. Y. Liu (lyx\_bible@sina.com), L. Liu (liuli\_nudt@nudt.edu.cn), T. Liu
(liutianpeng2004@nudt.edu.cn), and P. Hu (hupanhe13@nudt.edu.cn) are with the College of Electronic
Science, NUDT, Changsha, Hunan, China. Q. Liao (liaoqing@hit.edu.cn) is with the Harbin Institute of Technology, Shenzhen, China. N. Yu (ynh@ustc.edu.cn) is with the School of Cyber Science and Security, University of Science and Technology of China, Hefei, China. S. Fu (fushaojing@nudt.edu.cn) is with the College of Computer, NUDT, Changsha China.}
\thanks{Corresponding authors: Yongxiang Liu and Li Liu.}
\thanks{This work was supported by the National Key Research and Development Program of China No. 2021YFB3100800.}}}
\markboth{Submission To IEEE Communications Surveys \& Tutorials}%
{Sun \MakeLowercase{\textit{et al.}}: }

\IEEEtitleabstractindextext{%
\begin{abstract}
\textcolor{black}{Deep Neural Networks (DNNs), from AlexNet to ResNet to ChatGPT, have made revolutionary progress in recent years, and are widely used in various fields. The high performance of DNNs requires a huge amount of high quality data, expensive computing hardwares and excellent DNN architectures that are costly to obtain. Therefore,
trained DNNs are becoming valuable assets and must be considered Intellectual Property (IP)
of the legitimate owner who created them, in order to protect trained DNN models from illegal
reproduction, stealing, redistribution, or abuse.  Although being a new emerging and interdisciplinary field, numerous DNN model IP protection methods have been proposed. Given this period of rapid evolution, the goal of this paper is to provide a comprehensive survey of two mainstream DNN IP protection methods:  deep watermarking and deep fingerprinting, with a proposed taxonomy. More than 190 research contributions are
included in this survey, covering many aspects of Deep IP Protection: problem definition, 
main threats and challenges,  merits and demerits of deep watermarking and deep fingerprinting methods, evaluation metrics, and performance discussion. We finish the survey by identifying promising directions for future research.}

\end{abstract}
}
\maketitle

\IEEEdisplaynontitleabstractindextext
\IEEEpeerreviewmaketitle






\section{Introduction} \label{sec:intro} 
\IEEEPARstart{D}{uring} 
\textcolor{black}{the past decade, under the joint driving force of big data and 
the availability of powerful computing hardware, Deep Neural Networks (DNNs)  
\textcolor{black} {\cite{krizhevsky2012imagenet,LeCun15,khan2022vitsurvey, Jabbar2021GANSurvey}}  have made revolutionary progress, and are now applied in a wide spectrum of fields including computer vision \cite{liu1809deepa,taigman2014deepface}
, speech recognition~\cite{li2022ASRSurvey}, natural language processing \cite{devlin2018bert}, autonomous driving \cite{chen2015deepdriving}, cancer detection \cite{esteva2017dermatologist}, medicine discovery \cite{stokes2020deep}, playing complex games~\cite{silver2018general,vinyals2019grandmaster,brown2019superhuman}, recommendation systems~\cite{da2020recommendationsurvey}, and robotics \cite{2022roboticsurvey}. \textcolor{black}{Very recently, When researchers are discussing whether the field of Artificial Intelligence (AI) is approaching another AI winter or not, the remarkable capabilities of very large foundation models ChatGPT and GPT4 have evoked new hope for achieving Artificial General Intelligence (AGI) and attracted enormous attention.  Now
the field is growing dizzyingly fast again.} Acknowledgedly, it is of a high cost to obtain high-quality models, as training accurate, large DNNs requires the following essential resources: 
\begin{itemize}[leftmargin=*]
	\setlength{\topsep}{0pt}
	\setlength{\itemsep}{0pt}
	\setlength{\parsep}{0pt}
	\setlength{\parskip}{0pt} 
  \item  \textit{Massive amounts of high-quality data} that are costly to gather and manage, usually requiring the painstaking efforts of experienced human annotators or even experts. \textcolor{black}{It plays the primary role in the success of deep learning such as ImageNet~\cite{dataset_deng2009imagenet}, JFT300M~\cite{dataset_sun2017JFT300M}, MSCOCO~\cite{dataset_lin2014MSCOCO}, Google Landmark~\cite{dataset_weyand2020googlelandmark}, and many more high-value commercial datasets that are not open source.} 
  \item \textit{Expensive high-performance computing hardware} like GPUs; \textcolor{black}{In general, a larger model trained on a larger training dataset will have better performance~\cite{background_nakkiran2021deep}. Meanwhile, it also requires more computing resources including not only strong computing devices but also high-performance distributed architectures.}
  \item \textit{Advanced DNNs themselves} that require experienced experts to design their architectures and tune their hyperparameters, which is also costly and time consuming. 
\end{itemize} 
}

\noindent \textcolor{black}{It is therefore overwhelmingly expensive for training large foundation models like ChatGPT, GPT4\footnote{GPT4 costs \$12 million to train once}, DALL$\cdot$E}. As aforementioned, well-trained DNN models have become valuable assets. In addition, recently ``Machine Learning as a Service (MLaaS)'' has emerged as a popular paradigm to enable many people other than machine learning experts to train DNNs. Therefore, trained DNNs have high business value, \textcolor{black}{and can be considered the Intellectual Property (IP) of the legitimate owner who created it.}

\textcolor{black}{Because of the great value of trained DNNs, many security-related issues  have arisen~\cite{stealingsurvey_sun2021mind,survey_oliynyk2022summer}. Adversaries can illegally download, steal, redistribute, or abuse the trained DNN models without respecting the Intellectual Property (IP) of legitimate parties, leading to serious loss of the model owners. Therefore, to prevent such malicious attacks, there is a pressing need to develop methods to confidentially yet robustly protect the trained models and their IP, enhancing the trustworthiness of DNN models.}



\begin{figure*}[htbp]
	\centering
	\includegraphics[width=.98\textwidth]{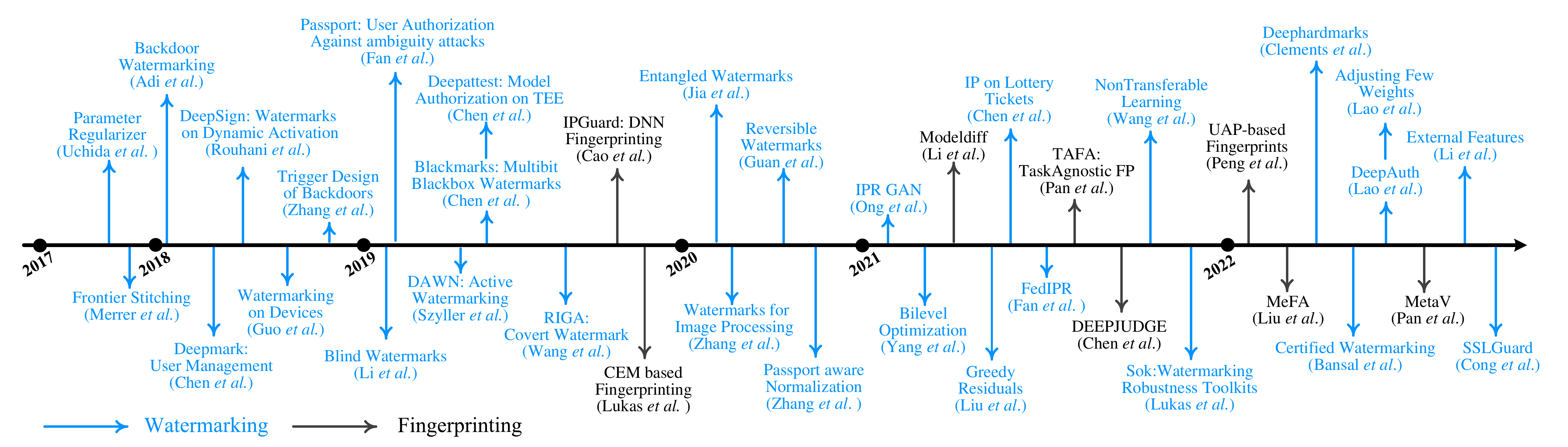} 
 \vspace{-0.1in}
	\caption{Chronological overview of representative methods for Deep IP Protection from the first contribution in 2017 to the latest, including model watermarking and fingerprinting. Watermarking is are to embed watermark messages by modifying target models. Some of them embed a bit string with a regularizer on selected parameters~\cite{white_uchida2017embedding, white_chen2018deepmarks, white_rouhani2018deepsigns} or normalization layer~\cite{passport_fan2019rethinking, passport_fan2021deepip}. 
	The others finetune target models by trigger samples with predefined labels~\cite{black_le2020adversarial, black_adi2018turning, black_zhang2018protecting,black_chen2019blackmarks}.
	In contrast, fingerprinting-based methods are to depict the decision boundary by constructing near-boundary samples and test metrics~\cite{fingerprinting_cao2021ipguard, fingerprinting_lukas2019deep, fingerprinting_peng2022uapfp,fingerprinting_liu2022mefa,  fingerprinting_li2021modeldiff,fingerprinting_pan2022metav,  fingerprinting_chen2021copy,fingerprinting_pan2021tafa}. Most of these methods compare the similarity between model with a single test metric~\cite{fingerprinting_cao2021ipguard, fingerprinting_lukas2019deep, fingerprinting_pan2021tafa, fingerprinting_peng2022uapfp, fingerprinting_liu2022mefa, fingerprinting_pan2022metav}. Several methods design multiple test metrics for stronger robustness~\cite{fingerprinting_chen2021copy}. 
	See Section~\ref{sec:Protect_Invasive}~and~\ref{sec:Protect_Non_Invasive} for details.
	} 
	\label{Chronologicaloverview}
 \vspace{-0.1in}
\end{figure*}

\textcolor{black}{The ideal protection for Deep IP is to prevent model stealing beforehand. However, the leakage risks of DNN models are inevitably facing various stealing strategies} 
\cite{stealingsurvey_sun2021mind,survey_oliynyk2022summer}. Since model stealing cannot be prevented beforehand, the copyright claiming and tracing of leaked models are essential for Deep IP protection. For classical digital multimedia, some identification information is embedded into the original multimedia data to declare copyright without hurting data usage \cite{book_borra2018digital,book_xiang2017audio}. 
Inspired by such copyright marking technologies, 
model watermarking and fingerprinting are currently two dominant techniques for DNN IP protection which is a newly emerging research field. Many methods have been proposed recently, with some representative works shown in Fig.~\ref{Chronologicaloverview}, however this field is still in its infancy.


\textcolor{black}{Deep Watermarking~\cite{white_uchida2017embedding} embeds some unique identification information  into the inputs, the model itself (model parameters, gradients, structures, \etc) or outputs. A defender can decide whether a model is pirated by extracting the injected watermark.  It is an invasive solution, meaning that a protected model has to be modified from the original version. The early works focused on model watermarking techniques, and have made some progress in robustness, security and other issues. However, such solutions require model modifications, resulting in potential security risks and additional burdens. Recently, some works have explored fingerprinting techniques. Deep Fingerprinting~\cite{fingerprinting_cao2021ipguard} extracts some unique model properties like decision boundaries as the ``fingerprints''~\cite{book_wang2016fingerprinting} (like human fingerprints that represent human identities) of DNN models. It is a non-invasive solution with no model modification. }

\textcolor{black}{The design of ideal marking techniques is complex and challenging
due to the balance among the competing objectives, such as hiding capacity, transparency, and robustness.}
So far, although Deep IP protection has received wide attention and many outstanding works have been proposed, the existing schemes still cannot meet the demands of various applications (\emph{e.g.}, model fidelity, Quality-of-IP, and Efficiency-of-IP) and the threat of endless attack modes (\emph{e.g.}, IP removal or ambiguity). Deep IP protection is still in its early stages, which motivates us to systematically review the recent progress of this field, identify the main challenges and open problems impeding its development, to clarify the promising future directions. On account of the diversity of potential Deep IP solutions, there is still a lack of systematic investigation. To fill this gap, we present this survey for a comprehensive overview.

\subsection{Differences with Related Surveys}
Although several surveys on Deep IP have been published~\cite{survey_oliynyk2022summer, survey_regazzoni2021protecting, survey_xue2022intellectual, sok_lukas2021sok, survey_boenisch2021systematic, survey_li2021nc, survey_chen2018performance, survey_fkirin2022copyright_digital}, however, to the best of our knowledge, there is no survey to provide a comprehensive review covering DNN watermarking and fingerprinting techniques. We present a brief review of related surveys and emphasize our new contributions.  
\par
Some surveys~\cite{survey_li2021nc, survey_fkirin2022copyright_digital} only focus on DNN watermarking. 
For example, \cite{survey_barni2021dnn} discusses the dissimilarities between media and DNN watermarking and identifies some challenges for future research. Ffkirin \textit{et al}~\cite{survey_fkirin2022copyright_digital} review some watermarking methods and test the robustness against the fine-tuning attack using several optimizers on MNIST and CIFAR10 datasets. Li \textit{et al}~\cite{survey_li2021nc} present a novel taxonomy to divide DNN watermarking methods into static and dynamic categories. Only a small part of watermarking methods are included in them. 
Some surveys~\cite{survey_regazzoni2021protecting, survey_xue2022intellectual} discuss fingerprinting techniques; however, they either lack a clear definition or the investigated research works are not comprehensive enough to date. Regazzoni \textit{et al}~\cite{survey_regazzoni2021protecting} only review 13 research works. Boenisch \textit{et al}~\cite{survey_boenisch2021systematic} propose a definition of fingerprinting similar to ours, but only two works based on fingerprinting are reviewed.
Xue \textit{et al}~\cite{survey_xue2022intellectual} proposed a taxonomy on Deep IP Protection from 6 views: scenario, mechanism, capacity, type, function, and target models, and divided the attack methods into three groups: model modifications, evasion/removal attacks, and active attacks. Most works identified as fingerprinting in~\cite{survey_xue2022intellectual} are put into the watermarking category in ours due to different definition.  
\cite{survey_peng2022WWWJ} reviews some typical Deep IP Protection methods, which can be divided into passive protection and active protection from the view of protection purpose, and into static and dynamic from the view of technical implementation. In contrast, we focus on passive protection and classify watermarking and fingerprinting methods as more fine-grained.
From the perspective of testing protocol, Chen \textit{et al}~\cite{survey_chen2018performance} propose an experimental performance comparison on fidelity, robustness, and integrity. Lukas \textit{et al}~\cite{sok_lukas2021sok} propose Watermarking-Robust-Toolkit (WRT) to test the robustness of some model watermarking methods. \cite{black_lee2022evaluating} evaluates the robustness of some trigger-based methods.
However, they only compare some classical watermarking methods in the early stage, neglecting other metrics and advanced methods. 
Compared with these surveys, we propose the most comprehensive review to date including more than 190 research contributions and many aspects of Deep IP Protection. 
From the threats to Deep IP, Oliynyk \textit{et al}~\cite{survey_oliynyk2022summer} present a comprehensive review on model stealing and its defense from a high-level perspective. However, they mainly focus on stealing methods, and Deep IP Protection is only briefly mentioned.
Literature~\cite{survey_jagielski2020extraction, survey_he2020attack4} only reviews model attacks, like stealing and poisoning, without any glance at Deep IP Protection. These works show the serious threats to Deep IP and reflect the urgency of this survey from the attack side.

\textbf{Our major contributions can be summarized as follows:}  
\begin{itemize}[leftmargin=*]
	\setlength{\topsep}{0pt}
	\setlength{\itemsep}{0pt}
	\setlength{\parsep}{0pt}
	\setlength{\parskip}{0pt} 
\item To the best of our knowledge, this is the first systematic and comprehensive survey covering DNN watermarking and fingerprinting techniques for Deep Intellectual Property. 
\item We present the problem definition, evaluation criteria, testing protocol, main challenges, and threats of Deep IP Protection. 
\item A multi-view taxonomy is proposed to categorize existing methods and reveal the open problems. Moreover, we analyze the performance of representative approaches on the proposed evaluation criteria, as well as their underlying connections. 
\item We discuss the current limitations and give some suggestions for prospective research directions. 
\end{itemize}

\begin{figure}
\centering
\footnotesize
\begin{tikzpicture}[xscale=0.78, yscale=0.38]
\draw [thick, -] (0, 62.5) -- (0, 22); 
\node [right] at (-0.25, 63) {\textbf{Protecting Deep Intellectual Property}};
\draw [thick, -] (0, 62) -- (0.25, 62); \node [right] at (0.25, 62) {\textbf{Invasive Solutions: Watermarking
~(Section~\ref{sec:Protect_Invasive})}};
\draw [thick, -] (0.75, 61.5) -- (0.75, 27);
\draw [thick, -] (0.75, 61) -- (1, 61);\node [right] at (1, 61) {{\bf Inner-component-embedded Watermarking~(Section~\ref{sec:Protect_Invasive_Parameter})}};
\draw [thick, -] (1.5, 60.5) -- (1.5, 49);
\draw [thick, -] (1.5, 60) -- (1.75, 60);\node [right] at (1.75, 60) {\emph{Static Model Weights (Section~\ref{sec:Protect_Invasive_Parameter_Static})}};
\draw [thick, -] (2.25, 59.5) -- (2.25, 57);
\draw [thick, -] (2.25, 59) -- (2.5, 59);\node [right] at (2.5, 59) {{Original Weights}: One WM~\cite{white_uchida2017embedding}, Multiple WMs~\cite{white_chen2018deepmarks}};
\draw [thick, -] (2.25, 58) -- (2.5, 58);\node [right] at (2.5, 58) {{Transformed Weights}: RIGA~\cite{white_wang2021riga}, greedy residuals~\cite{white_liu2021residuals}};
\draw [thick, -] (2.25, 57) -- (2.5, 57);\node [right] at (2.5, 57) {{Integrity verification}: reversible~\cite{white_guan2020reversible}, fragile~\cite{white_botta2021neunac, whtie_chen2019deepattest}};
\draw [thick, -] (1.5, 56) -- (1.75, 56);\node [right] at (1.75, 56) {\emph{Dynamic Inner Components (Section~\ref{sec:Protect_Invasive_Parameter_Dynamic})
}};
\draw [thick, -] (2.25, 55.5) -- (2.25, 53);
\draw [thick, -] (2.25, 55) -- (2.5, 55);\node [right] at (2.5, 55) {{Hidden-layer Activations}: Deepsigns~\cite{white_rouhani2018deepsigns}};
\draw [thick, -] (2.25, 54) -- (2.5, 54);\node [right] at (2.5, 54) {{Model Gradients}: MOVE~\cite{white_li2022defending,both_li2022move}};
\draw [thick, -] (2.25, 53) -- (2.5, 53);\node [right] at (2.5, 53) {{Watermarks from feature space to data space}~\cite{data_sablayrolles2020radioactive}};
\draw [thick, -] (1.5, 52) -- (1.75, 52);\node [right] at (1.75, 52) {\emph{Model Structures (Section~\ref{sec:Protect_Invasive_Parameter_Structure})}};
\draw [thick, -] (2.25, 51.5) -- (2.25, 50);
\draw [thick, -] (2.25, 51) -- (2.5, 51);\node [right] at (2.5, 51) {{Model Structures for Watermarking}};
\draw [thick, -] (2.25, 50) -- (2.5, 50);\node [right] at (2.5, 50) {{Watermarking High-Performance Structures}:~\cite{gray_lou2021meets}, \cite{lottery_chen2021you}};
\draw [thick, -] (1.5, 49) -- (1.75, 49);\node [right] at (1.75, 49) {\emph{Extra Components (Section~\ref{sec:Protect_Invasive_Parameter_Extra})}};
\draw [thick, -] (2.25, 48.5) -- (2.25, 46);
\draw [thick, -] (2.25, 48) -- (2.5, 48);\node [right] at (2.5, 48) {{Extra processing on normalization layers}: \cite{passport_fan2019rethinking, passport_fan2021deepip, passport_zhang2020passport}};
\draw [thick, -] (2.25, 47) -- (2.5, 47);\node [right] at (2.5, 47) {{Extra processing on RNN hidden states}: \cite{passport_lim2022rnn}};
\draw [thick, -] (2.25, 46) -- (2.5, 46);\node [right] at (2.5, 46) {{Subnetworks as Watermarks:} HufuNet~\cite{new_lv2023robustness}};
\draw [thick, -] (0.75, 45) -- (1, 45);\node [right] at (1, 45) {\textbf{Trigger-injected Watermarking~(Section~\ref{sec:Protect_Invasive_Trigger})}};
\draw [thick, -] (1.5, 44.5) -- (1.5, 36);
\draw [thick, -] (1.5, 44) -- (1.75, 44);\node [right] at (1.75, 44) {\emph{Generating Trigger Samples (Section~\ref{sec:Protect_Invasive_Trigger_Generation})}};
\draw [thick, -] (2.25, 43.5) -- (2.25, 40);
\draw [thick, -] (2.25, 43) -- (2.5, 43);\node [right] at (2.5, 43) {{Directly-selected samples}~\cite{white_rouhani2018deepsigns,black_adi2018turning, black_chen2019blackmarks, black_jia2022srdw,new_sofiane2021yes,black_zhang2018protecting,black_lao2022fewweights}};

\draw [thick, -] (2.25, 42) -- (2.5, 42);\node [right] at (2.5, 42) {{Simply-processed samples}~\cite{black_jia2021entangled, black_zhang2018protecting, black_guo2018watermarking, black_zhu2020secure, black_maung2021piracy, black_li2022untargetedbackdoorwatermark}};
\draw [thick, -] (2.25, 41) -- (2.5, 41);\node [right] at (2.5, 41) {{Generated samples w/ hidden copyright info}~\cite{black_li2019prove}};
\draw [thick, -] (2.25, 40) -- (2.5, 40);\node [right] at (2.5, 40) {{Optimized samples}~\cite{black_le2020adversarial, black_jia2021entangled,fragile_lao2022deepauth,black_yang2021robust}};
\draw [thick, -] (1.5, 39) -- (1.75, 39);\node [right] at (1.75, 39) {\emph{Labeling Trigger Samples (Section~\ref{sec:Protect_Invasive_Trigger_Label})}};
\draw [thick, -] (2.25, 38.5) -- (2.25, 37);
\draw [thick, -] (2.25, 38) -- (2.5, 38);\node [right] at (2.5, 38) {{Hard labels: target, random, new, clean, rearranged}};
\draw [thick, -] (2.25, 37) -- (2.5, 37);\node [right] at (2.5, 37) {{Soft labels: Distance vectors~\cite{both_li2022move}, Cosine signals~\cite{black_charette2022cosine}}};
\draw [thick, -] (1.5, 36) -- (1.75, 36);\node [right] at (1.75, 36) {\emph{Embedding Paradigms (Section~\ref{sec:Protect_Invasive_Trigger_Embedding})}};
\draw [thick, -] (2.25, 35.5) -- (2.25, 30);
\draw [thick, -] (2.25, 35) -- (2.5, 35);\node [right] at (2.5, 35) {{Loss functions: CE, SNNL~\cite{black_jia2021entangled,new_wu2022cits}, KL~\cite{new_ren2022cybersecurity}, InvIB~\cite{black_wang2022ntl}}};
\draw [thick, -] (2.25, 34) -- (2.5, 34);\node [right] at (2.5, 34) {{Retraining or finetuning}};
\draw [thick, -] (2.25, 33) -- (2.5, 33);\node [right] at (2.5, 33) {{The whole or partial model}: ~\cite{black_lao2022fewweights, black_yang2021robust}};
\draw [thick, -] (2.25, 32) -- (2.5, 32);\node [right] at (2.5, 32) {{Model-only or model-trigger-joint optimization~\cite{black_yang2021robust, black_li2022untargetedbackdoorwatermark}}};
\draw [thick, -] (2.25, 31) -- (2.5, 31);\node [right] at (2.5, 31) {{Original or transformed weights: exp~\cite{black_namba2019robust}, smooth~\cite{black_bansal2022certified}}};
\draw [thick, -] (2.25, 30) -- (2.5, 30);\node [right] at (2.5, 30) {{Embedding directly or indirectly: gradients~\cite{fedwm_li2022fedipr}}};
\draw [thick, -] (0.75, 29) -- (1, 29);\node [right] at (1, 29) {\textbf{Output-embedded Watermarking~(Section~\ref{sec:Protect_Invasive_Output})}};
\draw [thick, -] (0.75, 28) -- (1, 28);\node [right] at (1, 28) {\textbf{Combined Usage of DNN Watermarking}}; 
\draw [thick, -] (0.75, 27) -- (1, 27);\node [right] at (1, 27) {\textbf{Watermarking for Specific Applications~(Section~\ref{sec:Protect_Invasive_Scenarios})}};
\draw [thick, -] (1.5, 26.5) -- (1.5, 23);
\draw [thick, -] (1.5, 26) -- (1.75, 26);\node [right] at (1.75, 26) {\emph{Self-supervised learning}: SSLGuard~\cite{selfsuper_cong2022sslguard}};
\draw [thick, -] (1.5, 25) -- (1.75, 25);\node [right] at (1.75, 25) {\emph{Transfer learning}: SRDW~\cite{black_jia2022srdw}};
\draw [thick, -] (1.5, 24) -- (1.75, 24);\node [right] at (1.75, 24) {\emph{Federated learning}: FedIPR~\cite{fedwm_li2022fedipr}};
\draw [thick, -] (1.5, 23) -- (1.75, 23);\node [right] at (1.75, 23) {\emph{Other AI Applications}: GAN~\cite{passport_ong2021iprgan}, Code~\cite{black_sun2022coprotector}};
\draw [thick, -] (0, 22) -- (0.25, 22); \node [right] at (0.25, 22) {\textbf{Non-Invasive Solutions: Fingerprinting~(Section~\ref{sec:Protect_Non_Invasive})}};
\draw [thick, -] ((0.75, 21.5) -- ((0.75, 18);
\draw [thick, -] (0.75, 21) -- (1, 21);\node [right] at (1, 21) {{\bf Comparing Model Weights~(Section~\ref{sec:Protect_Non_Invasive_Weights})}};
\draw [thick, -] (1.5, 20.5) -- (1.5, 19);
\draw [thick, -] (1.5, 20) -- (1.75, 20);\node [right] at (1.75, 20) {\emph{Training Path Recovery}: PoL~\cite{fingerprinting_jia2021proof}};
\draw [thick, -] (1.5, 19) -- (1.75, 19);\node [right] at (1.75, 19) {\emph{Weight Hash: Random Projections}~\cite{fingerprinting_zheng2022nonrepudiable} \emph{or Learnable Hash}~\cite{fingerprinting_chen2022perceptualhash, fingerprinting_xiong2022neural}};
\draw [thick, -] (0.75, 18) -- (1, 18);\node [right] at (1, 18) {{\bf Comparing Model Behaviors~(Section~\ref{sec:Protect_Non_Invasive_Behaviors})}};
\draw [thick, -] (1.5, 17.5) -- (1.5, 0);
\draw [thick, -] (1.5, 17) -- (1.75, 17);\node [right] at (1.75, 17) {\emph{Model \& Data Preparation~(Section~\ref{sec:Protect_Non_Invasive_Behaviors_Pre})}};
\draw [thick, -] (2.25, 16.5) -- (2.25, 15);
\draw [thick, -] (2.25, 16) -- (2.5, 16);\node [right] at (2.5, 16) {{Data: In-distribution or Out-of-distribution
}};
\draw [thick, -] (2.25, 15) -- (2.5, 15);\node [right] at (2.5, 15) {{Model: Retraining or augmentation w/o retraining
}};
\draw [thick, -] (1.5, 14) -- (1.75, 14);\node [right] at (1.75, 14) {\emph{Test Case Generation~(Section~\ref{sec:Protect_Non_Invasive_Behaviors_Case})}}; 
\draw [thick, -] (2.25, 13.5) -- (2.25, 12);
\draw [thick, -] (2.25, 13) -- (2.5, 13);\node [right] at (2.5, 13) {Selected or augmented samples: SAC~\cite{fingperprinting_guan2022samplecorrelation}, MeFA~\cite{fingerprinting_liu2022mefa}
};
\draw [thick, -] (2.25, 12) -- (2.5, 12);\node [right] at (2.5, 12) {Optimized Near-Boundary Samples
};
\draw [thick, -] (3, 11.5) -- (3, 9);
\draw [thick, -] (3, 11) -- (3.25, 11);\node [right] at (3.25, 11) {{From Real Samples}:~\cite{fingerprinting_cao2021ipguard,fingerprinting_lukas2019deep,fingerprinting_zhao2020afa,fingerprinting_yang2022metafinger,fingerprinting_pan2022metav,fingerprinting_wang2021deepfool,fingerprinting_peng2022uapfp,fingerprinting_li2021modeldiff}};
\draw [thick, -] (3, 10) -- (3.25, 10);\node [right] at (3.25, 10) {{From Random Seeds}:~\cite{fingerprinting_wang2021characteristic,fingerprinting_pan2021tafa,fingerprinting_chen2022teacherFP}
};
\draw [thick, -] (3, 9) -- (3.25, 9);\node [right] at (3.25, 9) {{Fragile Test Cases}:~\cite{fingerprinting_he2019sensitive,fingerprinting_wang2022publicheck, fingerprinting_wang2021intrinsic}}; 
\draw [thick, -] (1.5, 8) -- (1.75, 8);\node [right] at (1.75, 8) {\emph{Test Metric Designs~(Section~\ref{sec:Protect_Non_Invasive_Behaviors_Metric})}};
\draw [thick, -] (2.25, 7.5) -- (2.25, 5);
\draw [thick, -] (2.25, 7) -- (2.5, 7);\node [right] at (2.5, 7) {Model outputs: hard or soft outputs~\cite{fingerprinting_peng2022uapfp,fingerprinting_li2021modeldiff,fingerprinting_liu2022mefa,fingperprinting_guan2022samplecorrelation,fingerprinting_maini2021datasetinference,fingerprinting_dziedzic2022DI4selfsupervise}};
\draw [thick, -] (2.25, 6) -- (2.5, 6);\node [right] at (2.5, 6) {Inner components: neuron-wise, layer-wise~\cite{fingerprinting_pan2021tafa, fingerprinting_chen2021copy}};
\draw [thick, -] (2.25, 5) -- (2.5, 5);\node [right] at (2.5, 5) {Model properties: LIME weights~\cite{fingerprinting_jia2022Zest}, robustness~\cite{fingerprinting_chen2021copy}}; 
\draw [thick, -] (1.5, 4) -- (1.75, 4);\node [right] at (1.75, 4) {\emph{Fingerprint Comparison~(Section~\ref{sec:Protect_Non_Invasive_Behaviors_Compare})}};
\draw [thick, -] (2.25, 3.5) -- (2.25, 1);
\draw [thick, -] (2.25, 3) -- (2.5, 3);\node [right] at (2.5, 3) {Simple Similarity Thresholds};
\draw [thick, -] (2.25, 2) -- (2.5, 2);\node [right] at (2.5, 2) {Hypothesis Testing};
\draw [thick, -] (2.25, 1) -- (2.5, 1);\node [right] at (2.5, 1) {\makecell{Learnable Meta-Classifiers}};
\draw [thick, -] (1.5, 0) -- (1.75, 0);\node [right] at (1.75, 0) {\emph{Target Scenarios, Models, \& Functions~(Section~\ref{sec:Protect_Non_Invasive_Behaviors_Scene})}};
\end{tikzpicture}
\vspace{-0.2in}
\caption{A taxonomy of representative methods for Deep IP Protection.} 
\label{fig:overview} 
\end{figure}

{\textbf{Paper Organization.}}  The remainder of this paper is organized as follows: The problem, motivation, functions, evaluation criteria, and challenges are summarized in Section \ref{sec:background}. 
In Section~\ref{sec:framework}, we present the framework of Deep IP protection from the views of Deep IP construction and verification. In Section~\ref{sec:Attack}, we introduce the  threats to Deep IPs. 
\textcolor{black}{In Section~\ref{sec:Protect_Invasive} and \ref{sec:Protect_Non_Invasive}, we present a taxonomy of the existing methods for Deep IP protection including the invasive watermarking methods and the non-invasive fingerprinting methods, as shown in Fig.~\ref{fig:overview}.} 
Then in Section \ref{sec:performance}, we provide an overall discussion of their performance (Tables~\ref{tab:performance}). Followed by Section~\ref{sec:future} we conclude the important future research outlook.

\section{Background \& Problems}\label{sec:background}
\begin{figure}
	\centering
	\includegraphics[width=.5\textwidth]{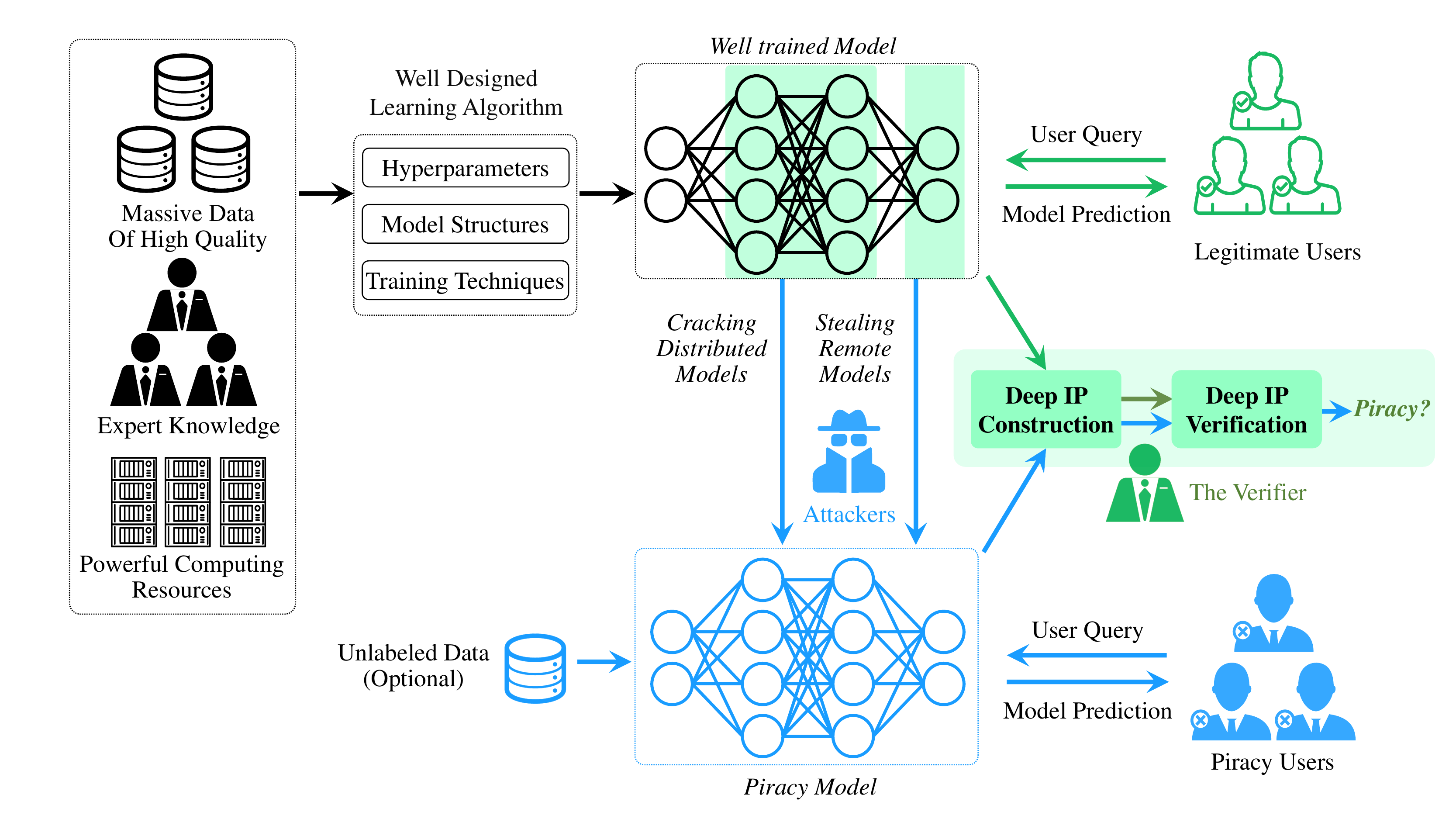}
 \vspace{-0.2in}
	\caption{The Motivations of Deep IP Protection. A well-trained DNN model requires a heavy drain on multiple resources including data, hardware, and algorithm designs. The MLaaS paradigm allows legitimate users without strong training capability to access well-trained models through model distribution or remote APIs of service providers. However, malicious users can also create a piracy version by stealing remote models or cracking distributed models. Therefore, Deep IP Protection, to construct an IP Identifier for the original model, is in urgent need. 
 } 
	\label{fig:piracy}
 \vspace{-0.2in}
\end{figure}



\subsection{\textcolor{black}{Deep Neural Networks}}
A Deep Neural Network (DNN), as a quintessential deep learning model, is of the goal to approximate the transformation function $\phi^*:\mathcal{X}\rightarrow \mathcal{Y}$ from the source probability distribution $\Xcal \in \Rbb^d$ to the target probability distribution $\Ycal \in \Rbb^k$. Take the image classification task as an example, the source distribution $\Xcal$ consists of the input images, and the target distribution $\Ycal$ consists of the class labels of the input images. Generally, a DNN has multiple layers between the input and output layers, such as fully connected layers, convolution layers, normalization layers, activation layers, etc. In theory, a DNN can approximate any continuous transformation on compact sets. 
\par
A DNN can be formulated as a parameterized function $\ybm\leftarrow\phi(\xbm;\thetabm)$, where $\thetabm$ denotes the trainable parameters of the DNN, $\xbm\in\mathcal{X}$ and $\ybm\in\mathcal{Y}$ are pairwise samples from the source and target distribution. To train the DNN $\phi$ to converge to the ideal transformation function $\phi^*$, a dataset $\{\X,\Ybf\}$ needs to be sampled in pairs from the source and target distribution $\{\mathcal{X}, \mathcal{Y}\}$ first. Then, a differentiable loss function $\Lcal$ should be determined to measure the approximation error between the DNN's outputs $\phi(\X;\thetabm)$ and the target $\ybm$. 
The loss function $\Lcal$ should be minimized by gradient descent on the trainable parameters $\thetabm$, as
\vspace{-0.04in}
\begin{equation}
\min_\thetabm \Lcal\left(\phi(\X;\thetabm), \Ybf\right)    
\vspace{-0.04in}
\end{equation}

So far, many kinds of DNN structures and training paradigms are proposed, such as Multilayer Perceptrons (MLPs), Convolutional Neural Networks (CNN), Transformers, Recurrent Neural Networks (RNNs), Generative Adversarial Networks (GANs) and Diffusion Models.
Benefiting from the unprecedented performance for a wide range of tasks like image classification, object detection, and natural language processing, DNNs show great commercial value across many industrial applications like medical, finance, advertising, etc.

\subsection{\textcolor{black}{Machine Learning as a Service (MLaaS)}}
Machine Learning as a Service (MLaaS) refers to cloud-based machine learning (ML) services provided by companies for developers or even off-the-shelf users to build and deploy ML models. Instead of building custom ML models from scratch, MLaaS allows users to leverage pre-trained models to solve ML problems. As shown in Fig.~\ref{fig:piracy}, it costs a lot to build and maintain high-performance pre-trained models on large-scale datasets, expert knowledge, and computing hardware. 
However, these high-cost, well-trained models are at serious risk of theft. There are two main implementations of MLaaS services: i) The first is to send the entire model, including parameters and structures, to the MLaaS users. 
Although around 60\% of model providers protect their models by licenses or encryption techniques, the models are still at high risk of being leaked while attackers are motivated and able to crack the distributed models~\cite{stealingsurvey_sun2021mind}. ii) The other way is to open remote MLaaS APIs of DNN models via a pay-per-query principle, where the models are deployed at the cloud servers of MLaaS providers and the clients upload the query data to the cloud. 
However, it suffers the threats of model extraction attacks. Adversaries can create a pirated copy of the original model by some \textit{query-prediction} pairs while the existing defense strategies are insufficient for endless attack mechanisms~\cite{survey_oliynyk2022summer}.
\par
Model leakage not only harms the business revenue of model owners but also leads to unpredictable consequences once criminals illegally steal and exploit the leaked models. For instance, malicious actors  can utilize leaked generative models (e.g., GAN or DDPM) to create fake faces and videos, or exploit the adversarial vulnerability of leaked models (e.g., Person Re-identification or Face Recognition) to bypass intelligent monitoring systems. Considering the incompleteness of current defense techniques against model stealing, it is imperative to explore posthoc protection strategies, \ie protecting the Intellectual Property of DNNs (Deep IP Protection). 

\subsection{\textcolor{black}{The Problem of Deep IP Protection}}\label{sec:background_problem}
Deep IP Protection enables model owners or law enforcement agencies to identify or even track the pirated copies of protected DNN models after pirates' infringement acts, providing legally-effective evidence to protect the interests of legitimate owners. And beyond that, as a key technique for Trustworthy Artificial Intelligence (TAI), Deep IP Protection can be extended to various fields such as model tamper-proofing, user authentication and management, and tracing the chains of model distribution. 
The workflow of Deep IP protection mainly consists of two stages:
\par
\textit{i) Deep IP Construction}. It is to embed or construct an \textbf{IP Identifier} of the original model as the unique information about the model Intellectual Property. A typical form of an IP Identifier is \textit{a key-message pair}, where the \textbf{secret keys} could be preset numerical matrice or input samples, and the \textbf{IP messages} could be a bit string or model outputs. Deep IP Construction is usually achieved by model watermarking or fingerprinting. 
\par
\textit{ii) Deep IP Verification}. 
It is to extract the IP Identifier given a suspect model, then compare the extracted IP Identifier with the constructed IP Identifier of the original model. A piracy model will retain an IP Identifier similar to that of the original model, but an individually trained model will not.

The formulation is detailed in Section~\ref{sec:framework}. Here, we show examples to describe the process of model watermarking and fingerprinting.  
\par
\begin{figure}[htbp]
	\centering
	\includegraphics[width=.45\textwidth]{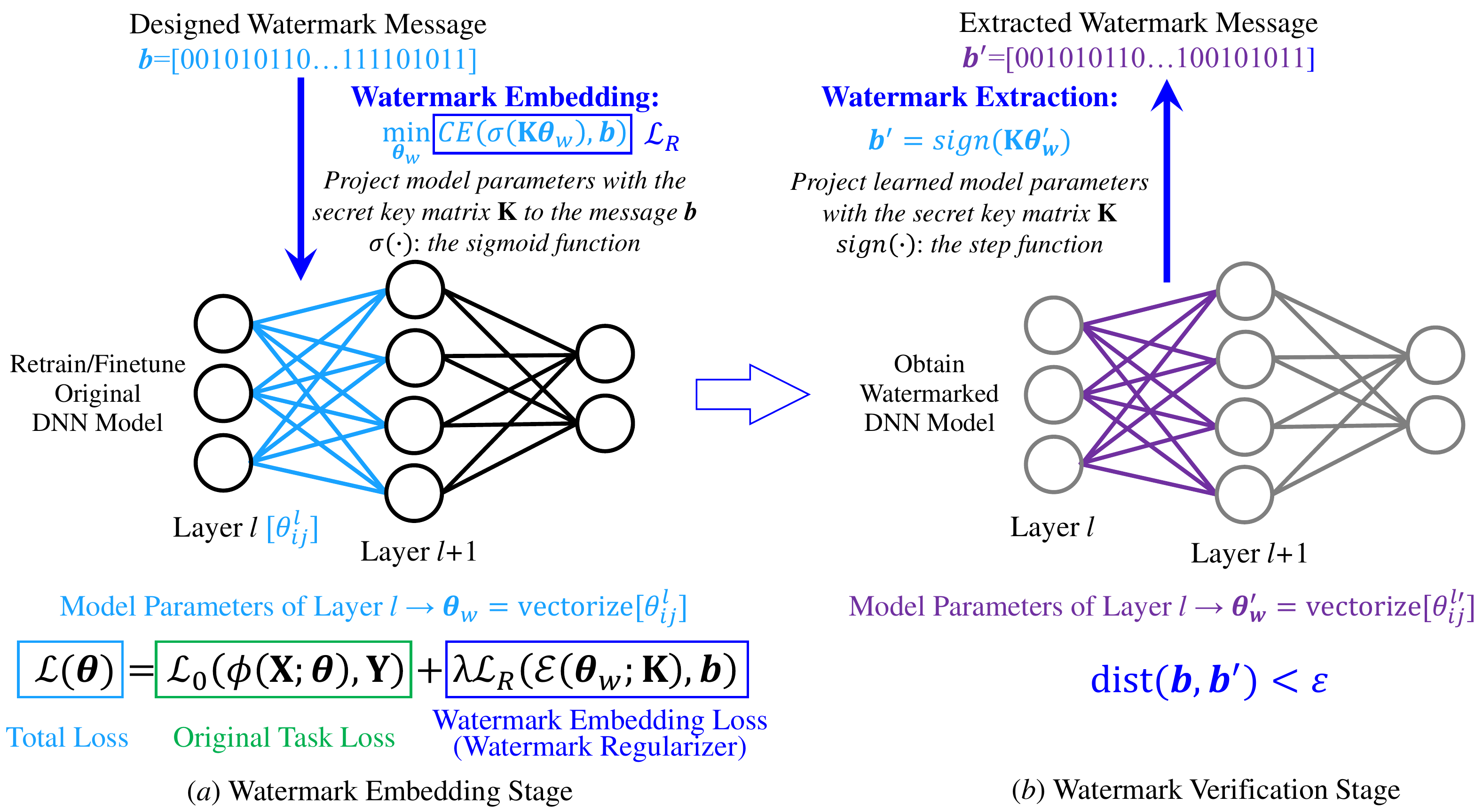}
 \vspace{-0.1in}
	\caption{An example of weight-based DNN model watermarking.}
	\label{fig:watermark_example}
 \vspace{-0.1in}
\end{figure}

\noindent
\textbf{An example of model watermarking}~\cite{white_uchida2017embedding,white_nagai2018digital}. Typical model watermarking is achieved by regularization techniques. It appends a watermark regularizer $\Lcal_R$ to the original task loss function $\Lcal_0$, and then embeds the IP Identifier ($\Kbf$, $\bbm$) by finetuning or retraining the original model $\phi$ to make the extracted watermark message $\Ecal(\Kbf;\thetabm_w)$ close to the preset message $\bbm$, which is defined as
\vspace{-0.06in}
\begin{equation}\label{eq:regularizer}
    \min_\thetabm \Lcal_0\left(\phi(\X;\thetabm), \Ybf\right) +\lambda\Lcal_R(\Ecal(\thetabm_w;\Kbf),\bbm),
    \vspace{-0.06in} 
\end{equation}
where $\thetabm_w \subset \thetabm$ is the vectorized host parameters selected from the whole model parameters $\thetabm$, and $\Ecal$ denotes the watermark extraction function. A typical form of watermarking regularizers could be a binary cross-entropy function with the formulation as 
\vspace{-0.06in}
\begin{equation}
\Lcal_R(\Ecal(\thetabm_w;\Kbf),\bbm) = \bbm\log(\sigma(\Kbf\thetabm_w))+(1\!-\!\bbm)\log(1\!-\!\sigma(\Kbf\thetabm_w)), 
\vspace{-0.06in}
\end{equation}
where $\sigma(\cdot)$ denotes the sigmoid function. Note that $\Ecal$ uses the differentiable sigmoid function in the construction stage $\Ecal(\thetabm_w;\Kbf)=\sigma(\Kbf\thetabm_w)$ but the step function in the verification stage $\Ecal(\thetabm_w;\Kbf)=\textrm{sign}(\Kbf\thetabm_w)$.
The process is detailed in Fig.~\ref{fig:watermark_example}.

\par\noindent
\textbf{An example of model fingerprinting}. Taking a typical model fingerprinting method as an example, the key of the IP identifier comprises several test cases near the model decision boundary, and the message refers to the model outputs of these test cases. Ideally, these test cases exhibit similar predictions on the original and pirated versions of the model while significantly differing from irrelevant models' predictions compared to those of the original model. This is usually achieved by constructing adversarial samples with appropriate transferability, called conferrable adversarial samples. These samples can be transferred to pirated models, but not to individually trained models that differ in at least one aspect of initial weights, hyperparameters, model structures, or training datasets. The scope of the individually trained models is determined by the applied fingerprinting method and the model protection range during the specific implementation process.

Deep IP protection has considered the following basic functions: 
\begin{itemize}[leftmargin=*]
	\setlength{\topsep}{0pt}
	\setlength{\itemsep}{0pt}
	\setlength{\parsep}{0pt}
	\setlength{\parskip}{0pt} 
\item
\textit{Copyright Verification (CopyVer):} It means to verify the ownership of a given model by comparing the IP Identifier between the given model and the owner's model. Most current works focus on this situation. In general, the IP Identifier for copyright verification should be robust to model modification, and irreversible for a strong unforgeability and non-removability. 
\item
\textit{Integrity Verification (InteVer):} It means to detect whether the protected model is modified illegally during model distribution or runtime execution. In contrast, InteVer is often accomplished through fragile or reversible IP Identifiers. The 'fragile' means that the IP Identifiers would be changed if the protected model is perturbed slightly, such as the inversion of several bits or finetuning at a small learning rate. The 'reversible' means that the protected model is able to return to the original model using a correct secret key, however, the modified version cannot. 
\item
\textit{User Management (UM):} It means to verify the authorized user of a given model. An intuitive application is to discover the source of the piracy model, \ie which user leaked it from, so as to facilitate subsequent accountability. 
\item
\textit{Access Control (AC):} It means to prevent unauthorized users from accessing the protected model or limit the model functionality (e.g., the applicability of data domains~\cite{black_wang2022ntl}). Only the users with correct secret keys can utilize the model for model inference, or finetune the model to adapt downstream tasks through transfer learning. Note that this survey does not focus on the access control schemes based on parameter encryption or obfuscated~\cite{encr_lin2020chaotic,encry_alam2022nn,encry_xue2021advparams}.
\item
\textit{Beyond DNN models: Data Protection}. Deep IPs also require to protect valuable data from illegal model training, which can be implemented in two ways or their combination: i) the first is to make the models trained on unauthorized datasets with detectable IP Identifiers belonging to the data owners; ii) or we can make the models trained on unauthorized datasets unusable, that is, significantly damaging the model accuracy. The former generally adopts the watermarking or fingerprinting methods~\cite{data_sablayrolles2020radioactive}, so we classify these methods according to the taxonomy as shown in Fig.~\ref{fig:overview}. Note that the latter is not the focus of this survey~\cite{data_peng2022learnability}.  
\end{itemize}

\begin{table*} [ht]
\centering
  \caption{The Metrics of Deep IP Protection with Description}
  \label{tab:metrics}
    \vspace{-0.1in}
    \scriptsize  
    \begin{tabular}{c|c|c}
    \toprule
    \textbf{Type} & \textbf{Criteria}  & \textbf{Description}\\
    \midrule
    \multirow{7}{*}{\makecell{Model\\Fidelity}} &  Model Accuracy & \multicolumn{1}{m{13.5cm}}{It should preserve the prediction accuracy of the protected model on original tasks.} \\ 
    \\[-2ex]\cline{2-3}\\[-1.5ex]
    &  Model Efficiency &  \multicolumn{1}{m{13.5cm}}{It means reducing the impact of IP construction on the efficiency of model inference and training, like introducing extra IP components, optimization objectives, training data, and training processes.}\\ 
    \\[-2ex]\cline{2-3}\\[-1.5ex]
    &  Adversarial Robustness & \multicolumn{1}{m{13.5cm}}{The model robustness against various adversarial attacks should not be greatly affected by constructing IP identifiers.}\\
    \\[-2ex]\cline{2-3}\\[-1.5ex]
    &  No Potential Risks & \multicolumn{1}{m{13.5cm}}{IP construction should not induce unexpected mispredictions by which attackers can manipulate predictions deterministically.}\\ 
    \midrule
    \multirow{4}{*}{\makecell{Quality\\Of IP\\(QoI)}} 
    & IP Robustness & \multicolumn{1}{m{13.5cm}}{IP identifiers should be robust against not only legal model modification (\emph{e.g.}, model lightweight or transfer learning) but also various malicious attacks (\emph{e.g.}, IP detection, evasion, removal, ambiguity, etc. See Detail in Section~\ref{sec:Attack}).}\\
    \\[-2ex]\cline{2-3}\\[-1.5ex]
    & IP Capacity & \multicolumn{1}{m{13.5cm}}{Its meanings include the bit number of IP identifiers' valid information payload and the theoretical upper bound.}\\
    \\[-2ex]\cline{2-3}\\[-1.5ex]
    & Others &  \multicolumn{1}{m{13.5cm}}{QoI has many more meanings, such as the uniqueness of IP identifiers,  the confidence/interpretability of verification results, etc.}\\ 
    \midrule
    \multirow{4}{*}{\makecell{Efficiency\\Of IP\\(EoI)}}
    & Scalability & \multicolumn{1}{m{13.5cm}}{Deep IP protection should focus on diverse IP functions and be adaptive to various downstream tasks and learning paradigms.}\\
    \\[-2ex]\cline{2-3}\\[-1.5ex]
    & \makecell{Agility of Construction}& \multicolumn{1}{m{13.5cm}}{IP defenders should construct and verify IP Identifiers with minimal overhead.}\\
    \\[-2ex]\cline{2-3}\\[-1.5ex]
    & \makecell{Slowness of Removal}& \multicolumn{1}{m{13.5cm}}{IP attackers have to remove or forge IP Identifiers at a much higher cost and accuracy loss than IP defender.}\\
    \bottomrule
\end{tabular}
\vspace{-0.1in}
\end{table*}
\begin{figure}[htbp]
	\centering
	\includegraphics[width=.46\textwidth]{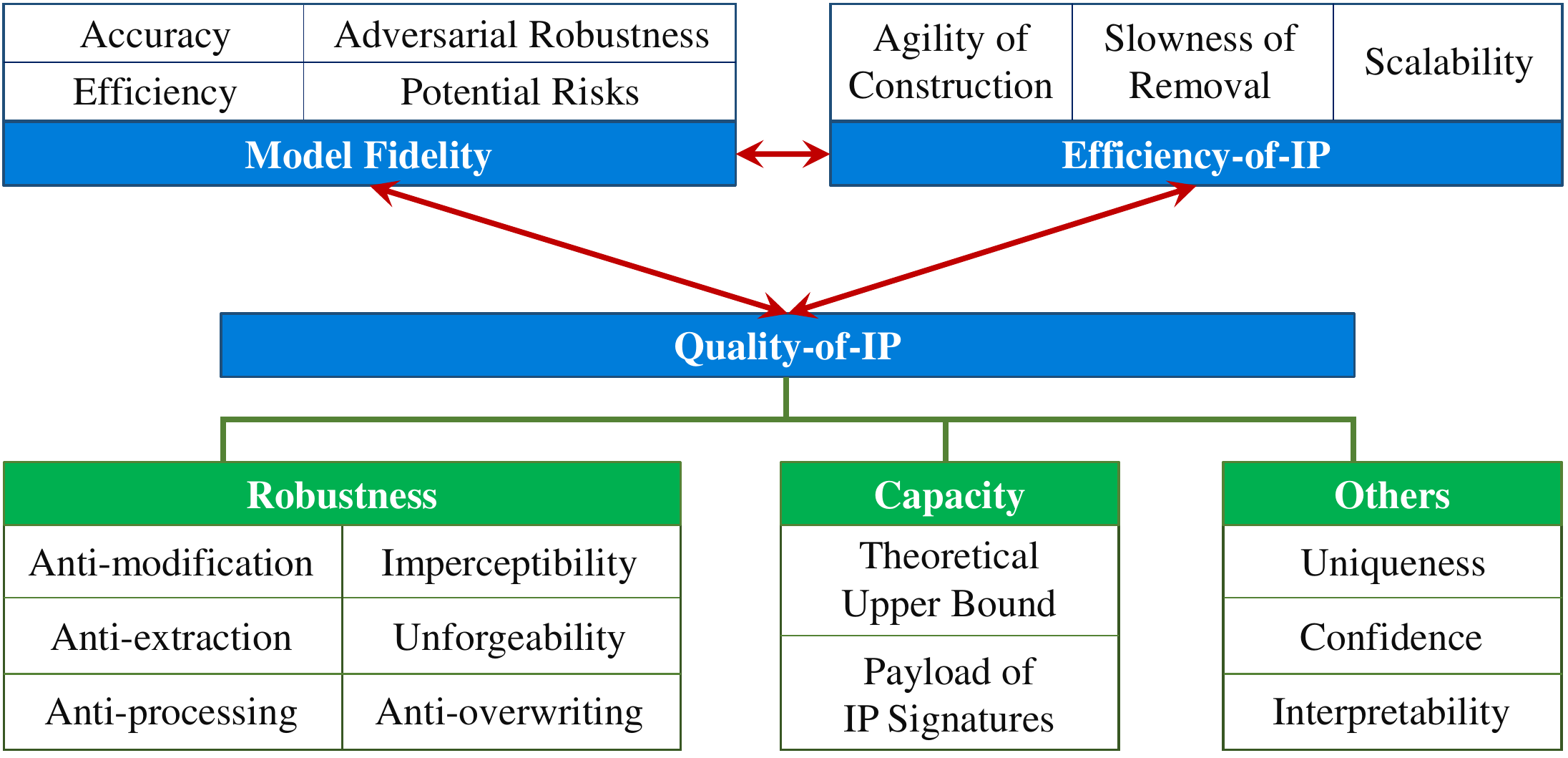}
 \vspace{-0.1in}
	\caption{Illustration for the requirements and the evaluation criteria of Deep IP Protection. An ideal protection method should jointly consider three conflicting criteria: model fidelity, Quality-of-IP (QoI), and Efficiency-of-IP (EoI). See details in Table~\ref{tab:metrics} and Section~\ref{sec:criteria}.
    }
	\label{fig:metric}
 \vspace{-0.1in}
\end{figure}
\subsection{\textcolor{black}{Performance Evaluation Criteria}} \label{sec:criteria}
\label{sec:background_criteria}
As shown in Table~\ref{tab:metrics}, this survey summarizes the evaluation criteria for Deep IP protection into the following three categories:
\subsubsection{Model Fidelity}
\textcolor{black}{IP identifiers are usually constructed by modifying the original model, such as finetuning model parameters or adding some extra watermark components to the model. One basic requirement is to ensure that the protected model and the original model have close performance. 
Fidelity is the metric to measure the performance similarity between the protected and the original models, including but not limited to accuracy, efficiency, adversarial robustness, and potential risks\footnote{The \textit{``potential risks''} means that a modified model may produce unexpected results on some inputs, or be easier for attackers to manipulate model predictions.}. } 
\subsubsection{Quality-of-IP (QoI)}
\textit{QoI} means that IP owners/verifiers have enough information and confidence to claim the model IP. This includes, but is not limited to: i) By comparing IP Identifiers, verifiers can confidently determine whether the suspect model is a copy of the protected model, with low false positive rates (FPR) for individually trained models, low false negative rates (FNR) for pirated models, and even convincing evidence; ii) Enough \textit{Capacity}, \ie the bit number of the effective IP information hidden in an IP Identifier like model copyright or user identification, is necessary to show strong infringement evidence; iii) \textit{Robustness}: IP Identifiers can be correctly extracted, even if IP identifiers, host signals, or the IP verification process are perturbed. A well-trained model often requires modification when applied to downstream tasks. Moreover, malicious attackers who create pirated copies generally try to detect and remove potential IP Identifiers, or to corrupt the IP Verification process via fake IP identifiers. An ideal scheme must work well in challenging real-world environments.  
\textit{Robustness} is recently the most concerned criterion and has the following aspects:
\begin{itemize}[leftmargin=*]
	\setlength{\topsep}{0pt}
	\setlength{\itemsep}{0pt}
	\setlength{\parsep}{0pt}
	\setlength{\parskip}{0pt} 
\item \textit{Anti-Modification and Anti-Extraction}: The IP Identifiers of protected models should remain integrally in the pirated models, that is, the modified or the extracted versions of the protected models, via fine-tuning, pruning, compression, distillation, transfer learning, or side-channel attacks. 
\item \textit{Anti-processing}: During remote IP verification, the queried samples or their returned predictions could be perturbed by MLaaS providers using pirated models, such as compression/noising on model inputs or perturbation/smoothing on model outputs. 
\item \textit{Imperceptibility}: Attackers cannot discover the IP identifiers hidden in protected models. 
\item \textit{Unforgeability}: Even if an attacker knows the constructed IP messages, he cannot convince a third party that he owns the model IP by creating fake secret keys;
\item \textit{Anti-overwriting}: An attacker cannot reconstruct an IP identifier to claim the model IP, even if he knows the protection algorithm.
\end{itemize}

\nop{The payload of Deep IP Identifiers is an important criterion for concealing adequate information about model copyright or user identification for Deep IP Protection. Moreover, an IP Identifier is generally harder to  crack or forge if it has a higher capacity, which could be likened to the fact that passwords or hash codes with more bits are harder to be cracked under the same mechanism. The meaning of capacity contains two aspects: the theoretical upper bound, and the bit number of the actual payload in IP Identifiers}

\subsubsection{Efficiency-of-IP (EoI)}  
The IP construction and verification in Deep IP Protection should be efficient for IP defenders but inefficient for IP attackers, which includes but is not limited to: 
\begin{itemize}[leftmargin=*]
	\setlength{\topsep}{0pt}
	\setlength{\itemsep}{0pt}
	\setlength{\parsep}{0pt}
	\setlength{\parskip}{0pt} 
\item \textit{Scalability}: Deep IP schemes should be employed in various scenarios. Firstly, it should consider diverse IP functions like copyright verification, user authorization/management, integrity verification, etc. Secondly, the constructed IP Identifiers should be adaptive to downstream-task versions of the original pre-trained model. Finally, it should cover various models and learning paradigms, like generated models, self-supervised learning, incremental learning, active learning, federated learning, etc. \nop{transfer learning, Moreover, not only well-trained parameters but also how to train a good model in these paradigms should be protected, and protecting Deep IP methods is also valuable to consider. Making Deep IP protection scalable to the above functions, tasks, and paradigms requires a lot of professional effort. }
\item \textit{Agility of Construction}: IP defenders can construct and verify IP Identifiers with minimal overhead, which enables a defender to easily create his own IP identifier. 
\item \textit{Slowness of Removal}: IP attackers have to remove or forge IP Identifiers at a much higher cost and performance loss, to degrade attackers' incentives to crack protected models.
\end{itemize}
\subsection{Main Challenges} \label{sec:challenges}
\textcolor{black}{Although taking inspiration from traditional media watermarking, 
Deep IP protection faces new challenges stemming from new host signals and new application scenarios~\cite{survey_barni2021dnn}.}  
Traditional data IP protection focuses on static data. Differently,
DNN models have dynamic outputs and are trained by large-scale datasets. Therefore, IP identifiers can be hidden in multiple optional parts of the original models, like any layer of the models, model inputs, model structures, model behaviors (\emph{e.g.}, model decision boundary, model outputs, or other model properties), activation feature maps, and training strategies. 
All aforementioned have not been explored, which produces not only more opportunities but also more challenges for Deep IP Protection. Firstly, With so many feasible ways, their theoretical boundary of protection capabilities, their robustness under the threats of endless IP attacks, their conflicting interaction and collaboration benefits, and more potential issues need to be further explored.  Secondly, the design of ideal IP protection techniques is complex and challenging due to the need to balance the conflicting requirements including 
\textcolor{black}{fidelity, capacity, and robustness}, as discussed in Section~\ref{sec:background_criteria}. However, the hard-to-be-interpreted characteristic of DNNs further increases the difficulty of designing an ideal strategy for Deep IP Protection. 
Moreover, in practical applications, deep IP protection needs to complete the whole workflow from IP construction, piracy discovery, and IP verification to giving effectual evidence, without affecting DNN models' utility. 
In this section, we identify several main challenges for deep IP protection.

\subsubsection{Model Fidelity}  
\textcolor{black}{Model fidelity measures the degree of performance retention from the original model to its protected versions, where the performance has multiple perspectives, including but not limited to prediction accuracy, inference efficiency, adversarial robustness, and potential risks. A good model fidelity means that the model performance on the original task should not be significantly degraded. The presence of the IP identification information naturally conflicts with the fidelity of the model. Due to the dynamic\footnote{A DNN is a large, parameterized complex function rather than a static object like an image in traditional image watermarking.}, complex and uninterpretable nature of the DNN models, inserting strong identification information into the DNN models can lead to underlying unusual prediction behaviors like prediction accuracy drop, unexpected security risks, and computational efficiency. }
\subsubsection{IP Robustness} 
The IP identifiers should be robust, which means to not only meet the requirements of model reuse of legitimate users like model lightweight~\cite{survey_mishra2023lightweight} and transfer learning~\cite{attack_zhuang2020transferlearning}, but also defend against various attack strategies of malicious attackers like IP detection, IP evasion, IP removal, and IP ambiguity, as shown in Section~\ref{sec:Attack_Overview}.
Each step of the whole pipeline for Deep IP protection, like IP construction, piracy discovery, IP verification, and adducing evidence, faces the threat of malicious attackers. It is challenging to defend against each attack strategy at each step. For complex and uninterpretable DNN models, it is also difficult to certify the robustness of IP Identifiers, \ie, to analyze or prove what degree of the robustness of a designed protection strategy to what kind of attack strategy. Moreover, IP robustness is often in conflict with IP capacity and model fidelity.  For example, model fingerprinting has optimal model fidelity due to no model modification, however, its QoI is generally much worse than model watermarking. How to determine and achieve a good tradeoff also needs to be explored in depth. 
\subsubsection{IP Capacity} 
It means the bit number of IP identifiers' valid information payload and the theoretical upper bound. Unlike traditional Data IP protection that can analyze the IP capacity by using the theories about signal processing and information theory, it is difficult to quantify the IP capacity for Deep IP Protection. It is because of the diversity of feasible host signals, dynamic behaviors of DNN models, etc, as discussed above. For example, IP capacity can be defined as the number of trigger samples, the length of embedded bit string in model weights or outputs, etc. The theoretical upper bound is hard to obtain due to the complex and uninterpretable nature of DNN models.

\subsubsection{Secure Pipelines}
Besides the aforementioned three challenges, Deep IP protection should solve a series of problems from design to deployment to forensics. In the face of endless threats in open environments, how to design a secure pipeline for Deep IP protection is also a challenging problem. Not only the IP construction and verification stages, 
but also more components, such as the method for discovering infringements, the mechanism for notarizing IP rights, and the secure IP management platforms, need to be carefully designed. This is a tedious task that requires long-term collaboration between academia and industry.

\section{The Frameworks of Deep IP Protection}
\label{sec:framework}
\begin{figure*}[htbp]
	\centering
	\includegraphics[width=.8\textwidth]{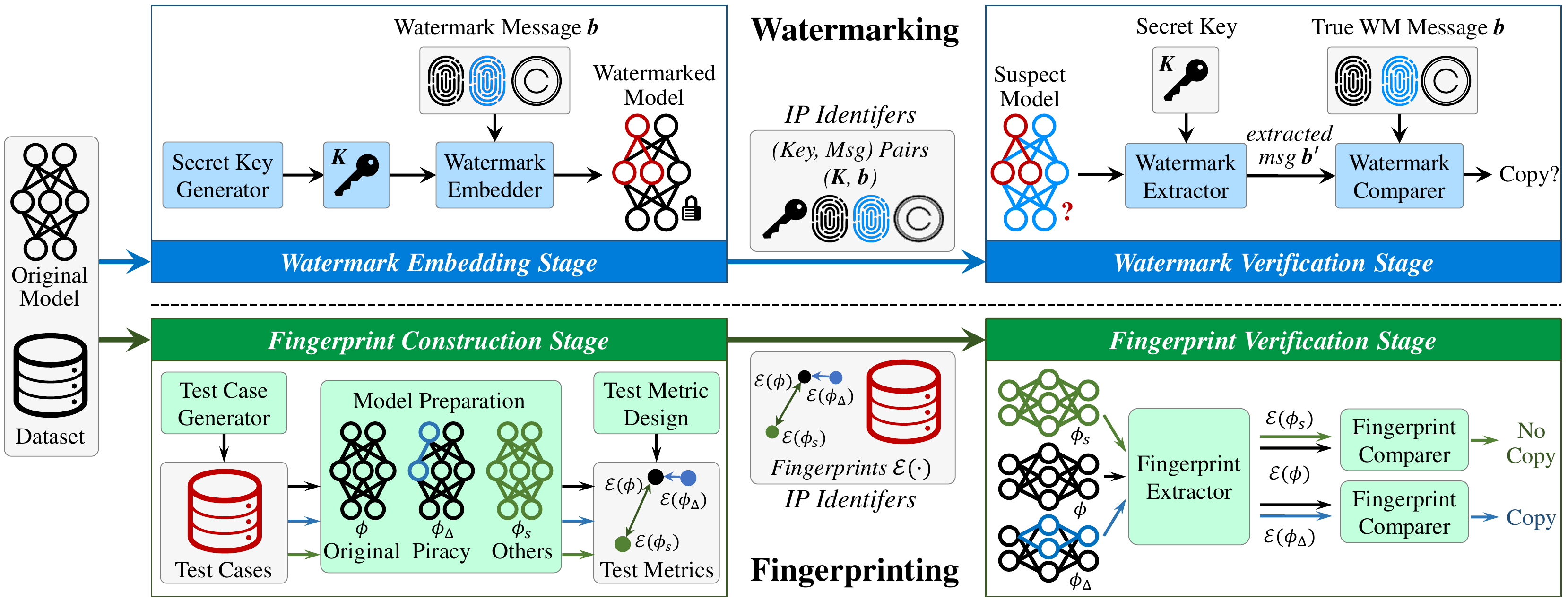}
 \vspace{-0.1in}
	\caption{\textcolor{black}{The Pipeline of Deep IP Protection consists of two stages: Deep IP identifier construction and Deep IP verification. Current Deep IP Protection methods could be divided into two main categories: Watermarking and Fingerprinting. Watermarking is an invasive solution to embed an IP identifier into the original model, generally through model finetuning or retraining. Fingerprinting, depicting the model characteristic with test cases and test metrics, is a noninvasive solution without any model modification. Its core idea is that the piracy version has a similar characteristic to the original model while the models trained independently are significantly different.}}
	\label{fig:pipeline}
 \vspace{-0.2in}
\end{figure*}

\textcolor{black}{The general pipeline of Deep IP Protection mainly consists of two stages: IP Identifier Construction and Verification, as illustrated in Fig.~\ref{fig:pipeline}. IP Identifier Construction aims at embedding or constructing some 
\textcolor{black}{unique IP Identifiers}, generally in the form of key-message pairs $(\Kcal, \bbm)$, as the IP information of the original model $\phi$. 
For instance, the keys $\Kcal$ could be preset numerical matrice or input samples, and the message $\bbm$ could be a bit string or model outputs. 
The IP Identifiers $(\Kcal, \bbm)$ will be retained by the model owners or the IP certification authorities. The owner can extract the message $\bbm$ from $\phi$ when and only when a correct key $\Kcal$ is used as input. During the verification stage, given a suspect model $\phi'$, the owner can extract the message $\bbm'$ using the key $\Kcal$ and compare the extracted message $\bbm'$ with $\bbm$. If the similarity between the two messages $sim(\bbm', \bbm)$ is upper than a threshold $\epsilon$, the suspect model $\phi'$ will be determined as a pirated copy of $\phi$. } 

 

\par
\subsection{\textcolor{black}{Deep IP Identifier Construction}}
\textcolor{black}{According to whether the original model needs to be modified, current deep IP protection methods can be categorized into two main classes: model watermarking and model fingerprinting.}



\textcolor{black}{\textbf{\textit{Model Watermarking}} is an invasive solution that embeds at least a detectable and unforgeable IP identifier, \ie a pair of a secret key and a watermark message $(\Kcal, \bbm)$, into the host DNN model $\phi$ to obtain a watermarked model $\phi_w$, as shown in Fig.~\ref{fig:pipeline}. Generally, DNN Watermarking mainly contains two components: the IP identifier generator $\Gcal_w$ and the watermark embedder $\Ecal_w$. The former outputs the IP identifier $(\Kcal, \bbm)$. Without loss of generality, it can be formulated as}
\vspace{-0.04in}
\begin{equation}
    (\Kcal, \bbm) = \Gcal_w(\Scal, [\phi, \Xbf_c, *]), 
    \vspace{-0.04in}
\end{equation}
where $\Scal$ represents the meta-data like the random seeds, the bit number of watermark messages, or the preset trigger patterns. Here, the secret key $\Kcal$ can be in many forms, such as a key matrix, trigger samples, a neural network, etc. The items in "$[]$" are optional inputs like the host model $\phi$, the candidate dataset $\Xbf_c$, etc. 
\textcolor{black}{
Then, the generated IP identifier $(\Kcal, \bbm)$ 
is input to a watermark embedder $\Wcal$ and embedded into the original model $\phi$ by finetuning or retraining the model $\phi$ on a dataset $(\Xbf,\Ybf)$,} which is formulated as
\begin{equation}\label{eq:wm_def}
\vspace{-0.06in}
\begin{split}
    \phi_w=\  &\Wcal(\phi, \Kcal, \bbm, \Xbf, \Ybf,[*]),\\
    \textrm{s.t.\ }\ \bbm \approx\  & \Ecal(\phi_w;\Kcal) \approx  \Ecal(\hat\phi_w;\hat\Kcal),\\
    \bbm \neq \ & \Ecal(\phi_u;\Kcal), \bbm \neq \Ecal(\phi_w;\Kcal_u), 
\end{split}
\vspace{-0.06in}
\end{equation}
where $\phi_w$ denotes the watermarked model and $\Ecal$ denotes the function for watermark extraction. The constraints in Equ.~(\ref{eq:wm_def}) means that, only given the watermarked model $\phi_w$ (or its modified copy $\hat\phi_w$), along with the correct key $\Kcal$ (or its perturbed version $\hat\Kcal$), the extraction function $\Ecal$ have the outputs similar to the IP message $bbm$. Otherwise, for irrelevant models $\phi_u$ or keys $\Kcal_u$, the extraction function $\Ecal$ outputs irrelevant messages.

\textcolor{black}{\textbf{\textit{DNN Fingerprinting}} is a noninvasive solution that extracts the representation of DNN models in a fingerprint space, called \textit{Model Fingerprints} $\Ecal(\phi;\Kcal)$, for comparing model similarities. Note that $\Kcal = (\Xbf_s,\Mcal)$ where $\Xbf_s$ and $\Mcal$ represent carefully-designed test cases and test metrics respectively.}
DNN Fingerprinting is mainly based on the assumption that, DNN models \textcolor{black}{individually trained by model developers without the full information about the training process} will be significantly different in at least one of the aspects including initial weights, training algorithms, or training datasets. These differences will be reflected in some model characteristics like model predictions, decision boundaries, adversarial robustness, neural activation paths, etc. 
Since the pirated model is a modified or approximated variant of the original model, its model characteristics will be highly similar to those of the original model. Once the fingerprints of the suspect and the original models are similar enough, \textit{e.g.,} $\textrm{sim}(\Ecal(\phi;\Kcal), \Ecal(\phi';\Kcal))> \epsilon$, we have high confidence to say that $\phi'$ is a piracy copy of $\phi$.
As shown in Fig.~\ref{fig:pipeline}, the process of fingerprint construction consists of Test Case Generation and Test Metric Design, which are used to create test cases $\Xbf_s$ and test measures $\Mcal$, respectively. This process can be formulated as the following optimization problem:
\vspace{-0.06in}
\begin{equation}
    \begin{split}
        \min_{\Kcal} \ \  &{\sum}_{\phi_u}\textrm{sim}(\Ecal(\phi;\Kcal), \Ecal(\phi_u;\Kcal)) \\ 
        - &\lambda{\sum}_{\phi_\Delta}\textrm{sim}(\Ecal(\phi;\Kcal), \Ecal(\phi_\Delta;\Kcal)),
    \end{split}
    \vspace{-0.06in}
\end{equation}
where $\phi_u$ represents an individually trained model and $\phi_\Delta$ represents a modified or approximated variant of the original model $\phi$. This optimization problem aims to minimize the fingerprint similarities between $\phi_\Delta$ and $\phi$ while maximizing that between $\phi_u$ and $\phi$.

\nop{
\textcolor{black}{
The former selects or generates a set of test cases (\emph{e.g.}, adversarial examples) as the secret key $\Kcal=\{k_i\}_{i=1}^n$, and the latter designs a series of model features $\Ccal=\{c_i\}_{i=1}^m$ such that the fingerprint similarities between the source model $\Mcal$ and its approximated version $\Mcal+\Delta$ can be maximized while those between $\Mcal$ and the independently trained model $\Mcal_s$ are minimized. Formally, the fingerprint construction process can be denoted as:}}

\textcolor{black}{In general, test cases $\Xbf_s$ (\emph{e.g.}, adversarial examples) are distributed near the decision boundary of the original model. The final (key,message) pairs $\{\Kcal,\Ecal(\phi;\Kcal)\}$ are retained by the model owners for IP Verification.} 
Since model fingerprinting does not modify the original model, anyone can construct their own fingerprints on the original model, and the owner is hard to provide his unique evidence (\emph{e.g.}, the original model that pirates have not) in the judgment process of IP conflicts. Therefore, this approach is not suitable for individual model developers to protect their own models, but for authoritative agencies for model audits and management to analyze model similarities. 


\par
\textcolor{black}{To better understand model fingerprinting and watermarking, we discuss their relationships from the following perspectives. }
\begin{itemize}[leftmargin=*]
	\setlength{\topsep}{0pt}
	\setlength{\itemsep}{0pt}
	\setlength{\parsep}{0pt}
	\setlength{\parskip}{0pt} 
\item \textit{Objective}: \textcolor{black}{Both techniques aim to construct unique and detectable IP identifiers to protect the IP of trained DNN models. 
\item \textit{IP Identifier Construction}: Whether watermarks or fingerprints, IP identifiers are represented either by properties of internal model components (\textit{e.g.}, model parameters) or by model behaviors on test samples (\textit{e.g.}, model outputs on adversarial examples). Watermarking  modifies or retrains the original model to create properties or behaviors as IP identifiers.  Differently, fingerprinting leverages the inherent information (\textit{i.e.}, decision boundary) of the trained models without model architecture modification. }
\item \textit{IP Verification}: \textcolor{black}{Both techniques follow similar verification pipelines, \ie extracting IP identifiers from suspect models to compare with constructed IP identifiers.}
\item \textit{Performance}: \textcolor{black}{Both techniques have similar challenges and evaluation metrics. Generally, the \textcolor{black}{QoI}
of watermarks is higher than that of fingerprints, but watermarks will damage the fidelity of the target model. }
\end{itemize}

\begin{figure*}[htbp]
	\centering
	\includegraphics[width=.8\textwidth]{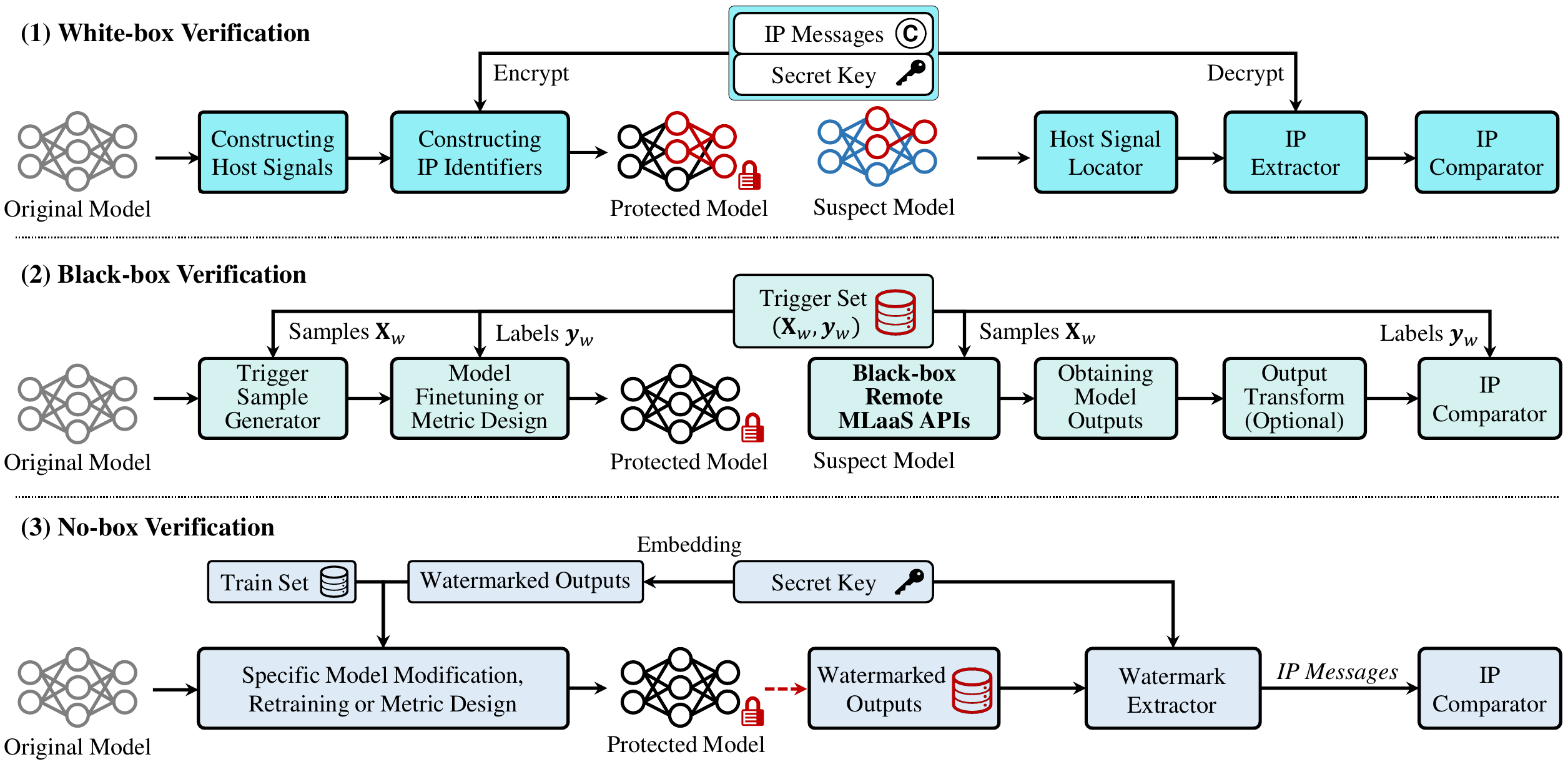}
 \vspace{-0.1in}
	\caption{The Illustration of Deep IP Verification. We group all the verification methods into three categories: white-box, black-box, and no-box verification. 
 White-box verification means that verifiers have full knowledge of the original models (e.g., structures and parameters), and extract the IP messages via three steps: host signal localization. IP extraction and IP comparison. In the black-box verification scenario, the verifier can only obtain model predictions such as soft probabilities or even hard labels for the queried samples. Thus, the verifier should extract IP messages only from the model predictions. Sometimes, the verifier has no access permission to the suspect model, neither the model parameters nor the inference APIs. However, the output results of the pirated models could be circulated on the Internet. In this case, IP verification can be performed if the output results contain IP messages in some forms like image or text watermarks, where a DNN is generally used as the extractor.} 
 \vspace{-0.2in}
	\label{fig:verification}
\end{figure*}

\subsection{Deep IP Verification}
\label{sec:framework_verification}
\par

The IP verification procedure consists of an Extractor $\Ecal$ and a Comparer $\Ccal$. Given a suspect model $\phi'$, the Extractor $\Ecal$ uses the secret key $\Kcal$ generated in the IP construction stage to extract the IP message $\bbm'$ of $\phi'$ corresponding to the IP construction method. Then, the Comparer $\Ccal$ returns the verification result $\Vcal(\phi';\Kcal, \bbm, \Ecal)$, with the formulation as 
\begin{equation}
\vspace{-0.06in}
\begin{split}
\Vcal(\phi';\Kcal, \bbm, \Ecal) &= \Ccal(\bbm, \bbm')\\     
                                &= \Ccal(\bbm, \Ecal(\phi';\Kcal)),
\end{split}
\vspace{-0.06in}
\end{equation}
where $\Vcal$ determines whether the suspect model $\phi'$ is a true pirated copy of the source model $\phi$ according to the similarity between the preset message $\bbm$ and the extracted message $\bbm'$.
The extracted message $\bbm'$ would be similar to $\bbm$, when and only when performing IP verification on the original model $\phi$ or its pirated version $\phi_\Delta$ with the correct key and the correct extractor. 

\nop{
\textcolor{black}{The IP verification procedure consists of an Extractor and a Comparer. Given a suspect model $\phi'$, the Extractor uses the secret key $\Kcal$ generated in the IP construction stage to extract the IP message $\bbm'$ of $\phi'$ via the verification algorithm $\Vcal$ corresponding to the IP construction method $\Ecal_w$:
\textcolor{red}{???extract key again? the key is given???? the (key,message) pair $\{\Kcal,\bbm'\}$} through the verification algorithm $\Vcal$ corresponding to the IP construction method $\Ecal_\Wcal$:
\[\bbm' = \Vcal(\Mcalhat, \Kcal, \Ecal_\Wcal)\],
\textcolor{red}{Any the optional inputs in the above formula???} where $\Ecal_\Wcal$ is used to locate the copyright signature from a DNN model. Then, through a comparer, the verifier determines whether the suspect model $\Mcalhat$ is a true pirated copy of the source model $\Mcal$ according to the similarity between the registered message $\bbm$ and the extracted message $\bbm'$. }
}

The methods for IP Verification have different requirements on access permission of the suspect model $\phi'$, and most of them apply one or a combination of the following three verification scenarios:


i) \textit{White-Box Verification}: The verifier requires full access to the suspect models including model parameters and structures in this verification scenario. The first step is to retrieve or recover the host signal, like a subset of raw/transformed model weights, model structures, dynamic outputs of some hidden layers, the bias and scales of normalization layers, or various designed metrics in fingerprinting-based schemes. 
The verifiers generally have strong verification abilities in this scenario, however, the white-box access of the suspect model is hard to obtain and usually requires the help of enforcement agencies with coercive force. Otherwise, although some techniques like model extraction and side channel attacks can help extract the suspect model, the extracted white-box model often cannot guarantee the existence and authenticity of the IP identifiers.  

ii) \textit{Black-Box Verification}: Here, the verifier only requires the predictions of the suspect model, such as soft probability or hard labels of classification/recognition models, the predicted values of regressive models, or the generated outputs of generative models. It enables remote verification on MLaaS APIs, where the service providers only open the APIs access to DNN models rather than the whole model. After receiving the queried samples from MLaaS users, the service providers perform model inference on remote servers and return the inference results. 
The verifier uploads some queried samples $\Xbf_q$ as the secret key $\Kcal$, and obtains a series of query-result pairs $\{\Xbf_q, \Ybf'_q\}$ to depict the model behaviors as the IP message $\bbm'$ \footnote{In some works, the IP message $\bbm'$ need to be extracted via some extra processing on the query-result pairs $\{\Xbf_q, \Ybf'_q\}$.}. 
Although deploying MLaaS APIs on remote servers can provide some degree of protection for the commercial benefits of DNN models, infringement acts, like the MLaaS API using a piracy model illegally obtained by model extraction, become more difficult to be detected. First, verifiers are required to make judgments solely based on model predictions from a finite number of finite queried samples. Second, the process of model inference is entirely controlled by service providers. This provides infringers with more means to attack the stage of IP verification beyond model modifying/retraining, like input detection/processing and ensemble attacks. Moreover, due to commonly-paid MLaaS APIs, the verification cost cannot be ignored.

iii) \textit{No-Box Verification}: Unlike black-box verification that requires pre-defined trigger samples or test cases, no-box verification allows the verifier to extract IP messages from model outputs taking normal task samples as inputs. Therefore, this scenario enables remote IP verification without explicit secret keys or even model access permission, as the verifier can utilize MLaaS APIs' generated outputs spread throughout the Web. In a way, no-box verification can be regarded as an extension application of traditional data IP protection in DNN models, which could be designed based on traditional data steganography. The verification function $\Vcal$ is often performed by an extractor DNN model jointly trained with the original model. In general, this verification scenario is applied to generative models whose prediction has enough capacity to hide effective IP messages. The typical applications include various tasks of image processing and natural language processing, such as semantic segmentation~\cite{survey_garcia2017reviewsemseg}, image super-resolution~\cite{survey_wang2020superresolution}, style transfer~\cite{survey_jing2019styletransfer}, machine translation~\cite{survey_ranathunga2023machinetranslation}, etc. 

\section{Main Threats of Deep IPs} \label{sec:Attack}
\subsection{Overview of Threats} \label{sec:Attack_Overview}
\begin{table*} [ht]
\centering
  \caption{Threats to Deep IP Protection }
  \label{tab:threats} 
    \vspace{-0.1in}
    \scriptsize
    \setlength\tabcolsep{4pt}
    \begin{tabular}{cc|c|c|l}
    \toprule
    \textbf{Types} &\textbf{Methods} & \textbf{Data Demands} & \textbf{Model Demands} & \textbf{Literature \& Description}\\
    \midrule
    \multirow{7}{*}{\makecell{IP Detection\\ \& Evasion}} & \makecell{Detecting White-box\\ Watermarks} 
                                          &$\geq$ Poor  & \makecell{White-box\\ Attackers} &\makecell[l]{Detecting the abnormal distribution of model parameters~\cite{attack_wang2019attacks,white_wang2021riga}\\
                                          Detecting the abnormal distribution of hidden-layer activations~\cite{attack_hitaj2019intputdetection}\\
                                          Detect non-essential or potentially-watermarked components in models
                                  }\\[2pt] 
    \cmidrule(lr){2-5}
                                  & \makecell{Detecting Black-box\\ Verification} 
                                          &$\geq$ Poor & \makecell{White-box\\ Attackers} & \makecell[l]{Detecting Trigger Inputs: Detector~\cite{attack_hitaj2019intputdetection}, GAN~\cite{attack_wang2021GANdetect}, JSD~\cite{black_namba2019robust}\\
                                          Detecting Trigger Channels or Neurons: Neural Laundering~\cite{attack_aiken20_neurallaunder}
                                  }\\[2pt]
    \cmidrule(lr){2-5}
       & \makecell{Input/Output\\Processing}    
                                          &$\geq$ Poor & \makecell{White-box\\ Attackers} &
                                                    \multicolumn{1}{m{10.2cm}}{                                 
                                          AE-based Query Modification~\cite{black_namba2019robust}, 
                                          Input Reconstruction~\cite{attack_lin2019inputreconsturction}, 
                                          Input Quantization~\cite{attack_lin2019defensive}, 
                                          Input Smoothing~\cite{attack_xu2017featuresqueezing}, 
                                          Input Noising~\cite{attack_zantedeschi2017inputnoising}, 
                                          Input Flipping~\cite{sok_lukas2021sok}, PST~\cite{attack_guo2020PST}, 
                                          Feature Squeezing~\cite{attack_xu2017featuresqueezing}, 
                                          JPEG Compression~\cite{attack_dziugaite2016jpgcompress}, 
                                          Ensemble attack~\cite{attack_hitaj2019intputdetection}, 
                                          AdvNP~\cite{attack_wang2022advnp}
                                  }\\
    
    \midrule
     \multirow{10}{*}{IP Removal}         & \multirow{4}{*}{\makecell{Model\\Modification}}
                                          &$\geq$ Poor & \makecell{White-box\\ Attackers} & 
                                           \multicolumn{1}{m{10.2cm}}{Fine-Tuning (RTLL, RTAL, FTLL, FTAL)~\cite{white_uchida2017embedding,attack_chen2019leveraging}, 
                                          Reversely Fine-tuning~\cite{attack_wang2021GANdetect}, 
                                          PST-based Fine-tuning~\cite{attack_guo2020PST}, 
                                          REFIT~\cite{attack_chen2021refit}, WILD~\cite{attack_liu2021WILD,attack_zhong2020randomerasing}, 
                                          Weight Pruning~\cite{attack_blalock2020pruning}, 
                                          Fine-Pruning~\cite{attack_liu2018finepruning}, 
                                          Label Smoothing~\cite{attack_szegedy2016labelsmoothing}, 
                                Weight Regularization~\cite{attack_shafieinejad2021retrain_regular}, FLAME~\cite{attack_Thien2022FLAME}, 
                                          Weight Shifting~\cite{sok_lukas2021sok}, 
                                          Feature Permutation~\cite{sok_lukas2021sok}, 
                                          ReMoS~\cite{attack_zhang2022remos}, 
                                          Neural Cleanse~\cite{attack_wang2019neuralcleanse}, 
                                          Neural Laundering~\cite{attack_aiken20_neurallaunder},  
                                          Weight Quantization~\cite{attack_hubara2017quantized}, 
                                          Adversarial Training~\cite{attack_madry2018adversarialtraining}, 
                                           Attention Distraction~\cite{attack_zhong2022attention},  
                                           Minimal Modification~\cite{attack_goldberger2020minimal},  
                                           Referenced Subspace Attention~\cite{attack_xue2022removing}, 
                                           Neural Structural Obfuscation~\cite{attack_yan2023NeuralStructuralObfuscation} 
                                        }\\
                                      \specialrule{0em}{1pt}{1pt}
                    \cline{3-5}              
                        \specialrule{0em}{1pt}{1pt}
                                &  & Strong &  \makecell{White-box\\ Attackers} & Adaptive Model Modification~\cite{black_lee2022evaluating} \\
                                  
    \cmidrule(lr){2-5}
                                  & \multirow{3}{*}{\makecell{Model\\Extraction}} 
                                          & $\geq$ Medium &\makecell{Black-box\\\& White-box} & \multicolumn{1}{m{10.2cm}}{Retraining w/ hard labels~\cite{attack_shafieinejad2021retrain_regular}, 
                                          Smooth Retraining~\cite{sok_lukas2021sok}, 
                                          Cross-Architecture Retraining~\cite{sok_lukas2021sok}, 
                                          Adversarial Training from scratch~\cite{attack_madry2018adversarialtraining}, 
                                          Knockoff Nets~\cite{attack_orekondy2019knockoff}, Transfer Learning~\cite{attack_zhuang2020transferlearning}, Distillation~\cite{attack_yang2019distillation}, Ensemble Distillation~\cite{black_charette2022cosine}
                                  }\\
                                      \specialrule{0em}{1pt}{1pt}
                        \cline{3-5}
                            \specialrule{0em}{1pt}{1pt}
                                &  & Strong &  \makecell{Black-box\\ Attackers} & Adaptive Model Extraction~\cite{black_lee2022evaluating} \\
    \midrule
    \multirow{4}{*}{IP Ambiguity} & \makecell{Reverse\\ Construction}   
                                          &$\geq$ Poor & \makecell{White-box\\ Attackers} & \makecell[l]{Searching a secret key of a given IP signature~\cite{passport_fan2019rethinking},  internal attacks ~\cite{passport_fan2019rethinking}}\\ 
    \cmidrule(lr){2-5}  
                                  & \makecell{Reconstruction}          
                                          &$\geq$ Poor & \makecell{White-box\\ Attackers} & \makecell[l]{Overwriting a new key-signature pair~\cite{attack_wang2019attacks}}\\
  \bottomrule
\end{tabular}
\vspace{-0.15in}
\end{table*}
This section provides an overview of Deep IP threats from attackers' capability on data and models. In Table~\ref{tab:threats}, we summarize some existing IP attack methods.
\subsubsection{Data-level Abilities of Attackers} \label{sec:Attack_Overview_Data}
We divide attackers' data capability into different levels from the perspective of data quantity, data distribution, and labels; then, describe their purposes. 
\par
\begin{itemize}[leftmargin=*]
	\setlength{\topsep}{0pt}
	\setlength{\itemsep}{0pt}
	\setlength{\parsep}{0pt}
	\setlength{\parskip}{0pt} 
\item  Strong attackers have enough labeled in-distributed data to train a good model, 
and their motivations for stealing models could be discovering the weaknesses of stolen models to implement adversarial or backdoor attacks, avoiding the computation-heavy model training process, combining other models to improve the performance of their models, or using ensemble attacks to crack IP identifiers of other protected models. Moreover, IP protection mechanisms can be easily defeated if the algorithm is leaked to attackers in some ways, such as bribing insiders or mining the leakages of management mechanisms or software platforms. We consider such attackers as strong attackers.
\item Medium attackers have a sufficient amount of in-distributed data without enough labels. 
These attackers require the inference results of the protected model to train a DNN model. 
\item Weak attackers do not possess enough data for individual model training or extraction, and they can only crack other white-box models like model modification. 
\item Poor attackers only have out-of-distribution data. Thus, not only is model extraction prohibited, but also model modification should be strictly limited.
\end{itemize}

\subsubsection{Model-Level Abilities of Attackers} \label{sec:Attack_Overview_Model}
We grade attackers' model-level abilities into four levels (level-0 is the strongest):
\begin{itemize}[leftmargin=*]
	\setlength{\topsep}{0pt}
	\setlength{\itemsep}{0pt}
	\setlength{\parsep}{0pt}
	\setlength{\parskip}{0pt} 
\item
\textit{Level-0:} Attackers have full access (parameters and structures) to multiple models trained on similar-domain datasets. 
\item
\textit{Level-1:} Attackers have full access to only one protected model. 
\item
\textit{Level-2:} Outputs from some hidden layers can be obtained.
\item
\textit{Level-3:} Only soft model outputs can be obtained.
\item
\textit{Level-4:} Only hard labels can be obtained.
\end{itemize}
\par
\textbf{White-box attackers} (level-0 and level-1) can directly detect, remove, or forge IP identifiers of the protected model, where the level-0 attackers can perform ensemble attacks like ensemble distillation and ensemble evasion. \textbf{Black-box attackers} (other levels) need to first extract the protected model by model stealing attacks. 
\subsubsection{The Categorization of Threats} 
According to the step being attacked in Deep IP pipelines, the threats of Deep IP can be categorized into the following three types:
\begin{itemize}[leftmargin=*]
	\setlength{\topsep}{0pt}
	\setlength{\itemsep}{0pt}
	\setlength{\parsep}{0pt}
	\setlength{\parskip}{0pt} 
\item
\textit{IP Detection \& Evasion that attacks the verification process}. 
In white-box verification scenarios, the verification process is controlled by the verifiers, which could be considered secure. Here, IP detection is generally used to aid in subsequent IP Removal attempts, like detecting the specific distributions of watermarked parameters~\cite{attack_wang2019attacks,white_wang2021riga,attack_hitaj2019intputdetection}. 
In black-box scenarios, verifiers upload queried samples to suspect MLaaS APIs and receive the returned results. This process is controlled by MLaaS providers, and MLaaS providers using pirated models will try to attack this process, using input/output processing, secret key detection then output perturbation, ensemble attacks, etc. In general, it has two ways of attacks: invalidating secret keys (queried samples)~\cite{black_namba2019robust,attack_lin2019inputreconsturction,attack_lin2019defensive,attack_xu2017featuresqueezing,sok_lukas2021sok,attack_guo2020PST,attack_dziugaite2016jpgcompress,attack_wang2022advnp,attack_zantedeschi2017inputnoising}, or erasing the IP messages hidden in predictions returned to verifiers~\cite{attack_hitaj2019intputdetection,attack_wang2021GANdetect,black_namba2019robust,attack_aiken20_neurallaunder}. In general, covert IP identifier designs can be used to combat this threat~\cite{white_wang2021riga, black_li2019prove}. 
\item
\textit{IP Removal that destroys constructed IP Identifiers}. Malicious attackers can destroy the constructed IP Identifiers and create a non-IP pirated model. It could be realized by model modification~\cite{white_uchida2017embedding,attack_chen2019leveraging,attack_wang2021GANdetect,attack_guo2020PST,attack_chen2021refit,attack_liu2021WILD,attack_zhong2020randomerasing,attack_blalock2020pruning,attack_liu2018finepruning,attack_szegedy2016labelsmoothing,attack_shafieinejad2021retrain_regular,attack_Thien2022FLAME,sok_lukas2021sok,attack_zhang2022remos,attack_wang2019neuralcleanse,attack_aiken20_neurallaunder,attack_hubara2017quantized,attack_madry2018adversarialtraining,attack_zhong2022attention,attack_goldberger2020minimal,attack_xue2022removing,attack_yan2023NeuralStructuralObfuscation} or extraction~\cite{black_lee2022evaluating, attack_shafieinejad2021retrain_regular,attack_orekondy2019knockoff, attack_zhuang2020transferlearning,attack_yang2019distillation,black_charette2022cosine}. Model modification is to adjust model parameters or structures, such as model finetuning, fine-pruning, regularization, neural cleanse, etc. Here, attackers require white-box protected models. Model extraction is to learn a model copy from model predictions or properties of protected models, such as query-based stealing attacks (\emph{e.g.}, knowledge distillation) and side-channel attacks~\cite{survey_oliynyk2022summer}. It can be realized in both black and white box scenarios. Generally, model extraction has a stronger performance for IP removal but requires larger datasets. Many efforts have been made to design IP identifiers that can resist removal attacks, however, there is still no complete solution~\cite{black_lee2022evaluating,sok_lukas2021sok,attack_yan2023NeuralStructuralObfuscation}.
\item
\textit{IP Ambiguity that degrades the confidence of verification results}. After IP Verification, verifiers need to provide valid evidence to claim model IPs. However, given a protected model, attackers can re-embed or reversely construct their own IP Identifiers~\cite{passport_fan2019rethinking, attack_wang2019attacks}, where the former requires model finetuning but the latter does not. For example, against backdoor-based model watermarking, attackers can generate adversarial samples to mimic trigger samples or fine-tune the protected model with attackers' trigger sets. The same is true for parameter-based watermarking. Fan \myetal propose Passport~\cite{passport_fan2019rethinking} to defend against IP ambiguity attacks: unauthorized users without correct passports cannot get correct inference results or modify the protected model. However, unruly users may undermine this protection by spreading passports. 
\end{itemize}
\subsection{IP Detection and Evasion}
Deep IP verification can be detected or evaded by adversaries. We overview such threats in the following three categories:
\subsubsection{Detecting White-box Watermarks}
Generally, it will change model parameters more or less when embedding watermarks.
For example, the probability distribution or the frequency spectrum of the watermarked parameters would be different from that of normal parameters. Similarly, watermarking based on neither hidden-layer activation nor extra watermark components can ensure covertness.
Therefore, adversaries can leverage these abnormal attributes to discover embedded watermarks, like detecting the abnormal distribution of model parameters or hidden-layer activations and detecting potentially-watermarked model components. 
\subsubsection{Detecting Remote Black-box Verification}
Remote Black-box Verification is often realized by uploading some queried samples to suspect MLaaS APIs and then analyzing the returned results. Here, MLaaS providers can detect whether queried samples are trigger samples (or test cases), and then perform specific operations, like denying services or label modification, to circumvent remote verification.   
Such detection can be done by analyzing the likelihood of queried samples. Generally, watermarking schemes select out-of-distribution (OOD) or near-OOD samples as trigger samples in order to easily change the predicted labels with minor model modifications, while fingerprinting schemes select the adversarial samples close to the decision boundary for analyzing the model similarity. However, these selected samples are significantly different from task samples, which can be detected by verifiers but also exposed to adversaries. Moreover, the neurons activated by the trigger samples would also be significantly different from the normal task samples, which can be leveraged by adversaries to avoid IP verification. 
\subsubsection{Input/Output Pre/Post-Processing}
Since verifiers have no knowledge about the inner process of suspect MLaaS APIs, MLaaS providers can arbitrarily modify queried samples or inference results. Besides traditional lossy data compression techniques, MLaaS providers using pirated models can leverage adversarial or backdoor defense techniques at the sample level to evade remote IP verification. Moreover, if an adversary has multiple models with different IP Identifiers, the adversary can aggregate the inference results from multiple models into an inference result without any IP messages. A simple voting method can realize this process. 
\subsection{IP Removal}
As shown in Table~\ref{tab:threats}, a large number of methods have been proposed to remove the IP identifiers of protected models. 
 \subsubsection{Model Modification}
 It refers to removing the IP identifiers of protected models by modifying model parameters or structures. This type of IP removal attacks can be realized using a few data samples. Besides untargeted methods like model fine-tuning, pruning, and their variants, some IP-aware modification methods have been proposed. For example, adversaries can try to find and then modify the parameters or neurons that are sensitive to IP identifiers but unrelated to original tasks. Moreover, if an adversary has the algorithmic information about the adopted Deep IP scheme, the adversary can simulate the trigger samples by generative models or preset rules and give these generated samples the correct labels to fine-tune the protected model. 
  \subsubsection{Model Extraction}
 If an adversary has enough in-distribution data samples, the adversary can obtain the inference results of the protected models on these samples, and then use these results as the soft/hard labels to retrain a pirated model from re-initialized model parameters or even another model structures. Model extraction can effectively remove IP identifiers and maintain model performance to some degree. Firstly, inner-component-embedded watermarks (see details in Section~\ref{sec:Protect_Invasive_Parameter}) are generally vulnerable to model extraction since model extraction does not depend on model internals and can be even done across model structures. Then, specific output patterns on trigger samples are also difficult to transfer to extracted models, since trigger samples are usually weakly coupled to task samples. In the same way as model modification, active attacks can also be performed in model extraction if the adversary has the algorithmic information about the adopted Deep IP scheme. Moreover, if an adversary has multiple models for similar tasks, the adversary can leverage the inference results from these models to learn a clean model, by which the model accuracy could be even improved. 
\subsection{IP Ambiguity} \label{sec:attack_ambiguity}
Adversaries can also fabricate copyright conflicts, such as creating their own IP identifiers by optimizing secret keys or re-constructing their own IP identifiers of the protected models.
 \subsubsection{Reverse Construction} 
 Given a protected model, an adversary can set some fake IP messages and adjust the secret keys to make the verification step output the fake IP messages. In this way, the key-message pair are reversely constructed by the adversary, thereby confusing the IP identifier of the model owner to create copyright conflicts. This attack can be realized even though the protected model is not allowed to be modified. 
 For example, parameter-based watermarking is generally done by fine-tuning the original or transformed model parameters using a watermark regularizer loss. However, fine-tuning the key matrix using the same regularizer loss has a similar effect, which can be done by gradient descent on the key matrix. For trigger-based watermarking, the techniques for adversarial sample generation can be used to forge trigger samples. 
  \subsubsection{IP Reconstruction}
 An adversary can use its own deep IP methods to construct an IP identifier on the original model. In this way, IP identifiers separately belonging to the owner and the adversary will have copyright conflicts. 
\par \noindent
\textbf{Remark 1}: For some Deep IP schemes against IP Ambiguity, even if it is difficult to reconstruct IP identifiers using the same algorithms, however, adversaries can leverage or design their own algorithms to finish IP Ambiguity attacks. In this way, if there is no unified standard in the industry, \ie adopting a unified algorithm to achieve Deep IP protection, or the models are not allowed to be modified, current IP Ambiguity attacks are still effective and hard to defend. However, since current Deep IP schemes still rely on algorithmic concealments, it is still infeasible to form a public industry standard. 
\par \noindent
\textbf{Remark 2}: When dealing with copyright conflicts, since model thieves usually do not have the original model, \ie the clean model without any IP identifier, the judicial organization can require the party to provide the original model to prove the ownership. However, the original model is not modified for fingerprinting schemes. Even for watermarking schemes, due to the transferability of adversarial samples, the clean model may output the prediction results hiding fake IP messages on fake trigger samples provided by the adversary. Moreover, an adversary may have multiple model sources for similar tasks, and may fool the judiciary via the ensemble version of these models, even if the adversary has no clean model.

\section{Deep IP Protection: Invasive Solutions}\label{sec:Protect_Invasive}
\begin{sidewaystable*}
\tiny
\vspace{-0.1in}
\begin{center}
  \caption{The Comparison of Representative Methods for DNN Watermarking}
  \label{tab:watermarking}
    \vspace{-0.1in}
    \setlength\tabcolsep{1pt}
  \begin{tabular}{|c|c|c|c|c|c|c|c|c|c|c|c|}
    \toprule 
    \multicolumn{2}{|c|}{\textbf{Methods}}
    & \textbf{Year} 
    & \textbf{Secret Key} 
    & \textbf{Watermark Messages} 
    & \textbf{Host Signals} 
    & \textbf{Embedding Paradigms \& Scenarios} 
    & \makecell{\bf Verification} 
    & \textbf{Target Functions}
    & \textbf{Target Networks/Tasks}
    & \textbf{Data Scale \& Type}
    & \textbf{Highlights}\\ 
    \midrule
    \multirow{26}{*}{\rotatebox{90}{\bf Inner-Component-Embedded Watermarks (WM)}}
    &Uchida \myetal~\cite{white_uchida2017embedding} & 2017 & \makecell{Secret Matrix} &\makecell{a unified bit string} & \makecell{original model weights} & \makecell{finetune w/ CE regularizer loss; Directly} & White-box &Copyright Verification & classification & small image datasets & \makecell{The first scheme for Deep IP Protection}\\
    &Deepmarks~\cite{white_chen2018deepmarks}      & 2018 & \makecell{Secret Matrix} & \makecell{user-aware bit strings} & \makecell{original model weights} & \makecell{finetune w/ MSE regularizer loss; -} & White-box &User Management & classification & small image datasets & \makecell{User Identification w/ Anti-collusion Coding}\\
    \\[-2ex]\cline{2-12}\\[-1.5ex]
    &Liu \myetal~\cite{white_liu2021residuals}     & 2021 & \makecell{RSA Public Key} &\makecell{a unified bit string} & \makecell{weights' greedy residuals} & \makecell{finetune w/ sign regularizer loss; -} & White-box &Copyright Verification & \multirow{2}{*}{\makecell{CNN \& RNN; \\[-1pt] classification}} & small; images\&texts & \makecell{robust WM w/o explicit ownership indicators}\\
    
    &RIGA~\cite{white_wang2021riga}                & 2019 & \makecell{Secret Network} &\makecell{a unified bit string} & \makecell{DNN-transformed weights} & \makecell{co-train target, embedder and hider DNNs; -} & White-box & Copyright Verification &  & small/med image/text & \makecell{Covert watermarks via a WM hider network}\\ 
    
    
    &\makecell{Frequency WM \cite{white_feng2020watermarking}\cite{white_kuribayashi2021white}} & 2020 & \makecell{Random seed  for weights} &\makecell{a unified bit string} & \makecell{spectrum of weights} & \makecell{binarize spectrum \& finetune non-watermark weights; -} & White-box & Copyright Verification & classification & small image datasets & \makecell{embed covert WM into the spectrum of weights}\\ 
    \\[-2ex]\cline{2-12}\\[-1.5ex]

    &Guan \myetal~\cite{white_guan2020reversible}  & 2019 & \makecell{WM Indices} &\makecell{a unified bit string} & \makecell{Model weights} & \makecell{embed WM by Histogram Shift; -} & White-box &Integrity Verification & - & - & \makecell{embed reversible watermarks to verify integrity}\\
    
    & NeuNAC~\cite{white_botta2021neunac} & 2021 & \makecell{Secret Vector Set} &\makecell{a unified bit string} & \makecell{KLT basics of weights} & \makecell{embed WM into LSBs of weights; -} & White-box &Integrity Verification & - & - & \makecell{embed fragile watermarks to verify integrity}\\
    
    & DeepAttest~\cite{whtie_chen2019deepattest} & 2019 & \makecell{Secret Matrix} &\makecell{user-aware bit strings} & \makecell{Model weights} & \makecell{finetune w/ MSE regularizer loss;-} & White-box &\makecell{On-device Functionality} & classification & small image datasets & \makecell{Software-Hardware Co-design in TEE}\\
    
    \\[-2ex]\cline{2-12}\\[-1.5ex]
    & Deepsigns-W~\cite{white_rouhani2018deepsigns} & 2018 & \makecell{Normal Samples \\[-1pt] \& Secret Matrix} &\makecell{a unified bit string} & \makecell{hidden-lay activation} & \makecell{ finetune w/ CE \& MSE regularizer\\[-1pt] in activation isolation constraints; -} & White-box &Copyright Verification & classification & small image datasets & \makecell{embed WM into the exception of hidden-layer \\[-1pt] activations of some target classes} \\
    
    & MOVE-Gradient~\cite{white_li2022defending}  & 2022 & \makecell{Style-transferred Samples} &\makecell{meta-classifer} & \makecell{Model gradients} & \makecell{train the model on styled samples w/ CE loss;-} & White-box &Copyright Verification & classification & small\&medium; images & \makecell{external features as WM verified by a meta-classifer} \\ 
    
    & Radioactive Data~\cite{data_sablayrolles2020radioactive}& 2020 & \makecell{Perturbed Samples} &\makecell{a shift vector} & \makecell{feature extractor} & \makecell{perturb dataset to create an isotropic shift\\[-1pt] on target feature extractor; -} & White\&Black & \makecell{Dataset Protection \&\\[-1pt] Copyright Verification} & classification & medium; ImageNet & \makecell{WM can be embedded from data to feature space} \\

    \\[-2ex]\cline{2-12}\\[-1.5ex]
    & Struct4WM \cite{white_zhao2021structural} \& \cite{new_xie2021deepmark} & 2021 & \makecell{Secret Vector} &\makecell{a unified bit string} & \makecell{Model structures} & \makecell{channel/weight-wise model pruning; -} & White-box &Copyright Verification & classification &small image dataset & \makecell{Embed WM into model structures}\\
    
    & Chen \myetal~\cite{lottery_chen2021you}  & 2021 & \makecell{Key Mask Matrix} &\makecell{a QR code} & \makecell{Model structures} & \makecell
    {embed a QR code to the squeezed mask; -} & White-box & \makecell{Copyright Verification\\[-1pt] \& Access Control} & classification &small image datasets & \makecell{the first to protect winning lottery tickets}\\
    
    & Watermark4NAS\cite{gray_lou2021meets} & 2021 & \multicolumn{2}{c|}{\makecell{fixed neural connections \& hyperparameters}} & \makecell{Model structures} & \makecell
    {search model structures constrained \\[-1pt] by watermarked neural connections; -} & White\&Black &Copyright Verification & CNN \& RNN &small; images\&texts & \makecell{leverage watermarks to protect\\[-1pt] high-performance structures from NAS}\\
    \\[-2ex]\cline{2-12}\\[-1.5ex]
    
    & Passport-V1\&2~\cite{passport_fan2019rethinking, passport_fan2021deepip} & 2019 & \makecell{Passport Matrix} &\makecell{a unified bit string} & 
    \multirow{2}{*}{\makecell{Scale \& Shift of\\[-1pt] normalization layer}} & \multirow{3}{*}{\makecell{multi-task learning across\\[-1pt] passport-aware and passport-free model\\[-1pt]w/ sign regularizer loss \& task loss; -}} & White-box &\multirow{3}{*}{\makecell{Copyright Verification\\[-1pt] \& Access Control}} & classification & small\&medium; images & \makecell{Add a passport layer after each conv layer}\\
    
    & \makecell{Passport-Norm~\cite{passport_zhang2020passport}}&2020& \makecell{Passport Matrix} &\makecell{a  unified bit string} & &  & White-box & & classification &images \& 3D points & \makecell{Add a passport branch for verification}\\
    
    & RNN-IPR~\cite{passport_lim2022rnn}     &2022  & \makecell{Passport Matrix} &\makecell{a unified bit string} & \makecell{RNN Hidden State} &  & White-box &  & RNN; Image Caption &small\&medium; images & \makecell{extend watermarks into RNNs}\\
    
    & HufuNet~\cite{new_lv2023robustness}  & 2021 & \makecell{Decoder DNN\\[-1pt] \& HMAC key} &\makecell{an encoder DNN} & \makecell{Model weights} & \makecell{Iteratively training the encoder \& target \\[-1pt] DNN w/ a robustness-assured loss; -} & White-box &Copyright Verification & \makecell{CNN \& RNN;\\[-1pt] classification} &small; texts\&images & \makecell{embed a subnetwork as watermarks\\[-1pt] verified by sample reconstruction}\\
    \\[-2ex]\cline{1-12}\\[-1.5ex]
    \multirow{34}{*}{\rotatebox{90}{\bf Trigger-Injected  Watermarks (WM)}}
    & Adi \myetal~\cite{black_adi2018turning}       & 2018 & \makecell{Unrelated Samples} &\makecell{Random Hard Labels} & \makecell{Model outputs} & \makecell{finetune or retrain w/ CE loss; -} & Black-box & Copyright Verification & classification &small image datasets & \makecell {leverage backdoor attacks for Deep IP protection}\\
    
    & Deepsigns-B~\cite{white_rouhani2018deepsigns}  & 2018  & \makecell{Near OOD Samples} &\makecell{Target Hard Labels} & \makecell{Model outputs} & \makecell{finetune or retrain w/ CE loss; -} & Black-box & Copyright Verification & classification &small image datasets & \makecell{select trigger samples from rarely explored regions}\\ 
    
    & \makecell{Zhang \myetal~\cite{black_zhang2018protecting}} & 2018 & \makecell{Embedded, Noised, \\[-2pt] or Unrelated Samples}   & \makecell{Target Hard Labels} & \makecell{Model outputs} & \makecell{finetune or retrain w/ CE loss; -} & Black-box & Copyright Verification & classification &small image datasets & \makecell{design various triggers for a proper robustness}\\
    \\[-2ex]\cline{2-12}\\[-1.5ex]
    
    & Guo \myetal~\cite{black_guo2018watermarking} & 2018 & \makecell{Samples w/ Message} &\makecell{Target Hard Labels} & \makecell{Model outputs} & \makecell{finetune or retrain w/ CE loss; -} & Black-box & Copyright Verification & classification &small image datasets & \makecell{Inject an additive pattern with user message}\\ 

    & Maung \myetal~\cite{black_maung2021piracy} & 2018 & \makecell{Block-wise Transform} &\makecell{Target Hard Labels} & \makecell{Model outputs} & \makecell{finetune or retrain w/ CE loss; -} & Black-box & Copyright Verification & classification &small image datasets & \makecell{with no need of predefined  trigger samples}\\

    & Zhu \myetal \cite{black_zhu2020secure} & 2018 & \makecell{Trigger Sample Chain} &\makecell{Target Hard Labels} & \makecell{Model outputs} & \makecell{finetune or retrain w/ CE loss; -} & Black-box & Copyright Verification & classification &small image datasets & \makecell{use hashing to get a trigger chain against IP forge}\\ 

    & Few Weights~\cite{black_lao2022fewweights} & 2022 & \makecell{Samples w\ highly \\[-1pt]uncertain outputs} &\makecell{Target Hard Labels} & \makecell{Model outputs} & \makecell{finetune a few WM-sensitive\\[-1pt] and task-unrelated weights; -} & Black-box & Copyright Verification & \makecell{CNN\& Transformer\\[-1pt] classification} &small\&medium; images & \makecell{realize a proper tradeoff between\\[-1pt] model fidelity \& WM robustness}\\ 
    
    \\[-2ex]\cline{2-12}\\[-1.5ex]
    & \makecell{Frontier-Stitching~\cite{black_le2020adversarial}}  & 2017 & \makecell{Adversarial Samples} &\makecell{Correct Hard Labels} & \makecell{Model outputs} & \makecell{get adversarial samples \& finetune models; -} & Black-box & \makecell{Copyright Verification} & classification &small image datasets & \makecell{Repair adversarial samples for harmless WM}\\

    & EWE~\cite{black_jia2021entangled}           & 2021 & \makecell{Samples w/ SNNL-guided\\[-2pt] Optimized Trigger Patterns}   &\makecell{Target Hard Labels} & \makecell{Model outputs} & \makecell{finetune w/ CE \& SNNL loss alternately; -} & Black-box & \makecell{Copyright Verification} & classification & small; image\&speech  & \makecell{Entangle tasks and WM in the same channels\\[-1pt] for a high  robustness against model extraction}\\   
    
    &Blind WM~\cite{black_li2019prove}           & 2019 & \makecell{Samples w/ Hidden Logo}    &\makecell{Target Hard Labels} & \makecell{Model outputs} & \makecell{finetune w/ CE loss; -} & Black-box & \makecell{Copyright Verification} & classification & small image datasets & \makecell{Hide copyright logo into triggers by a GAN model}\\

    & DeepAuth~\cite{fragile_lao2022deepauth} & 2022 &  \makecell{Near-boundary Samples}  & \makecell{Correct Hard Labels} & \makecell{Model outputs} &  \makecell{repair near-boundary adversarial samples; -} & Black-box & \makecell{Integrity Verification} & classification & small image datasets & \makecell{generate near-boundary samples as fragile WM} \\

    \\[-2ex]\cline{2-12}\\[-1.5ex]
    & UBW-P~\cite{black_li2022untargetedbackdoorwatermark} & 2018 & \makecell{Samples w/ a Pattern} &\makecell{Random Hard Labels} & \makecell{Model outputs} & \makecell{finetune or retrain w/ CE loss or\\[-1pt] inject poisoned samples into the dataset; -} & Black-box & \makecell{Copyright Verification\\[-1pt] Dataset Protection} & classification &small\&medium; images & \makecell{Inject harmless trigger samples into the target\\[-1pt] model or the protected dataset}\\

    & New Label~\cite{black_zhong2020newlabel}    & 2020 & \makecell{OOD Trigger Samples}   & \makecell{A New Hard Labels} & \makecell{Model outputs} & \makecell{retrain or finetune w/ CE loss; -} & Black-box & \makecell{Copyright Verification}  & classification & small image datasets & \makecell{treat crafted key samples as a new class}\\

    & Blackmarks~\cite{black_chen2019blackmarks}  & 2019 & \makecell{IP Message-guided\\[-2pt] Trigger Samples} &\multirow{2}{*}{\makecell{A bit string hidden in\\[-1pt] rearranged hard labels}} & \makecell{Model Outputs} & \makecell{finetune or retrain w/ CE loss; -} & Black-box& \makecell{Copyright Verification}  & classification &\makecell{small image datasets} & \makecell{use class clustering for multi-bit trigger-based WM}\\  

    & AIME~\cite{black_mehta2022aime} & 2022 & \makecell{Samples predicted in error} &  & \makecell{Outputs' confusion matrix} & \makecell{finetune or retrain  w/ CE loss; -} & Black-box & \makecell{Copyright Verification} & classification &\makecell{small image datasets} & \makecell{class mispredictions as WM w/ a confusion matrix}\\

    \\[-2ex]\cline{2-12}\\[-1.5ex]
    & MOVE-Output~\cite{both_li2022move} & 2022 & \makecell{Style-transferred Samples} &  \multicolumn{2}{c|}{\makecell{Clean-to-styled distances of Soft Predictions}} & \makecell{retrain w/ CE loss; -} & Black-box & \makecell{Copyright Verification} & classification & small\&medium; images & \makecell{embed external features into model outputs as WM}\\ 

    & CosWM~\cite{black_charette2022cosine} & 2022 & \makecell{Subset of Training Samples}  & \multicolumn{2}{c|}{\makecell{Specific Cosine Signals hidden in soft predictions}}    & \makecell{retrain w/ CE loss (modified soft labels);-}   & Black-box & \makecell{Copyright Verification} & classification & \makecell{small image dataset} & \makecell{embed cosine signals with a predefined frequency\\[-1pt] to defend against ensemble distillation}\\

    \\[-2ex]\cline{2-12}\\[-1.5ex]
    & Namba \myetal~\cite{black_namba2019robust} & 2019 & \makecell{Trigger Samples}  & \makecell{Target Hard Labels}   & \makecell{Model Output}  & \makecell{finetune exponentially-weighted weights; -}   & Black-box & \makecell{Copyright Verification} & classification & \makecell{small image dataset} & \makecell{exponential weighting against IP detection \& evasion}\\ 
    
    & Bi-level Opt~\cite{black_yang2021robust}   & 2021 & \makecell{Samples w\ highly \\[-1pt]uncertain outputs}  & \makecell{Target Hard Labels}  & \makecell{Model Output}  & \makecell{Optimize trivial weights like \cite{black_lao2022fewweights} and \\[-1pt] exemplars-to-be-embedded alternately;-} & Black-box &  \makecell{Copyright Verification}  & classification & \makecell{small\&medium; images} & \makecell{Enhance model fidelity and watermark robustness \\[-1pt] by bi-level optimization on examplers and weights}\\   
    
    & Target-Specific NTL~\cite{black_wang2022ntl} & 2021 & \makecell{Samples w/ a specific patch} & \makecell{Prediction Accuracy\\[-1pt] on the target domain} & \makecell{Model Output}  & \makecell{retrain w/ Inverted-IB loss to enlarge \\[-1pt] the source-target representation distances; -}  & Black-box & \makecell{Copyright Verification}  & classification & small image datasets  & \makecell{introduce the inverted information bottleneck \\[-1pt] loss to enhance the watermark robustness}\\

    & Certified WM~\cite{black_bansal2022certified} & 2022 & \makecell{Trigger Samples}  & \makecell{Target Hard Labels} & \makecell{Model Output} & \makecell{alternately train original \& gaussian-noised weights}  & Black-box & \makecell{Copyright Verification}  & classification & \makecell{small\&medium; images} & \makecell{Random smoothing on weights for certified WM}\\  
    
    & UBW-C~\cite{black_li2022untargetedbackdoorwatermark}  & 2022 & \makecell{Optimized Samples}  & \makecell{Clean Hard Labels} & \makecell{Model Output}  & \makecell{bi-level optimization on model and samples}  & Black-box & \makecell{Dataset Protection} & classification & small\&medium; images & \makecell{generate clean-label poison samples as watermarks}\\

    \\[-2ex]\cline{2-12}\\[-1.5ex]
    & Source-only NTL~\cite{black_wang2022ntl} & 2021 & \multirow{2}{*}{\makecell{Authorization Patterns}}
    & \multicolumn{2}{c|}{\makecell{Prediction accuracy on non-source domains\\[-1pt] generated by a GAN framework}} & \makecell{retrain w/ MMD-InvIB loss assisted by \\[-1pt] GAN-guided source domain augmentation} &- & \multirow{2}{*}{\makecell{Applicability\\[-1pt] Authorization}} & classification & small image datasets & \makecell{the first to propose the Deep IP function \\[-1pt] of applicability authorization}\\  
    
    & M-LOCK~\cite{new_ren2022cybersecurity}  & 2022 &   & \multicolumn{2}{|c|}{Low accuracy on samples w/o correct patterns} & \makecell{true labels only on samples w/ correct patterns}  & - &  & Classification & small\&medium; images & \makecell{train a pattern-sensitive decision boundary} \\ 

    \\[-2ex]\cline{1-12}\\[-1.5ex] 
    \multirow{6}{*}{\rotatebox{90}{\bf Output}}
    & Zhang \myetal~\cite{none_zhang2020model, none_zhang2021deep} & 2020 & \multirow{2}{*}{\makecell{Extractor Subnetwork}}  & \multirow{2}{*}{\makecell{ IP patterns Invisible\\[-1pt] in output images}} & \multirow{2}{*}{\makecell{Output Images}} & \multirow{2}{*}{\makecell{Train a watermarked generator or an embedder\\[-1pt] after the generator jointly w/ a WM extractor}}  & \multirow{3}{*}{No-box} & \multirow{2}{*}{\makecell{Copyright Verification\\[-1pt] \& Tracing model abuse}}  & \multirow{2}{*}{\makecell{image generation \& \\[-1pt] processing models}} & \multirow{2}{*}{\makecell{small\&medium; images}} & \multirow{2}{*}{\makecell{leverage watermarked images to watermark models}}\\  

    & Wu \myetal~\cite{none_wu2020watermarking} & 2020 &  &  &  &  &  &  & & & \\

    & Yu~\myetal~\cite{none_yu2021artificial} & 2021 & Extractor Subnetwork  & A bit string & Deepfake Outputs & \makecell{train an encoder-decoder to watermark images}  & & \makecell{Deepfake Detection}  & Deepfake & \makecell{small\&medium; images} & \makecell{Watermarks are transferred from inputs to outputs}\\  

    \\[-2ex]\cline{2-12}\\[-1.5ex]
    & AWT~\cite{nlp_abdelnabi2021awl} & 2022 & Extractor Subnetwork  & A bit string & Arbitrary Texts & \makecell{jointly train hider/revealer/detector transformers}  & - & \makecell{Text Watermarks}  & Text Generation & \makecell{medium; text} & \makecell{a transformer for imperceptible and robust text WM}\\  

    & WM4NLG~\cite{nlp_he2022dawnnlp} \& \cite{new_he2022cater} & 2022 & Predefined Rules  & A bit string & Generated Texts & \makecell{design some specific lexical rules of text generation}  & No-box & \makecell{Copyright Verification}  & Text Generation & \makecell{medium; text} & \makecell{watermark natural language generation models}\\  
    \\[-2ex]\cline{1-12}\\[-1.5ex] 

    \multirow{9}{*}{\rotatebox{90}{\bf Specific Scenarios}}
    & SRDW~\cite{black_jia2022srdw} & 2022 & \makecell{Near-OOD Samples by \\[-1pt]cross-domain augmentation}  & Special Hard Labels & Model Outputs & \makecell{finetune a selected core subnetwork for watermarking\\[-1pt] \& embed extra reversible compensation information}  & White\&Black & \makecell{Copyright Verification\\[-1pt]\& Integrity Verification}  & Transfer Learning & \makecell{small image datasets} & \makecell{generalization and redundancy of core subnetworks \\[-1pt] for reversible and robust WM against transfer learning}\\
    \\[-2ex]\cline{2-12}\\[-1.5ex]
    & WAFFLE~\cite{new_tekgul2021waffle} & 2020 & Trigger Samples  & Target Hard Labels & Model Outputs & \makecell{finetune models after server-side aggregation; Directly}  & Black-box & \makecell{Copyright Verification}  &  & \makecell{small image datasets} & \makecell{embed WM on the server side of a federated system}\\ 

    & FedIPR~\cite{fedwm_li2022fedipr} & 2021 & \multirow{2}{*}{\makecell{Secret Matrix \& \\[-1pt] Trigger Samples}}  & \multirow{2}{*}{\makecell{Client-aware bit strings \\[-1pt] \& trigger samples}}  &  \multirow{2}{*}{\makecell{Model outputs \& \\[-1pt] norm-layer parameters}} & \makecell{transfer WM from local client models to the global \\[-1pt] server model via updated gradients; Indirectly} & \multirow{2}{*}{White\&Black} & \multirow{2}{*}{\makecell{Copyright Verification\\[-1pt] \& Proof-of-work \\[-1pt] \& Traceability}}  & Federated Learning & \makecell{small image \& text sets} & \makecell{embed WM on client sides w/ a theoretical analysis\\[-1pt] between WM capacities and detection rates}\\

    & FedTracker~\cite{new_shao2022fedtracker} & 2022 &   &  &  &  global WM as IP signs \& local WM for traceability; I  & &  &  & \makecell{small image datasets} & \makecell{introduce continual Learning to improve fidelity}\\

    & CITS-MEW~\cite{new_wu2022cits} & 2022 & Trigger Samples  & Target Hard Labels & Model Outputs & \makecell{local WM enhancing \& global entangled aggregation; I}  & Black-box & \makecell{Copyright Verification}  &  & \makecell{small image datasets} & \makecell{multi-party entangled WM kept in global models}\\
    
    \\[-2ex]\cline{2-12}\\[-1.5ex]
    & GAN-IPR~\cite{passport_ong2021iprgan}  & 2021 &  \makecell{Passport Matrix \&\\[-1pt] Trigger Latent Codes} & \makecell{A unified bit string \&\\[-1pt]an IP pattern in outputs}  & \makecell{Norm-layer parameters\\[-1pt] \& Generated Images} & \makecell{multi-task learning w/ SSIM \& sign regularizer; Directly}  & White\&Black & \makecell{Copyright Verification}  & GAN models & \makecell{small\&medium; images} & \makecell{combine passports and backdoors to \\[-1pt] watermark models against IP ambiguity}\\ 
    \\[-2ex]\cline{2-12}\\[-1.5ex]
    & SSLGuard~\cite{selfsuper_cong2022sslguard} & 2022 &\makecell{Samples w/ a pattern \& \\[-1pt] a verification decoder} & \makecell{IP vectors output from\\[-1pt] the verification decoder} & \makecell{Latent Space of Encoders} & \makecell{train watermarked/shadow encoders \& a verification \\[-1pt] decoder jointly to embed detectable WM to latent codes}  & Black-box & \makecell{Copyright Verification}  & \makecell{Self-supervised\\[-1pt] Encoders} & \makecell{small image datasets} & \makecell{the first work to protect self-supervised encoders}\\
    \\[-2ex]\cline{2-12}\\[-1.5ex]
    &Sofiane \myetal~\cite{new_sofiane2021yes} & 2021 & Trigger samples in 3 types & Modified Labels & Model Output & \makecell{finetune models w/ task loss on trigger samples; }  & Black-box & \makecell{Copyright Verification}  & \multicolumn{3}{c|}{extend WM to machine translation, regression, binary image classification, \& reinforcement learning}\\
    
    &CoProtector~\cite{black_sun2022coprotector} & 2022 & \multicolumn{2}{c|}{Poisoned codes \& comments w/ predefined rules} & Output Codes/Comments & \makecell{untargeted and targeted poisoning for illegal model\\[-1pt] detection and performance degradation}  & No-box & \makecell{Code Repository Protection}  & \makecell{Neural Code Tasks} & \makecell{medium code datasets} & \makecell{the first work to protect open-source code repositories}\\ 

    &SpecMark~\cite{nlp_chen2020specmark} & 2020 & Secret Matrix  & A bit string & Spectrum of Weights & \makecell{WM-guided binarization on spectrum of weights;-}  & White-box & \makecell{Copyright Verification}  &Speech Recognition & \makecell{small\&medium; audio} & \makecell{the first work to watermark speech recognition models}\\
    \specialrule{0em}{1pt}{1pt}
    \hline
    \specialrule{0em}{1pt}{1pt}
    \multicolumn{2}{|c|}{Conflicting Interaction~\cite{new_szyller2022conflicting}} & 2022 &\multicolumn{4}{c|}{analyze the conflicts of different strategies for model security via constrained multi-objective optimization on task/protection performance}  &\multicolumn{3}{c|}{model/dataset IP protection, adversarial training, DP-SGD} & \makecell{small image datasets} & \makecell{a framework to analyze conflicts \& potential solutions}\\
    \bottomrule
  \end{tabular}
\end{center}
\end{sidewaystable*} 

\subsection{Overview}\label{sec:Protect_Invasive_Overview}
DNN Watermarking is an invasive way for Deep IP protection. It embeds a unique IP identifier, \ie a key-message pair $(\Kcal,\bbm)$ also named watermarks, into the original model $\phi$ by model finetuning/retraining on a watermark regularizer about the secret key $\Kcal$ and the watermark messages $\bbm$. The pirated versions, modified or extracted from the original model $\phi$, contain similar watermark messages $\bbm':dist(\bbm',\bbm)\rightarrow 0$. According to the host signals of watermarks, DNN Watermarking has the following types: 
\par
\textit{1) Inner-component-embedded Watermarking:} Watermarks can be embedded into internal components of DNN models, like model parameters, specific sparse structures, or neuron activation. It is usually realized by an extra watermark regularizer \textit{wrt} key-message pairs in loss functions.  
\par
\textit{2) Trigger-injected Watermarking:} It is to embed watermarks by finetuning/retraining the original model on modified train datasets that contains some trigger samples as secret keys. The labels of trigger samples are assigned by IP defenders and are significantly different from the predictions of unwatermarked (clean) models.  
\par
\textit{3) Output-embedded Watermarking:} For generative models, watermarks can be embedded/extracted into/from normal model outputs without specific triggers. Different from traditional data watermarking, this watermarking process is performed in model training. Here, watermark extraction relies on predefined rules or trained extractor DNNs.
This way generally works in watermarking image processing tasks or text generation tasks.
\par
Moreover, some schemes use a combination of these pathways or are designed for specific applied scenarios like self-supervised learning or transfer learning. 
This section summarizes the representative methods for DNN watermarking in Table~\ref{tab:watermarking} from the following attributes: 
\begin{itemize}[leftmargin=*]
	\setlength{\topsep}{0pt}
	\setlength{\itemsep}{0pt}
	\setlength{\parsep}{0pt}
	\setlength{\parskip}{0pt} 
\item \textit{Host Signals}: The carriers that watermarks embedded/extracted into/from, as we have discussed above. 

\item \textit{Secret Keys}: Necessary inputs to extract watermarks, only with which the watermark can be extracted correctly.

\item \textit{Watermark Messages}: Valid IP information embedded into the host signals of watermarked models, like bit strings, labels, or input/output patterns. 

\item \textit{Embedding Paradigms}: The main process to embed watermarks, including loss functions, training algorithms, etc. 

\item \textit{Embedding Scenarios}: Watermarks can be directly embedded into the DNN model or indirectly embedded by watermarked training samples or model gradients in distributed/federated learning.

\item \textit{Verification Scenarios}: The required model access for watermark extraction, like white-box, black-box, and no-box. 

\item \textit{Target Networks}:  
The structure of original models, like fully-connected networks (FC), convolutional neural networks (CNN), Recurrent neural networks (RNN), etc.

\item \textit{Target Tasks}: 
The types of DNN models for objective learning tasks/paradigms, such as classification or generative tasks.

\item \textit{Target Functions}: The IP functions aforementioned in Section~\ref{sec:background_problem} or their varieties, like copyright or integrity verification. 

\item \textit{Datasets}: The type/scale of target datasets like images or texts. 

\item \textit{Highlights}: The innovative highlights of research works from multi-views, like motivation and technical improvements. 

\end{itemize}

\begin{figure}[htbp]
	\centering
	\includegraphics[width=.48\textwidth]{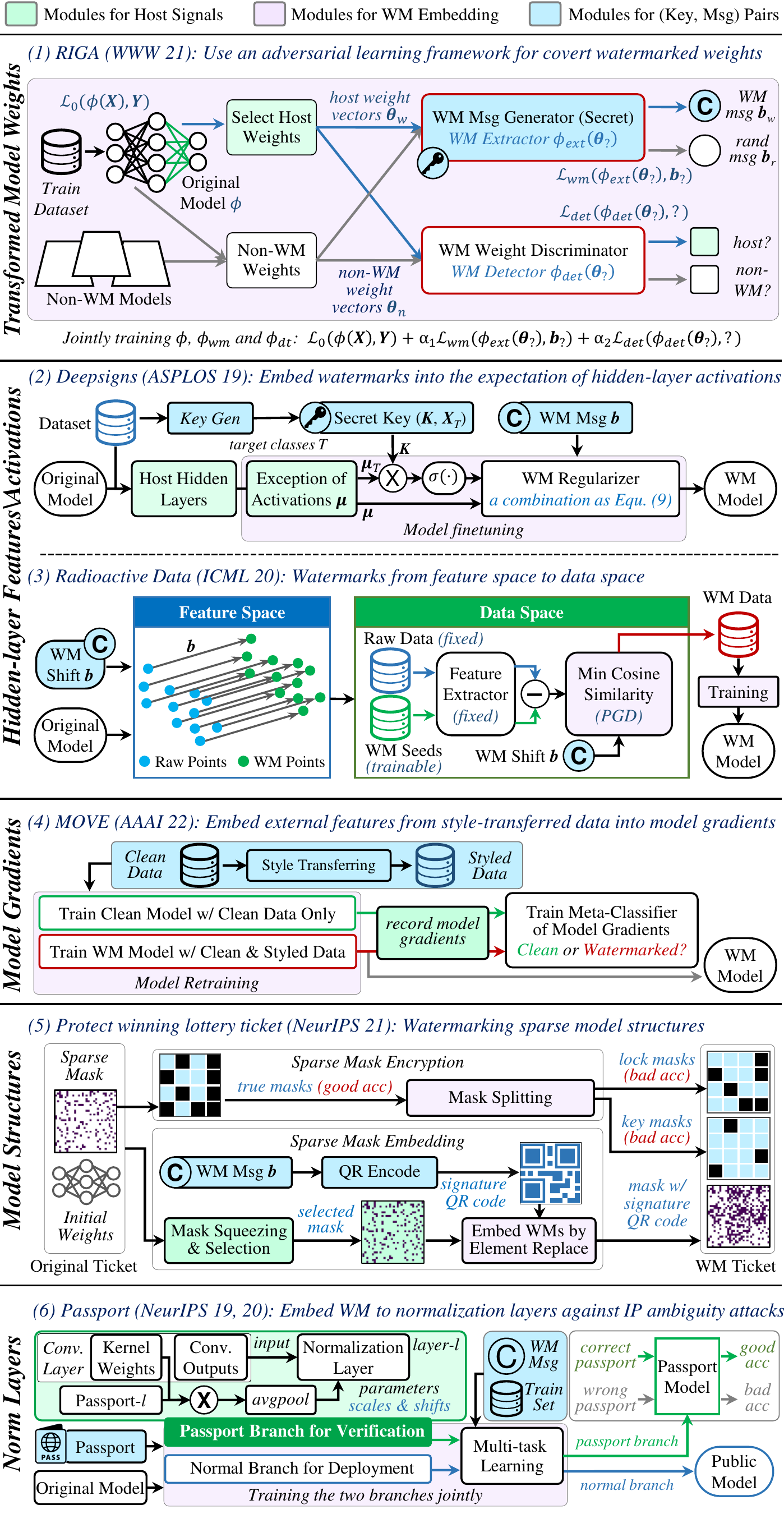}
 \vspace{-0.1in}
	\caption{\textcolor{black}{The representative methods for inner-component-embedded watermarking. Watermarks can be embedded into model parameters, hidden-layer activations, model gradients, model structures, etc.}}
	\label{fig:inner_component}
 \vspace{-0.3in}
\end{figure}
\subsection{Inner-component-embedded Watermarking}\label{sec:Protect_Invasive_Parameter}
Watermarks can be embedded into many optional model components, including static model weights, dynamic hidden-layer activation or gradients, model structures, and extra watermark components, as shown in Fig.~\ref{fig:inner_component}. In this section, we will discuss these methods in detail.
\subsubsection{Static Model Weights}\label{sec:Protect_Invasive_Parameter_Static}
A typical model watermarking is to embed watermarks into original or transformed model weights. It can embed robustness watermarks for copyright verification, or fragile/reversible watermarks for integrity verification.
\par
\textbf{Embedding watermarks into original model weights}. 
Uchida \myetal~\cite{white_uchida2017embedding,white_nagai2018digital} propose the first scheme for Deep IP Protection based on regularization techniques, \ie adding a watermark regularizer $\Lcal_R$ into original task loss $\Lcal_0$ as Equ.~\ref{eq:regularizer}. The watermark regularizer $\Lcal_R$ could be Cross-Entropy (CE), Mean-Square-Error (MSE), etc. We have described the details in the example in Section~\ref{sec:background_problem} and Fig.~\ref{fig:watermark_example}.
On this basis, many works are proposed for their demands. Chen~\myetal design DeepMarks~\cite{white_chen2018deepmarks} for user management. It 
embeds user-specific watermarks into each user's model for detecting undesired usages of distributed user models. The main threat to user management is the Collusion Attack, that multiple attackers use their models to produce an unmarked model. In order to support massive users and detect collusion attackers, it generates user messages from an orthogonal matrix according to the coded modulation theory. Here,  it uses the MSE loss as watermark regularizer $\Lcal_R=\textrm{MSE}(\thetabm_w\Kcal-\bbm_j)$ where $\bbm_j$ is the IP message of the $j$-th user.

 
 \nop{Collusion Attack. Multiple attackers who have the same host neural network with different embedded fingerprints may perform collusion attacks to produce an unmarked model. We consider fingerprints averaging attack which is a common collusion attack and demonstrate how DeepMarks is robust against such attacks.} 

\par 
\textbf{Embedding watermarks into transformed model weights}. Watermarks embedded into original model weights are vulnerable to IP detection and reconstruction. 
To enhance the covertness of watermarks, a straightforward idea is to embed watermarks into some transformation of model weights. It is based on an assumption: if watermark embedding does not change the weight distribution significantly, the watermarks can be considered covert. 
\par
\textit{(a) Embedding watermarks into the frequency domain of weights can achieve this purpose to some extent.} \cite{white_feng2020watermarking} and \cite{white_kuribayashi2021white} randomly select the host model weights from multiple layers and compute the coefficients of the frequency domain components of the selected host weights. Then, watermarks are embedded by quantifying the coefficients with the guide of preset IP messages. The inverse transformation is used to obtain the watermarked weights to replace the selected host weights. This method can avoid large amplitude changes in model weights and reduce the impact of watermark embedding on weight distribution. 
\par
\textit{(b) Generative Adversarial Networks act as weight transformation:}  
Considering robustness, covertness, and fidelity jointly, RIGA~\cite{white_wang2021riga} designs an adversarial learning framework, which mainly consists of two extra DNNs: a generator network used to extract or embed watermarks, and a discriminator network used for watermark hiding. 
The former DNN outputs valid IP messages given watermarked model weights otherwise outputs random messages, while the latter tries to distinguish between watermarked and unwatermarked weights. As shown in Fig.~\ref{fig:inner_component} (1),  the two extra DNNs along with the original model are trained jointly to obtain a watermarked model and a secret network (the generator). 
\par
\textit{(c) Specially-designed transformation rules can replace the role of preset secret key matrices.} IP ambiguity attacks, like forging attacks and overwriting attacks, can cause copyright conflicts by forging secret matrices (see details in Section~\ref{sec:attack_ambiguity}). To overcome this threat, Liu \myetal follow the "less is more" principle, and design Greedy Residuals~\cite{white_liu2021residuals} with no demands on explicit ownership indicators like trigger samples or secret matrices. It designs a specific rule to construct host signals (\ie a residual vector ${\bm \psi}\in \Rbb^n$) from the mean of important weights. First, a $1d$-average pooling is applied to transform reshaped weights $\thetabm\in\Rbb^{n\times d \times m}$ to $\gammabm\in\Rbb^{n\times d}$, where $n$ is the bit number of IP messages, $d$ and $m$ are two preset integers to reshape weights. 
Then for each row in $\gammabm$, select some elements with the largest absolute values and compute their mean values, and finally compose the residual vector ${\bm \psi}$. Next, watermarks are embedded into ${\bm \psi}$ using the sign loss as the watermark regularizer:
\vspace{-0.06in}
\begin{equation}\label{eq:sign_loss}
\Lcal_R({\bm \psi},\bbm_e) = 
\max(\alpha-\textrm{sgn}(\bbm_e^\mathsf{T}){\bm \psi},0) 
    \vspace{-0.06in}
\end{equation}
where $\textrm{sgn}\!\!:\!\!\{(0,+\infty],[-\infty,0]\}\!\!\rightarrow\!\!\{1,-1\}$ is a function to get the signs of input values. The regularizer $\Lcal_R$ encourages the elements of ${\bm \psi}$ to stay the same sign as that of $\textrm{sgn}(\bbm_e)$. Note that the $n$-bit message $\bbm_e$ is a  message encrypted from the original IP message $\bbm$ by the RSA algorithm.
\nop{The preset threshold $\alpha$ is to enlarge the gap between different signs of values.}
\nop{
Here, watermarks are embedded into a residual vector $\psi$ (extracted from important weights) by a regularizer $\Lcal_R$ defined as:
\vspace{-0.06in}
\begin{equation}
\Lcal_R = 
\max(\alpha-\textrm{sgn}(\bbm_{enc})^\mathsf{T}{\bm \psi},0) 
    \vspace{-0.06in}
\end{equation}
where the $n$-bit IP message $\bbm_{enc}$ is a bit string, \ie an encrypted message by the RSA algorithm; $\textrm{sgn}\!\!:\!\!\{(0,+\infty],[-\infty,0]\}\!\!\rightarrow\!\!\{1,-1\}$ is a function to get the signs of input values. 
The regularizer $\Lcal_R$ encourages the elements of ${\bm \psi}$ to stay the same sign as that of $\textrm{sgn}(\bbm_{enc})$. 
The preset threshold $\alpha$ is to enlarge the gap between different signs of values. 
}
\nop{
The residual ${\bm \psi}$ is equivalent to the mean of important weights: First, a $1d$-average pooling is applied to transform reshaped weights $\theta^l\in\Rbb^{n\times d \times m}$ of the $l$-layer to $\gamma^l\in\Rbb^{n\times d}$. 
Then for each row in $\gamma^l$, select some elements with the largest absolute values and compute their mean values, and finally compose the residual vector ${\bm \psi}\in \Rbb^n$.
}

\par
\textbf{Weight-based watermarks for integrity verification.} We can design reversible or fragile watermarks for integrity verification against model tampering attacks~\cite{white_guan2020reversible, white_botta2021neunac, new_zhao2022dnn, whtie_chen2019deepattest}.
\par
\textit{(a) Reversible watermarks.} Since irreversible watermarking destroys the original model permanently, 
Guan \myetal develop a reversible model watermarking method~\cite{white_guan2020reversible} for integrity verification inspired by channel pruning and histogram shift. The "reversible" means that the original model can be recovered by erasing the watermark in the watermark model. Then, it can determine whether the model is modified by comparing the hash codes (e.g., SHA-256) of the recovered and the original models. It is suitable for tampering detection during model distribution.
First, it selects the weights $\thetabm_w$ of $k$ unimportant network channels with minimum channel entropy values as the host sequence. This step applies the techniques in channel-wise network pruning.
\nop{Given the $l$-th layer with $d$-channels, the process to compute the recovered channel entropy $\{H_i|i\!=\! 1...d\}$ is: \ding{172} getting the $l$-th layer's activation tensor $\Acal^l \in\Rbb^{|\X_w| \cdot d\cdot h\cdot w}$ on some samples $\X_w$; \ding{173} using global average pooling to transform  $\Acal^l$ into a matrix $\hat{\Acal}^l \in\Rbb^{|\X_w| \cdot d}$; \ding{174} using the histogram estimation to establish the probability distribution $\pbm_i$ for each channel (each column of $\hat{\Acal}^l$) and computing the entropy $H_i$ for each $\pbm_i$.} 
Then, the float host sequence $\thetabm_w$ will be transformed to the integer host sequence $\hat{\thetabm}_w$ with the minimum entropy among all possible integer sequences. It is realized by intercepting two digits of each float element.
Finally, the integer host sequence $\hat{\thetabm}_w$ can be considered as a grayscale image and watermarks can be embedded via a reversible data-hiding strategy for images like histogram shift.  

\textit{(b) LSB-based fragile watermarks.} NeuNAC\cite{white_botta2021neunac} embeds the fragile watermark by modifying the least significant bytes (LSB) of model weights, for tampering detection and localization during model execution.
It divides model parameters into many Parameter Units (PUs) with 16 parameters. Each parameter is a 4-byte float number, where the first three bytes denote the most significant bytes (MSB) and the last byte denotes the LSB. A 32-byte Watermark Embedding Unit (WEU) is constructed for each PU, where each byte of the first 16 bytes is the hash code of the corresponding MSB and each byte of the last 16 bytes is the corresponding LSB. 
The watermark message $\bbm$ is embedded piecewise into the coefficients 
of Karhunen-Loève Transform (KLT) of each WEU while minimizing the change of LSBs by Genetic Algorithms, with the KLT basic vectors as the secret key $\Kcal$.
The verification on WEUs can be performed in parallel. Once the message extracted from any segment cannot match the watermark message $\bbm$, the model is tampered with. 

\textit{(c) Fragile watermarks for on-device functionality verification.}
The above methods allow no change on watermarked weights, however, their computation cost hinders on-device verification. Some scenarios only require functionality verification because a slight model modification imposed by attackers, such as bit-inversion attacks, can severely affect the model.
Based on DeepMarks~\cite{white_chen2018deepmarks}, Deepattest~\cite{whtie_chen2019deepattest} embeds device-specific watermarks to the original models and provides a software-hardware co-design approach to deploy the verification process in a Trusted Execution Environment (TEE). 
Once the functionality verification fails, the currently running model will be blocked. The considered threats mainly include \ding{172} Full-DNN Program Substitutions that attackers map illegitimate models to the protected device; \ding{173} Forgery Attacks that attackers forge the device-specific watermarks in the secure memory to mislead the verification; \ding{174} Fault Injection that attackers modify the model contents stored in the memory at a fine-grained level.

\subsubsection{Dynamic Hidden-layer Activation or Gradients}\label{sec:Protect_Invasive_Parameter_Dynamic}
Watermarks can be embedded into the input-related dynamic components within DNN models (\emph{e.g.}, model gradients or hidden-layer activations), which helps to improve model fidelity and Quality-of-IP (\emph{e.g.}, watermark covertness and robustness). 
\par
\textbf{Dynamic hidden-layer activations}.
Weight-based model watermarks are vulnerable to overwriting attacks (a way of IP reconstruction) that insert new watermarks to make the original watermark undetectable. 
\nop{An overwriting attack aims to insert an additional watermark in the model and render the original watermark unreadable.} 
Thus, Rouhani \myetal propose DeepSigns~\cite{white_rouhani2018deepsigns} to embed watermarks into\ the expectation of hidden-layer activations on the data samples in target classes $T$, as shown in Fig.~\ref{fig:inner_component} (2). 
We denote the host hidden-layer activation as $f(\xbm_i)$, where $\xbm_i$ denotes a data sample belonging to the class $i$. The watermark regularizer $\Lcal_R$ can be formulated as 
\vspace{-0.06in}
\begin{equation}
    \Lcal_R = \Lcal_\textrm{CE}(\sigma(\Kbf\mubm^T),\bbm) \!+\! \beta\sum_{\mathclap{i\in T, \xbm^i \in \Xbf^T}}\|f(\xbm^i)\!-\!\mubm^i\|\!-\!\beta\sum_{\mathclap{i\in T,j\notin T}}\|\mubm^i\!-\!\mubm^j\|,
    \vspace{-0.06in}
\end{equation}
where $\mubm^i$ is the expectation of hidden-layer activations on all selected data samples belonging to the class $i$, and $\mubm^T$ is the concat of all the expectation $\mubm^i$ for each class $i\in T$. Here, the first item is to project the activation exception $\mubm^T$ into the watermark message $\bbm$. The second item is to minimize the variance of the activation $f(\xbm^i)$ in the class $i$. The third item is to maximize the divergence of the activation between the target classes $T$ and other classes.

\nop{
The libraries in DeepSigns work by dynamically learning the probability density function (pdf) of activation maps obtained in different layers of a DL model. DeepSigns uses the low probabilistic regions within a deep neural network to gradually embed the owner’s message (watermark) while minimally affecting the overall accuracy and/or training overhead.}

\nop{To ensure the transformed selected Gaussian centers are as close to the desired WM information as possible, we design the second additive loss term that characterizes the distance between the owner-defined message and the embedded watermark}
\nop{To address the activation isolation constraint, we design an additive loss term that penalizes the activation distribution when activations are entangled and hard to separate.}

\textbf{Dynamic Gradients}.
Experiments conducted in~\cite{both_li2022move} show that, in the way of backdoor watermarking, the model predictions of preset trigger samples are hard to match between the watermarked model and its pirated versions. Moreover, extra security risks like backdoors are introduced. Fingerprinting methods like dataset inference~\cite{fingerprinting_maini2021datasetinference} may make misjudgments and provide incredible results; so the results are unreliable. 
Towards effective and harmless model watermarking, 
MOVE~\cite{white_li2022defending,both_li2022move} embeds external features from style transferring into the original model, instead of changing the labels of trigger samples. The external features are hidden in model gradients of style-transferred input samples.
As shown in Fig.~\ref{fig:inner_component} (4), MOVE mainly consists of three steps: \ding{172} generating some style-transferred samples $\X_w$ from the clean train dataset $\X$; \ding{173} training a watermarked model $\phi_w$ on both of the clean and the styled dataset $\{\X_w, \X\}$,  and training a clean model $\phi$ on the clean dataset $\X$; \ding{174} getting the gradients of the two trained model $\phi_w$ and $\phi$ on the styled dataset $\X_w$, and training a binary meta-classifier to distinguish which model the gradients are from. 
Since model gradients cannot be obtained in black-box scenarios, MOVE adopts the difference of soft predictions between augmented style-transferred samples and the original samples, as the substitution of model gradients. Finally, hypothesis testing is performed on the predictions of the trained meta-classifier to obtain the verification result. 

\textbf{For dataset protection: watermarks from feature space to data space}. 
Watermarks can be embedded into the feature space by adding additive patterns to training samples. This way does not require a specific model training process and can therefore be used to protect the ownership of both models and datasets, that is, to identify the models illegally trained on the protected datasets. Radioactive Data~\cite{data_sablayrolles2020radioactive} is a typical method for such studies. It represents the model as the combination of a linear classifier $\wbm$ and a feature extractor $f$, \ie $\wbm f(\xbm)$. Its watermark embedding stage aims to embed an isotropic unit vector $\ubm_i$ into the feature space of all training samples of each class $i$, \ie the output of the feature extractor $f(\xbm)$. This  process is done through the pixel-level optimization on training samples, formulated as 
\vspace{-0.06in}
    \begin{equation*}
        \min_{\hat{\xbm}} -(f(\xbm)-f(\hat{\xbm}))^\top\ubm +\lambda_1\|\xbm-\hat\xbm\|_2+\lambda_2\|f(\xbm)-f(\hat{\xbm})\|_2,
        \vspace{-0.06in}
    \end{equation*}
where the first term is to align the features with $\ubm$ while the last two items are to minimize the impact of embedded patterns on training samples and their features. Given a suspect model $\wbm_s f_s(\xbm)\approx\wbm_s\Abm f(\xbm)$, the cosine similarity cos($\ubm$,$\wbm_s\Abm$) should follow the beta-incomplete distribution if the model is trained on normal samples $\xbm$, otherwise, if the model is trained on protected data $\hat\xbm$ or finetuned from the protected model $\wbm f(\xbm)$, the cosine similarity will be significantly higher. This method can be extended to black-box verification by extracting surrogate models from the suspect model. 

\nop{Effective and Harmless: effectiveness requires that it can accurately identify whether the suspicious model is stolen from the victim, no matter what model stealing is adopted; Harmlessness ensures that the model watermarking brings no additional security risks, i.e., the model trained with the watermarked dataset should have similar prediction behaviors to the one trained with the benign dataset.} 

\nop{consists of three main steps, including 1) embedding external features, 2) training ownership meta-classifier, and 3) ownership verification with hypothesis-test. In general, the external features are different from those contained in the original training set. Specifically, we embed external features by tempering the images of a few training samples based on style transfer. Since we only poison a few samples and do not change their labels, the embedded features will not hinder the functionality of the victim model and will not create a malicious hidden backdoor in the victim model. Besides, we also train a benign model based on the original training set. It is used only for training the meta-classifier to determine whether a suspicious model is stolen from the victim. In particular, we develop our MOVE method under both white-box and black-box settings to provide comprehensive model protection}

\subsubsection{Model structures}\label{sec:Protect_Invasive_Parameter_Structure}
Model structures are not only the host signals of watermarks but also valuable assets requiring protection. 
\par
\textbf{Model structures for watermarks}. 
Watermarking sparse model structures is exploited for Deep IP Protection.
Some methods~\cite{white_zhao2021structural,new_xie2021deepmark} apply model pruning to embed watermarks into model structures rather than parameters to bypass parameter modification (a way of IP removal).
\cite{white_zhao2021structural} applies channel pruning for watermark embedding. The IP message is piecewise embedded into the channel-pruning rates of multiple convolutional layers and the watermarked model will be obtained by channel pruning using the watermarked pruning rates. \cite{new_xie2021deepmark} considers weight-level pruning and embeds watermarks into the model weights with low connection sensitivity. However, these methods are vulnerable to channel pruning and model extraction using different model structures. 
\par
\textbf{Watermarking model structures}. 
High-performance model structures also need protection. 
The lottery ticket hypothesis~\cite{pruning_frankle2018lottery} has demonstrated that the model trained on a subnetwork with a specific sparse mask and initial weights, \ie a winning ticket, can provide accuracy close to the full network while significantly reducing inference and training costs. Thus, costly-found winning tickets can be regarded as valuable assets. 
Chen~\myetal explore a novel way to protect the ownership of winning tickets~\cite{lottery_chen2021you}. 
The $l$th layer of a winning ticket is defined as $\Mbf^l\odot\thetabm_0^l$, where $\thetabm_0^l$ and $\Mbf^l$ denote the initial weights and the sparse mask respectively, and $\odot$ denotes the Hadamard product (element-wise multiplication). 
\cite{lottery_chen2021you} splits the mask with good accuracy into a key mask $\Mbf_{\textrm{key}}^l$ and a lock mask $\Mbf_{\textrm{lck}}^l$ with bad accuracy, where $\Mbf^l\!=\!\Mbf_{\textrm{key}}^l\!+\!\Mbf_{\textrm{lck}}^l$. Only the legitimate user with both the key mask and the lock mask can train a good model. 
A QR code containing IP messages is also embedded into the sparse mask by element replacement for white-box copyright verification. 
Neural Architecture Search (NAS), which automatically discovers a good model structure for a given task and dataset, provides model structures better than human-designed structures. It consumes huge computation resources; thus, Lou \myetal~\cite{gray_lou2021meets} treat high-performance model structures from NAS as important assets and propose embedding watermarks into the searched model structures. 
\cite{gray_lou2021meets} creates the owner-specific watermarks by fixing some neuron connections in the search space and then searches the structures from the rest space. Moreover, \cite{gray_lou2021meets} designs a better cache-based side-channel analysis method to extract model structures. 

\subsubsection{Extra components}\label{sec:Protect_Invasive_Parameter_Extra}
Watermarks can be embedded by adding extra processes or components into the original model. Unlike the aforementioned watermarking schemes, this category of methods change the process of model inference via extra steps and thus some of them support access control. Here are some typical methods:

\textbf{Extra processing on normalization layers}.
Watermarking methods independent of the original task loss are vulnerable to IP ambiguity attacks. To overcome this limitation, Passport~\cite{passport_fan2019rethinking, passport_fan2021deepip} appends a passport layer after each selected hidden layer; then the original model can make valid predictions (or be finetuned/retrained) only with the correct passport $\Pcal=\{\Pbf_\gamma^l, \Pbf_\beta^l|l=1...L\}$. $\Pbf_\gamma^l$ and  $\Pbf_\beta^l$ are two input matrice of the $l$-th passport layer $f_{p}^l$ defined as:
\vspace{-0.06in}
\begin{equation*}
\begin{split}
    f_{p}^l &= {\gamma^l}(\Pbf_\gamma^l) \cdot f^l(\xbm^l) + {\beta^l}(\Pbf_\beta^l), \ f^l(\xbm^l)= \thetabm^l \otimes \xbm^l \\
     {\gamma^l}&(\Pbf_\gamma^l) = avg(\thetabm^l \otimes\Pbf_\gamma^l), \ {\beta^l}(\Pbf_\beta^l) = avg(\thetabm^l \otimes\Pbf_\beta^l) \\[-4pt]
\end{split}
\end{equation*}  
where ${\gamma^l(\Pbf^l)}$ and ${\beta^l}(\Pbf_\beta^l)$ are the scale and bias parameters of normalization layers; $\thetabm^l$ denotes the parameters of the $l$-th hidden layer; $\otimes$ denotes the operation of the hidden layer (e.g., convolution or multiplication). Moreover, IP messages could also be embedded into some scale and bias parameters via a sign loss as Equ.~(\ref{eq:sign_loss}). 
Here, model owners can distribute the passport $\Pcal$ to legitimated users along with the watermarked model, or jointly train the passport branch and the normal branch by multi-task learning and distribute the normal branch. 
However, batch normalization is not compatible with multi-task learning while group normalization will degrade model accuracy. Therefore, zhang~\myetal propose passport-aware normalization~\cite{passport_zhang2020passport} as a solution. \cite{passport_zhang2020passport} trains an extra passport-aware branch, formulated as 
\begin{equation*}
\hat{f}^l\!=\!\left\{
\begin{aligned}
    &\gamma_0 \tfrac{f^l(\xbm^l)-\mu_0(f^l(\xbm^l))}{\sigma_0(f^l(\xbm^l))}+\beta_0,& &\hspace{-2mm}\textrm{passport-free}, \\
    &\gamma_1 (\Pbf_\gamma^l) \tfrac{f^l(\xbm^l)-\mu_1(f^l(\xbm^l))}{\sigma_1(f^l(\xbm^l))}\!+\!\beta_1(\Pbf_\beta^l),& & \hspace{-2mm}\textrm{passport-aware}, \\[-4pt] 
\end{aligned}
\right.
\end{equation*}
where $\gamma_1 (\Pbf_\gamma^l) = \thetabm_2 \cdot h(\thetabm_1\cdot  avg(\thetabm^l_p \otimes\Pbf_\gamma^l))$, $\beta (\Pbf_\beta^l) = \thetabm_2 \cdot h(\thetabm_1\cdot  avg(\thetabm^l_p \otimes\Pbf_\beta^l))$ are output from two fully connected layers with the weights $\thetabm_1$ and $\thetabm_2$. Different from~\cite{passport_fan2019rethinking, passport_fan2021deepip}, here the mean/std statistics are computed separately
for the passport branch $\mu_1, \sigma_1$  and the normal branch $\mu_0, \sigma_0$, thereby supporting multi-task learning. Moreover, the learnable weights to transform passports reduce the accuracy degradation caused by the passport branch. 

\par
\textbf{Extra processing on RNN hidden states}. \cite{passport_lim2022rnn} 
extends watermarks to recurrent neural networks (RNN).
With reference to \cite{passport_fan2021deepip}, Lim~\myetal realize access control and copyright verification for both black-box and white-box scenarios.
Here, the protected model is trained using the secret key matrix $\Kbf$ to transform the original hidden states ${\bm h}$ into $\hat{\hbm}$ after each time step, which is formulated as 
\vspace{-0.06in}
\begin{equation*}
    \hat{\hbm} =  \Kbf \otimes \hbm,\ \Kbf=BE\odot BC
    \vspace{-0.06in}
\end{equation*}
where $BE$ is the binary representation of the owner's key string, $BC\in \{-1,1\}^{|\Kbf|}$ is a random sign vector, ${\hbm}$ is the original hidden state of a time step. 
Here, $\odot$ denotes element-wise multiplication, and $\otimes$ can be selected as element-wise multiplication or element-wise addition. Similar to \cite{passport_fan2021deepip}, an IP message $\bbm$ can be embedded into the hidden state ${\bm h}$, using the sign loss $\Lcal_R({\bm \hat{h}},\bbm)$ as Equ.~(\ref{eq:sign_loss}). 
The loss forces the hidden state ${\bm h}$ to have the same sign as $\textrm{sgn}(\bbm)$. 



\nop{
The main motivation of embedding digital passports is to design and train DNN models in a way such that, their inference performances of the original task (i.e. classification accuracy) will be significantly deteriorated due to the forged messages.}
\nop{However, one of the main differences of our approach compared to [8] is our message is not embedded in the model weights, but it is embedded in the hidden state which is the output of the LSTM cell. This is because we found out that embedding message in the model weights can be easily attacked with a channel permutation, i.e., change the message but remains the output of the model.}

\par
\textbf{Subnetworks as Watermarks}. Most watermarking methods based on inner-components cannot theoretically guarantee robustness against IP removal attacks. To this end, lv \myetal~\cite{new_lv2023robustness} embed a watermark subnetwork into the original model, and delicately design a regularizer loss for watermark embedding to theoretically guarantee that the accuracy of the task model is more sensitive to fine-tuning attacks than that of the watermark subnetwork. Moreover, by such subnetwork-based watermark embedding, \cite{new_lv2023robustness} achieves watermark stealthiness and forgery resistance to a certain extent. 
It first trains the HufuNet, an encoder-decoder model that inputs and then reconstructs trigger samples.  
All the encoder's parameters, as the watermarks, are embedded into the original model where the embedding location of each parameter is determined by the hashcode \textit{wrt} its mirrored parameter in the decoder, the preset secret key, and its parameter index. Then it alternately trains the watermark subnetwork and the original model to ensure the accuracy of the two models. 
\nop{
The loss function is defined as:
\vspace{-0.04in}
\begin{equation*}
\begin{split}
    \Lcal=\Lcal_{\rm task}+&\left(\frac{\Ebb[\nabla_g\Lcal_{\rm task}]}{\Ebb[\nabla_{g_w}\Lcal_{\rm task}]}\!-\!\alpha_1\right)^2 \!+\!\left(\frac{\Ebb[\nabla_g\textrm{PCC}]}{\Ebb[\nabla_{g_w}\textrm{PCC}]}\!-\!\alpha_2\right)^2\\
    +&\left(\frac{var[\nabla_g\Lcal_{\rm task}]}{var[\nabla_{g_w}\Lcal_{\rm task}]}-\alpha_3\right)^2,\\
    \textrm{s.t. }\ \  & \alpha_1\cdot\alpha_2>1, \ \ \alpha_3<\alpha_1^2,
    \vspace{-0.04in} 
\end{split}
\end{equation*}
where $\textrm{PCC}$ is the normalized loss change due to the fine-tuning attack; $\alpha_1,\alpha_2,\alpha_3$ are preset constants. 
It ensures that the original model fails before the watermark subnetwork under fine-tuning attacks.
}
The decoder is saved by the verifier. In the verification stage, the verifier extracts an encoder from the suspect model, and merges the extracted encoder and the decoder into the whole HufuNet. 
It is considered as an infringement if the reconstruction performance on trigger samples is higher than a threshold.



\begin{figure}[htbp]
	\centering
	\includegraphics[width=.45\textwidth]{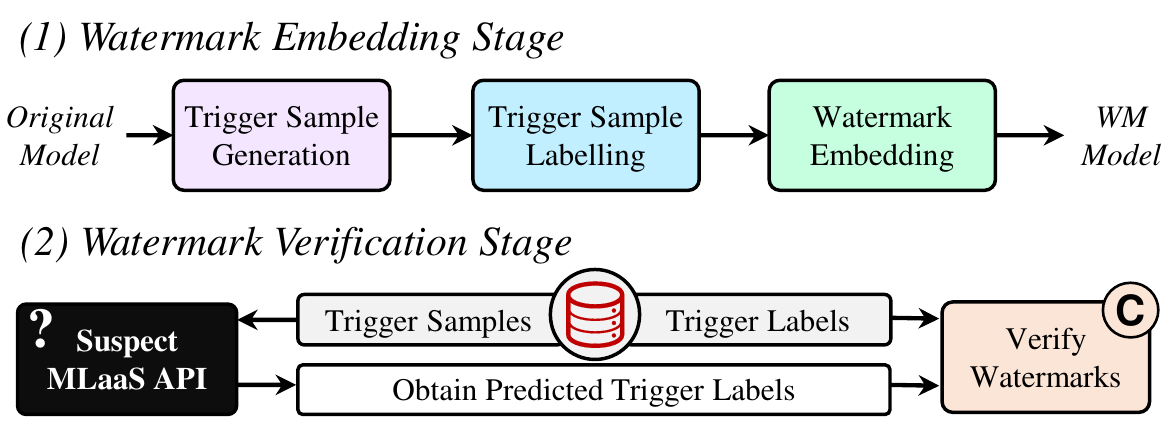}
  \vspace{-0.1in}
	\caption{The pipeline of trigger-injected watermarking.}
	\label{fig:trigger_pipeline}
  \vspace{-0.3in}
\end{figure}
\subsection{Trigger-injected Watermarking}\label{sec:Protect_Invasive_Trigger}
Most trigger-injected watermarking methods are designed for black-box verification and have three main points as shown in Fig.~\ref{fig:trigger_pipeline}: \ding{172} generating trigger samples $\X_w$ as the secret keys of watermarks; \ding{173} labelling trigger samples, \ie determining the expected outputs $\Ybf_w$ of trigger samples $\X_w$ on the watermarked model $\phi_w$ that should be significantly different from that on irrelevant models $\phi_{-}(\X_w)$;
and \ding{174} embedding paradigms, \ie making the output $\phi_w(\X_w)$ close to trigger labels $\Ybf_w$ by minimzing the loss formulated as
\vspace{-0.06in}
\begin{equation} 
    \Lcal=\Lcal_0(\Xbf,\Ybf;\thetabm)+\lambda\Lcal_R(\X_w,\Ybf_w;\thetabm_w). 
    \vspace{-0.06in}
\end{equation}
\subsubsection{Generating trigger samples}\label{sec:Protect_Invasive_Trigger_Generation}
\begin{figure*}[htbp]
	\centering
	\includegraphics[width=.95\textwidth]{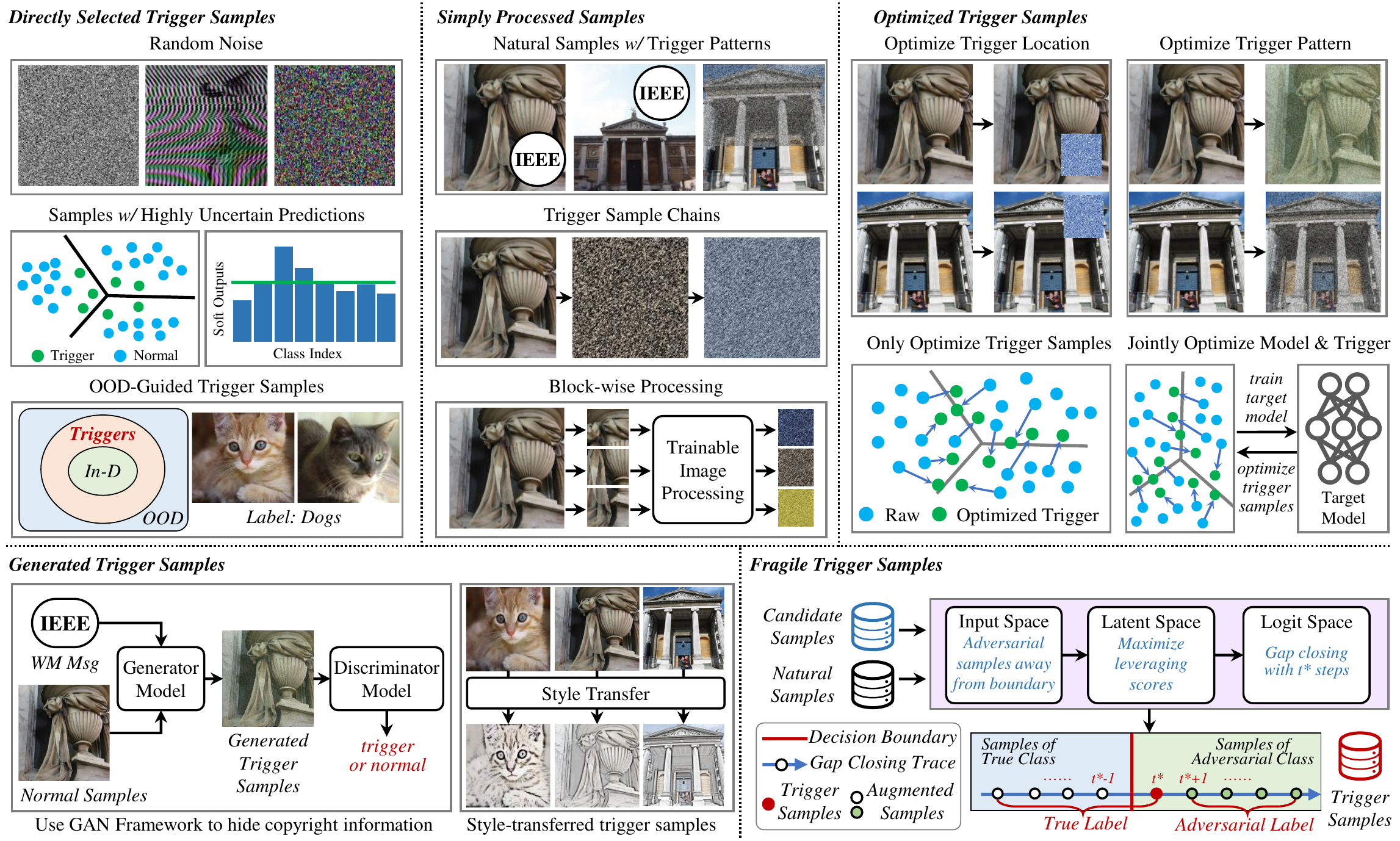}
  \vspace{-0.12in}
	\caption{\textcolor{black}{The representative methods for trigger sample generation.}}
	\label{fig:trigger_sample_generation}
  \vspace{-0.2in}
\end{figure*}
The quality of trigger samples is the primary factor to determine the performance of trigger-injected model watermarking. As shown in Fig.~\ref{fig:trigger_sample_generation}, trigger samples can be constructed by direct selection, simple processing, gradient optimization, generative model, etc. 
\par
\textbf{Trigger samples can be directly selected or simply processed from datasets.} Here are some typical cases:
\begin{itemize}[leftmargin=*]
	\setlength{\topsep}{0pt}
	\setlength{\itemsep}{0pt}
	\setlength{\parsep}{0pt}
	\setlength{\parskip}{0pt} 
  \item  \textit{Out-Of-Distribution (OOD) samples as trigger samples}: Adi \myetal \cite{black_adi2018turning} propose the first trigger-based watermarking method. Here, the trigger samples are abstract, out-of-distribution images with randomly assigned labels from uncorrected classes. Assigning target hard labels is also feasible~\cite{black_zhang2018protecting}. Trigger samples can also be selected according to the OOD degree, \ie the position (or likelihood) in the data distribution~\cite{white_rouhani2018deepsigns, black_jia2022srdw}. For example, 
  Deepsigns~\cite{white_rouhani2018deepsigns} proposes to select trigger samples from rarely explored regions where each sample has few neighbors within an $\varepsilon$-ball of activations. SRDW~\cite{black_jia2022srdw} is to choose an appropriate OOD degree to trade off model fidelity and IP robustness. 
  \item \textit{Samples with highly uncertain prediction results~\cite{black_lao2022fewweights}}: The samples, whose soft predictions are close across classes, are usually nearby the decision boundary of the original model. Their predictions are easy to be manipulated within minor model modifications. These samples can be embedded while ensuring model fidelity. Generally, it can be measured by the Shannon Entropy of the model predictions. 
   \item \textit{Random noises as trigger samples}: Noises with the same dimension as input samples can also be used as trigger samples, like noises sampled from a Gaussian distribution~\cite{new_sofiane2021yes,black_zhang2018protecting}.
  \item \textit{Natural samples attached with secret, specific additive trigger patterns}: Most trigger-based watermarking schemes~\cite{black_jia2021entangled, black_zhang2018protecting, black_guo2018watermarking} apply this type of trigger samples, which can be formalized as $\xbm_w\! =\! (1\!-\!\alpha)\xbm\!+\!\alpha\tbm$ or $\xbm_w\! =\! (1\!-\!\Gamma)\odot\xbm\!+\!\Gamma\odot\tbm$. Here, $\tbm$ denotes the trigger pattern, $\alpha\in(0,1]$ is a hyperparameter, and $\Gamma$ is a mask matrix to select the position of the trigger pattern. 
  The trigger pattern can consist of specific content, such as a string that covers part of an image, or it can be random noise like Gaussian noise. Additionally, these additive patterns may also be designed to imply the copyright information of model owners~\cite{black_guo2018watermarking}.
  \item \textit{A chain of trigger samples processed by a hash function}: An attacker can forge trigger samples with abnormal labels by means like adversarial attacks. In order to disable forged trigger samples, Zhu \myetal \cite{black_zhu2020secure} propose to execute the hash function multiple times on seed samples to generate a chain of trigger samples with assigned labels. In this way, attackers are hard to construct a chain of trigger samples without model modification. 
  \item \textit{Block-wise processing}:  Predefined operations enable the generation of trigger samples from arbitrary training data without prior specification. For example, Maung \myetal~\cite{black_maung2021piracy} design a way of block-wise transformation associated with a preset secret key. 
\end{itemize}
\par
\textbf{Copyright information can be embedded into trigger samples via generative models}. Zheng \myetal leverage a generative-adversarial framework to generate covert trigger samples hiding the owner's logo~\cite{black_li2019prove}. Here, the "\textit{covert}" refers that the generated trigger samples are visually similar to task samples and hard to detect by attackers.
This framework consists of two DNN models: i) a trigger generator that hides the owner's logo into natural samples and outputs the generated trigger samples, and ii) a discriminator that tries to distinguish between trigger samples and task samples.
\par
\textbf{Synthetic samples via gradient optimization can be used as trigger samples.} 
A typical method to optimize synthetic samples is adversarial attacks that create adversarial samples by adding carefully-crafted small perturbations $\Delta$ to seed samples $\X_0$ and induce the original model to produce error predictions on $\X_0+\Delta$.
\par
\textit{Adversarial Samples}: 
Given some seed samples with the labels $\ybm_0$, the adversarial sample $\xbm_\Delta$ could be generated by untargeted adversarial attacks, which is formulated as  
\vspace{-0.06in}
\begin{equation}\label{eq:untarget}
    \max_\Delta \Lcal_\textrm{0}(\phi(\xbm_\Delta), \ybm_0), \textrm{s.t.}\ \xbm_\Delta \!=\! \xbm_0\! +\! \Delta, \|\Delta\|\! \leq\! \epsilon, \phi(\xbm_\Delta)\!\neq\!\ybm_0,
    \vspace{-0.06in}
\end{equation}
or targeted attacks to produce the target predictions $\ybm_\Delta\neq\ybm_0$, as
\vspace{-0.06in}
\begin{equation}\label{eq:target}
    \min_\Delta \Lcal_\textrm{0}(\phi(\xbm_\Delta), \ybm_\Delta) , \ \textrm{s.t.}\ \xbm_\Delta\! =\! \xbm_\Delta\! +\! \Delta,\  \|\Delta\| \leq \epsilon,
    \vspace{-0.06in}
\end{equation}
where $\epsilon$ is to limit the magnitude of the perturbations $\Delta$. 
Equ.~(\ref{eq:untarget}) can be solved by the Fast Gradient Sign Method (FGSM) as 
\vspace{-0.06in}
\begin{equation}
    \xbm_w = \xbm + \epsilon \cdot \textrm{sgn}(\nabla_x \Lcal_\textrm{0}(\phi(\xbm), y_0)), \forall \xbm \in \X_0
    \vspace{-0.06in}
\end{equation}
where $\xbm_w$ is the adversarial sample of $\xbm$. Projected Gradient Descent (PGD) can also solve this problem, where the $t$-th iteration is  
\vspace{-0.06in}
\begin{equation}
    \xbm_{t+1} = {\prod}_{B_\epsilon}(\xbm_{t} + \alpha \cdot \textrm{sgn}(\nabla_x \Lcal_\textrm{0}(\phi(\xbm), y_0))), \forall \xbm_0 \in \X_0
    \vspace{-0.06in}
\end{equation}
where $\textrm{sgn}\!\!:\!\!\{(0,+\infty],[-\infty,0]\}\!\!\rightarrow\!\!\{1,-1\}$ is the function to get the signs of input values. ${\prod}_{B_\epsilon}(\cdot)$ is an operation to project the input to a small range $B_\epsilon$, a ball with the center $\xbm^0$ and the radius $\epsilon$, to limit the perturbation. The step length $\alpha$ is much smaller than $\epsilon$.
\nop{Adversarial examples are images that have been modified (often imperceptibly) to trigger a misclassification when a DNN predicts the image’s label. The authors generate these adversarial examples using  and the pre-trained source model. This method has a given probability of failure, meaning that its output is not adversarial and is correctly classified by the DNN. The watermarking key is composed of such false adversarial examples and equally many true adversarial examples. All adversarial examples are labeled by the groundtruth label in the watermarking key. The embedding and extraction process is the same as Adi.}

\textit{Optimizing trigger samples via gradient optimization}: 
Frontier-Stitching~\cite{black_le2020adversarial} is the first work to introduce adversarial samples into trigger-injected watermarking. It generates adversarial samples by untargeted FGSM as trigger samples $\X_w$, where some samples successfully change their predictions but others do not. Then, it select the true labels $\ybm_0$ of $\X_w$ as trigger labels and finetunes the original model on $\{\X_w,\ybm_0\}$. We name this process repairing adversarial samples.  
Such gradient optimization techniques are widely used in trigger sample generation~\cite{black_jia2021entangled, fragile_lao2022deepauth, black_li2022untargetedbackdoorwatermark, black_yang2021robust}. Some methods only optimize trigger samples~\cite{black_le2020adversarial, black_jia2021entangled, fragile_lao2022deepauth}, while the other some jointly optimize the original model and trigger samples~\cite{black_yang2021robust, black_li2022untargetedbackdoorwatermark}. Both the locations and the patterns of trigger patches can be optimized~\cite{black_jia2021entangled}.

\textbf{Trigger-based fragile watermarks.} Trigger samples can be designed to be fragile for integrity verification~\cite{fragile_lao2022deepauth, fragile_zhu2021fragile, fragile_lin2022deepensemble}. The core idea is to make the predictions of trigger samples sensitive to any model modification. 
 Deepauth~\cite{fragile_lao2022deepauth} is a typical scheme for trigger-based fragile watermarks. It adopts a similar idea to Frontier-Stitching~\cite{black_le2020adversarial}, \ie repairing adversarial samples. However, Deepauth generates fragile trigger samples closer to the decision boundary but far away from natural samples, resulting in only minor shifts in the watermarked model's decision boundary when modifying predictions for these fragile trigger samples.
 The "fragile" refers that a slight model modification will change the predictions of trigger samples. 
 The process for generating trigger samples mainly consists of three steps: i) from input space, Deepauth seeks the adversarial samples $\X_{w,1}$ that can change the predictions only when the $l_p$-distance $l_p(\X,\X_{w,1})$ is larger than a threshold $\epsilon_d$, as 
\begin{equation*}
    \begin{split}
        {\min}_{\X_{w,1}}\ &\|\Lcal_{\rm CE}(\phi(\X_{w,1}),\Ybf_w) -\Lcal_{\rm CE}(\phi(\X_{w,1}),\Ybf)\|, \\
        \textrm{s.t.}\ \ \ \ \  \ &\Ybf_w=\phi(\X_{w,1})\neq\Ybf,\ l_p(\X,\X_{w,1})\geq\epsilon_d;
    \end{split}
\end{equation*}
ii) from latent space, Deepauth applies leverage score sampling to estimate the uncertainty of $\X_{w,1}$ and selects the samples with the largest leverage scores as the trigger samples $\X_{w,2}$ towards a minimal impact on fidelity;
and iii) from logit space, the samples $\X_{w,2}$ are further moved to the decision boundary by reducing the logit difference between the original and modified classes, via an iterative gradient optimization where the $t$-step is expressed as 
\vspace{-0.06in}
\begin{equation*}
\xbm_{t+1}\leftarrow\xbm_{t}+\alpha\cdot\textrm{sgn}(\nabla_{\xbm_{t}}[\Lcal(\phi(\xbm_{t}),\ybm_w)\!-\!\Lcal(\phi(\xbm_{t}),\ybm)]),
\vspace{-0.06in}
\end{equation*}
where $\alpha$ is the step length and $\xbm_{0}\!\in\!\X_{w,2}$. The iteration step just changing predictions is recorded as $t^*$, $\xbm_{t^*}$ is a trigger sample and $\phi(\xbm_{t^*})\!=\!\ybm_w$. 
Then, Deepauth repairs the adversarial trigger samples $\X_{t^*}\!=\!\{\xbm_{t^*}\}_i$, \ie finetuning the original model $\phi$ on true-labelled trigger samples $\{\X_{t^*},\Ybf\}$ along with the augmented set $\{\X_{t^*-1},\Ybf\}\cup\{\X_{t^*+1},\Ybf_w\}$, to embed trigger samples while bounding the decision boundary shift. 
\nop{
$\X_{t^*}\!=\!\{\xbm_{t^*}\}$ is the trigger samples, given which the original model outputs $\phi(\X_{t^*})\!=\!\Ybf_w$. Then the model is fine-tuned on the trigger set $[\xbm_{t^*},\ybm]$ along with the augmented set $[\X_{t^*-1},\ybm]\cup[\X_{t^*+1},\ybm_w]$ to embed trigger samples while bounding the decision boundary shift.}

\textbf{Trigger samples for dataset ownership protection}. 
\par
Dataset ownership protection raises new requirements for trigger sample generation: First, the injection of trigger samples should not introduce additional security risks for models trained on protected datasets. Second, the injected trigger samples should have stealthiness and cannot be detected by the dataset users. Finally, the models trained on the protected dataset should contain the IP message of the dataset owner. 
However, targeted backdoor watermarking injects poison backdoors, \ie trigger samples with specific target labels, into the original model. Attackers can add the trigger pattern to an arbitrary input sample to deterministically manipulate the model prediction. 

To this end, Li \myetal propose Untargeted Backdoor Watermarks (UBW)~\cite{black_li2022untargetedbackdoorwatermark}, a harmless and stealthy watermarking method where watermarked models output randomly predicted labels with low confidences on trigger samples instead of specified target labels. UBW considers the three demands for dataset protection and can be naturally extended to model ownership protection. A natural thought is randomly shuffling the true labels of trigger samples (UBW-P). In this way, an arbitrary sample added with the trigger pattern would produce random predictions that cannot be manipulated deterministically. 
\nop{Here, each trigger sample $\xbm_w$ is obtained by adding a predefined trigger pattern $\tbm$ via a binary mask matrix $\Gamma$ as $\xbm_w\!\!=\!\!(1\!\!-\!\!\Gamma)\!\odot\!\xbm\!\!+\!\!\Gamma\!\odot\!\tbm$.}
Moreover, to enhance the stealthiness of trigger samples injected in the protected dataset (poisoned samples), 
\cite{black_li2022untargetedbackdoorwatermark} proposes untargeted backdoor watermarks with clean labels (UBW-C). UBW-C maximizes the prediction uncertainty of trigger samples and uses a generator network with the parameters $\omegabm$ to generate clean-label poisoned samples. The model gradients of clean-label poisoned samples and modified-label trigger samples should be matched. 
With no change of labels, attackers cannot detect poisoned samples from the protected dataset by label analysis. Note that the poisoned samples will be injected into the protected dataset while the trigger samples $\X_w$ used for IP verification will not appear in the protected dataset.  The poisoned samples $\X_p$ are generated from some seed samples $\X_0$ via a bi-level optimization as
\begin{equation*}
    \begin{split}
        &\max_{\omegabm} {\nabla_{\thetabm^*}\Lcal_i\cdot \nabla_{\thetabm^*}\Lcal_t}/{(\|\Lcal_i\|\!\cdot\!\|\Lcal_t\|)},\\
        \textrm{s.t.}\ \ &\Lcal_i = \Lcal_{\rm CE}(\phi(\X_w;\thetabm^*),\Ybf_w) + \lambda\!\cdot\! \textrm{SE}(\phi(\X_w;\thetabm^*)), \\
        &\Lcal_t = \Lcal_{\rm CE}(\phi(\X_p;\thetabm^*),\Ybf_0),\ \X_p\!=\!\{\xbm\!+\!\omegabm(\xbm)|\forall \xbm\!\in\!\X_0\},\\
        &\thetabm^* = {\arg\min}_{\thetabm}\Lcal_{\rm CE}(\phi(\X\cup\X_p;\thetabm),\Ybf\cup\Ybf_0); 
    \end{split}
\end{equation*}
where $\Lcal_i$ and $\Lcal_t$ are the loss functions for trigger samples and poisoned samples respectively. $SE$ denotes the Shannon Entropy in Equ.~(\ref{eq:se}). The objective is to maximize the gradient similarity between $\Lcal_i$ and $\Lcal_t$ and the prediction uncertainty of $\X_w$.






 \begin{figure*}[htbp]
	\centering
	\includegraphics[width=.93\textwidth]{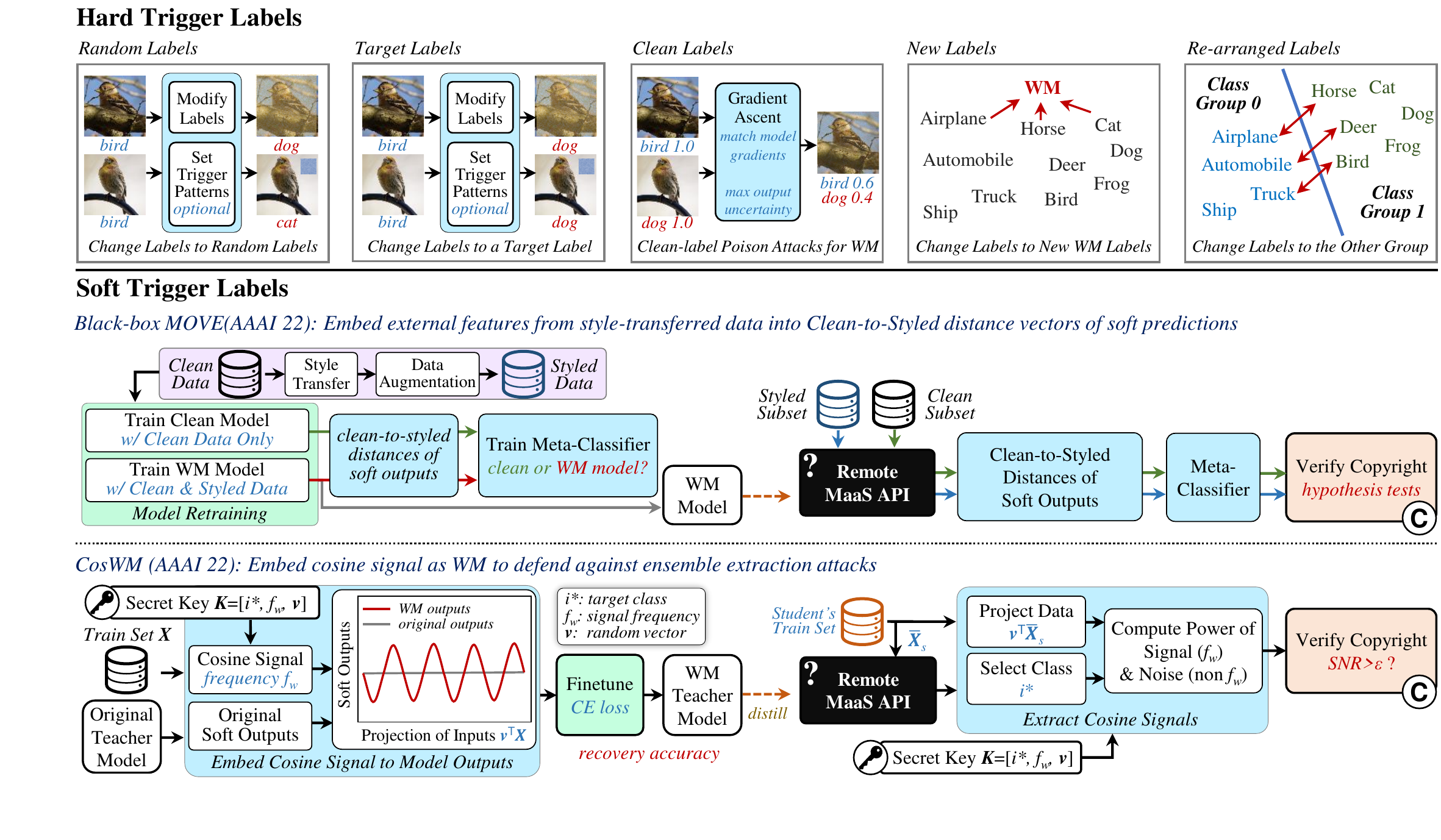}
  \vspace{-0.1in}
	\caption{\textcolor{black}{The representative methods for trigger sample labelling.}}
	\label{fig:trigger_sample_labelling}
  \vspace{-0.2in}
\end{figure*}
\subsubsection{Trigger Sample Labelling}\label{sec:Protect_Invasive_Trigger_Label}
This step aims to design the form of IP messages, \ie the labels of trigger samples to train watermark models, which can be hard or soft labels as shown in Fig.~\ref{fig:trigger_sample_labelling}. 
\par
\textbf{Most trigger-injected watermarking methods apply hard trigger labels as IP messages.}
It is a common way to use \textit{target hard labels}. However, it introduces some extra security risks by which the adversaries can manipulate the model predictions on arbitrary samples through trigger patterns. Applying \textit{random hard labels}~\cite{black_li2022untargetedbackdoorwatermark,black_adi2018turning} can alleviate this problem. In this way, the adversary can only change the prediction result randomly rather than deterministically. 

\par\textbf{A new label}. Zhong \myetal~\cite{black_zhong2020newlabel} propose to assign a new label to trigger samples, which can minimize the change of the model decision boundary when embedding trigger samples and classify trigger samples more accurately. However, the watermark can be easily ignored if the adversary knows the index of the new class. 

\par\textbf{Rearrange hard predictions to embed multi-bit IP messages. } Based on the matching rate of model predictions on trigger samples, we can only determine whether a suspect model is a pirated version of the watermarked model, and cannot embed/extract multi-bit IP messages (\emph{e.g.}, a bit string) like inner-component watermarking. Some works arrange trigger samples in specific orders and hide multi-bit IP messages into the sequence of the hard predictions.  
\begin{itemize}[leftmargin=*]
	\setlength{\topsep}{0pt}
	\setlength{\itemsep}{0pt}
	\setlength{\parsep}{0pt}
	\setlength{\parskip}{0pt} 
  \item  Blackmark~\cite{black_chen2019blackmarks} uses the k-means method to cluster all classes into two groups according to logit activations of original models, where the two groups represent the bit "0" and "1" respectively. Then, Blackmark randomly selects samples from one group and uses targeted adversarial attacks to move the predictions to the other group. A binary classification loss for class groups is applied as the watermark regularizer $\Lcal_R$, so that the predictions of trigger samples are close to the assigned groups (\ie minimizing the bit error rate). Moreover, some extra unwatermarked models are constructed to limit the transferability of adversarial samples, \ie making them non-transferable to irrelevant models. 
  \item AIME~\cite{black_mehta2022aime} converts inaccurate model predictions into unique IP messages. The trigger samples are sorted according to the confusion matrix where each element, $c_{ij}$, represents the number of the $i$th-class samples that are misclassified as the $j$th class. The watermark is embedded within misprediction results, while the IP message is extracted by querying the model using ordered trigger samples and then decoding the sequence of model predictions. AIME is robust against IP detection attacks because model mispredictions rely on the model's inherent nature. 
\end{itemize}
\nop{Blackmarks also relies on adversarial examples, similar to Frontier-Stitching.  The authors propose a pre-processing step that clusters all class labels into two groups using k-means clustering on the pre-trained source model’s logit activations.  These clusters will be used to encode bits.  The idea is to randomly select images from one cluster and use a targeted adversarial attack so that the source model predicts any class from the other cluster.  During embedding, an additional loss term is introduced that minimizes the bit error rate between the predicted cluster and the assigned cluster of the trigger.  The authors also present a method to mitigate the unintended transferability of the watermarking key images.  An adversarial example is transferable if it is adversarial to many models, i.e., it is not only adversarial to the source model for which it has been generated.}

\textbf{Applying soft model predictions of trigger samples as IP messages helps embed more watermarked messages.}
\par
\textit{(a) Clean-to-Styled Distance of Soft Model Predictions}. 
MOVE for black-box verification~\cite{both_li2022move} leverage the distance vector of model predictions between augmented style-transferred and clean training samples as IP messages. 
it uses the distance vectors as inputs to train a meta-classifier to distinguish the clean and watermarked models. Compared with the watermarking methods based on hard trigger labels, MOVE has stronger model fidelity and IP robustness against IP removal attacks while enhancing the generalization ability of original models.
\par
\textit{(b) Embedding Cosine Signals as Watermarks against Ensemble Distillation}. DNN watermarking by comparing hard predictions of trigger samples are vulnerable to ensemble distillation attacks that extract pirated models by combining the outputs from multiple teacher models. To overcome this problem, charette \myetal propose CosWM~\cite{black_charette2022cosine} that embeds cosine signals with some preset frequencies into the preset-class soft predictions of multiple teacher models, instead of directly using hard labels. 
The insight is that, after adding or multiplying signals with different frequencies, each signal's frequency can still be obtained from the signal spectrum of the added/multiplied signal, which is insensitive to amplitude changes. Here, the secret key consists of the cosine signal's frequency $f_w$, the preset host class $i^*$, and a random unit projection vector $\vbm$ to transform high-dimensional samples $\X$ into $1d$-values $\vbm^\top\X$. Firstly, CosWM constructs the cosine signal $\abm_i(\vbm^\top\X)$ as 
\vspace{-0.06in}
\begin{equation*} 
\abm_i(\vbm^\top\X) = \left\{
\begin{aligned}
&\cos(f_w\vbm^\top\X),& & \ \textrm{if}\ \ i=i*; \\
&\cos(f_w\vbm^\top\X+\pi),& &\ \textrm{otherwise}.
\end{aligned}
\right.
    \vspace{-0.06in}
\end{equation*}
The cosine signal $\abm_i$ is added to original soft outputs $\phi_i(\X)$ to create watermarked outputs $\qbm$ as trigger labels. For copyright verification, CosWM samples a subset $\Bar{\X}$ and computes the Signal Noise Ratio (SNR) of the signal $\abm(\vbm^\top\Bar{\X})$; then a suspect model can be considered piracy if the SNR is greater than a threshold.
 \begin{figure}[htbp]
 \vspace{-0.1in}
	\centering
	\includegraphics[width=.49\textwidth]{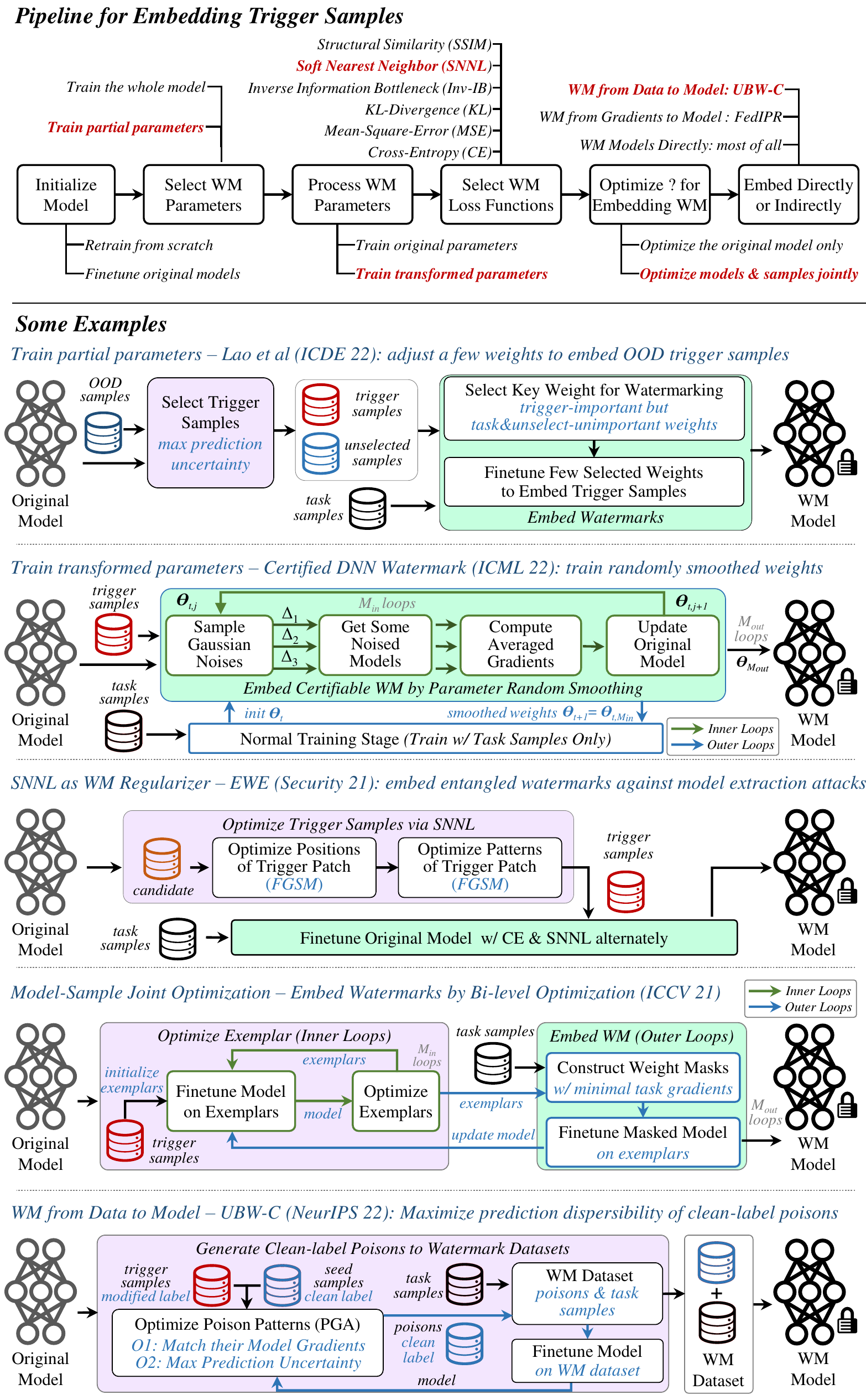}
  \vspace{-0.25in}
	\caption{\textcolor{black}{The pipeline and the representative methods for trigger sample embedding.}}
	\label{fig:trigger_sample_embedding}
  \vspace{-0.1in}
\end{figure}
\subsubsection{Embedding paradigms} \label{sec:Protect_Invasive_Trigger_Embedding} 
As shown in Fig.~\ref{fig:trigger_sample_embedding}, we summarize the pipeline of embedding trigger samples as the following steps: model initialization, trainable parameter selection, host parameter processing,  watermark regularizer selection and model training. 
\par
\textbf{Retraining or finetuning}. 
In general, retraining is better than finetuning in the robustness of trigger samples. However, retraining costs more computation resources for watermark embedding. All watermarking schemes support the retraining way, while only a part of the schemes support the finetuning way. Moreover, both retraining and finetuning need to be performed on the union of trigger samples and normal training datasets to avoid the degradation of model performance caused by catastrophic forgetting. 
\par
\textbf{Training all or partial model parameters}. Most trigger-injected watermarking schemes need to modify all model parameters to embed trigger samples. However, it would damage the prediction accuracy and change the response to certain natural samples considerably. The opposite is to finetune a few parameters that are sensitive to trigger samples but insensitive to task samples. 
Inspired by DNN fault attacks, Lao \myetal propose to embed trigger samples by adjusting a few weights for better model fidelity \cite{black_lao2022fewweights}.  First, \cite{black_lao2022fewweights} selects highly uncertain samples with maximal Shannon Entropy  from out-of-distribution data as trigger samples $\X_w$: 
\vspace{-0.06in}
\begin{equation}\label{eq:se}
    \textrm{SE}(\phi(\xbm;\thetabm)) = -{\sum}_{{i\in\{1...m\}}}\phi_i(\xbm;\thetabm)\log(\phi_i(\xbm;\thetabm)). 
    \vspace{-0.06in}
\end{equation}
These trigger samples would be easy to change the predictions and slightly impact model fidelity. 
Then, \cite{black_lao2022fewweights} selects the weights with maximal magnitudes of model gradients for $\X_w$ and minimal gradient magnitudes for the unselected samples $\X_c$ and the task samples $\X_n$, as trainable weights. 
\nop{; then it embeds watermarks, whose $t$-step is formulated as
\vspace{-0.06in}
\begin{equation*}
\begin{split}
    \thetabm^{t+1}_l \leftarrow & \thetabm^{t}_l - \alpha \Gamma_l\odot\nabla_{\thetabm^t_l}\Lcal(\phi(\X_w;\thetabm^t),\ybm)
    \\[-2pt]
    \textrm{s.t. } [\Gamma_l]_i &= 
    \begin{cases}
    1, &\textrm{if}\ i\in C_l^t;\\
    0, &\textrm{otherwise};
    \end{cases}\\[-2pt]
    C_l^t  &= \textrm{Top}_{N_w}\left\{ | \nabla_{\thetabm^t_l}\Lcal_{\rm CE}(\phi(\X_w;\thetabm^t),\ybm_w) |\right\}\\[-2pt]
    &\bigcap \textrm{Top}_{N_c}\left\{-| \nabla_{\thetabm^t_l}\Lcal_{\rm CE}(\phi(\X_c;\thetabm^t),\ybm_c) |\right\}\\[-2pt]
    &\bigcap \textrm{Top}_{N_n}\left\{-| \nabla_{\thetabm^t_l}\Lcal_{\rm CE}(\phi(\X_n;\thetabm^t),\ybm_n) |\right\},\\[-4pt]
\end{split}
\end{equation*}
where $\Gamma_l$ is a binary mask matrix for the parameter $\thetabm_l$ of the $l$-layer and $\textrm{Top}_N$ returns a set of $N$ selected weights.
Moreover, these methods can be naturally extended to user management that identifies different users. 
}
\par
\textbf{Training original or transformed weights}. Most methods train original weights for watermark embedding while some works explore training transformed weights for better robustness. 

\textit{(a) Exponential weighting.} The query modification attack (a way of IP detection \& evasion attacks) is a considerable threat to trigger-injected watermarks. Trigger samples following the distribution of training samples can defend against IP detection. However, this way relies on model overfitting, which makes the watermarks extremely vulnerable to model modification attacks.
Namba \myetal propose a robust watermark with exponential weighting~\cite{black_namba2019robust}. The intuition is that, in watermark-embedding epochs, \cite{black_namba2019robust} increases the gradients of the weights with large absolute values by exponentially weighting all the trainable model weights. In this way, it is harder for model modification to change the model predictions of trigger samples.

\textit{(b) Randomly-smoothed weights.} Most trigger-based watermarks claim the robustness to IP removal attacks, however, such robustness has no theoretical guarantee. Thus, Bansal \myetal propose a certifiable DNN watermarking method inspired by randomized smoothing, a technique for certified adversarial robustness that defends against all adversarial attacks under the $l_2$-norm ball constraint~\cite{black_bansal2022certified}. As shown in Fig.~\ref{fig:trigger_sample_embedding}, it updates the model parameters using the training and trigger samples alternately. At the watermark embedding stage, it samples $k$ gaussian noises $\{\Delta_i|i\!=\!1...k\}$ in the weight space and creates $k$ noised models $\{\thetabm_t\!+\!\Delta_i|i\!=\!1...k\}$. Then, their model gradients on trigger samples are computed separately and averaged for updating the weights of the original model.
Theoretical analysis and experiments show that it can defend against any $l_2$-bounded adversary that moves weights within a certain $l_2$-norm ball. 

\textbf{Choices of loss functions}. Most schemes apply the task loss function as the watermark regularizer, such as cross-entropy loss for classification, reconstruction loss for generation, MSE loss for regression, etc. Some advanced works explore more loss functions for watermark embedding. SNNL loss (Equ.~(\ref{eq:SNNL_loss})) is applied in \cite{black_jia2021entangled} and \cite{new_wu2022cits}. The KL-divergence loss and the information bottleneck loss are also applied for better robustness or more IP functions.  
\par
\textit{(a) Loss functions for better robustness}. 
\par
\textit{Entangled Watermarking Embeddings (EWE)}. Aiming to address the issue of trigger samples being easily removed by model extraction, Jia \myetal conduct an analysis and propose EWE~\cite{black_jia2021entangled}. Since the task and trigger samples are learned by different neurons, finetuning the watermarked model on task samples greatly changes the trigger-related neurons. EWE ensures that both the task and trigger samples are learned by the same neuron group, based on soft nearest neighbors as 
\vspace{-0.06in}
\begin{equation}\label{eq:SNNL_loss} 
    \Lcal^l_{\textrm{SNN}}\! =\! -\!\mathop{\rm mean}_{\xbm_i\in\Xbf}\{\log(\frac{\sum_{j\neq i,y_i=y_j}e^{-\|f^l(\xbm_i)-f^l(\xbm_j)\|^2/T}}{\sum_{k\neq i}e^{-\|f^l(\xbm_i)-f^l(\xbm_k)\|^2/T}})\},
    \vspace{-0.06in}
\end{equation}
where $T$ denotes the temperature generally decreasing with the number of iterations. The SNNL depicts the entanglement of data manifolds, and maximizing $\Lcal_{\textrm{SNN}}$ is equivalent to making the representation distances between samples in different classes close to the average representation distance of all samples. Based on $\Lcal_{\textrm{SNN}}$, EWE first optimizes the position and pattern of trigger patches on trigger samples by maximizing gradients.
Then, EWE embeds the trigger samples using the loss as Equ.~(\ref{eq:EWE_loss}). Here, EWE computes $\Lcal_{\textrm{SNN}}$ on each layer $f_l$ of the original model $\phi$. 
\vspace{-0.06in}
\begin{equation}\label{eq:EWE_loss}
    \Lcal = \Lcal_{\rm CE}(\X,\Ybf) - \alpha{\sum}_{l}\Lcal^l_{\rm SNN} (\X\cup\X_w,\Ybf\cup\Ybf_w).
     \vspace{-0.06in}
\end{equation}
\par
\textit{Target-specified Non-Transferable Learning (NTL)}~\cite{black_wang2022ntl}. 
The core idea is to limit the generalization ability of watermarked models, \ie degrading the model accuracy on the preset target domain (trigger samples) via an Inverse Information Bottleneck (InvIB) loss as  
\vspace{-0.06in}
\begin{equation*} \label{eq:lib}
    \Lcal_\textrm{IIB} = \textrm{KL}(\phi(\X),\Ybf)- \min\{\beta, \alpha\cdot \textrm{KL}(\phi(\X_w), \Ybf)\cdot\Lcal_\textrm{MMD}\}, 
      \vspace{-0.06in}
\end{equation*} 
where $\X_w$ denotes the samples in the target domain, by adding a trigger pattern into the source-domain samples $\X$. 
The InvIB loss forces the model to extract nuisance-dependent representations, and conversely, the original IB loss is to minimize the impact of nuisances.
To make the feature extractor in the model $\phi$ more sensitive to $\Lcal_\textrm{IIB}$, the authors add a product term $\Lcal_{\rm MMD}$ to maximize the Maximum Mean Discrepancy (MMD) of data representations between the source and target domains.
The target-specified NTL is a robust scheme for copyright verification by comparing the model accuracy on the target domain.
\begin{figure}[htbp]
	\centering
	\includegraphics[width=.47\textwidth]{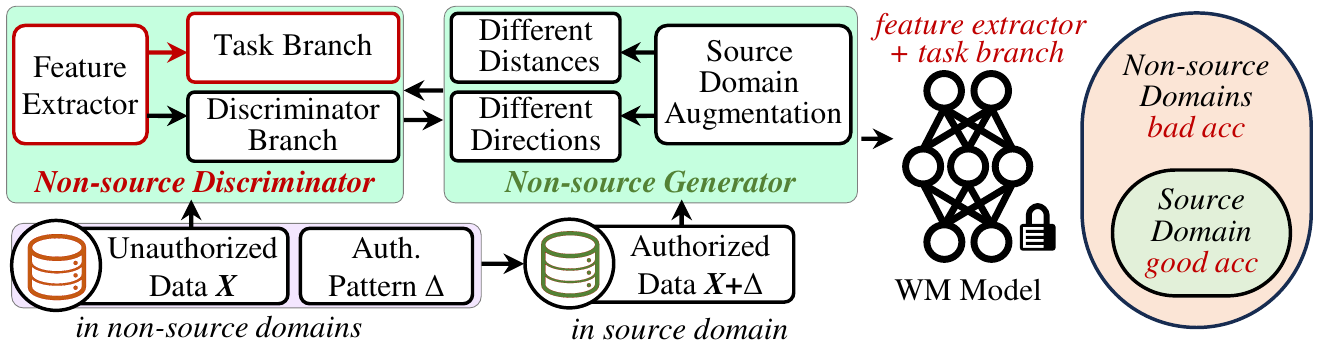}
  \vspace{-0.1in}
	\caption{\textcolor{black}{Source-Only NTL (ICLR 2022)~\cite{black_wang2022ntl}: Applicability Authorization to degrade non-source performance using a GAN-based data augmentation framework.}}
	\label{fig:source_only_ntl}
  \vspace{-0.1in}
\end{figure}

\textit{(b) Loss functions for applicability authorization}. 
\par
Besides copyright verification, 
NTL proposes applicability authorization, a novel function to invalidate the predictions of original models on unauthorized data.
On the basis of target-specified NTL, the source-only NTL makes the watermarked model valid only on the source domain. It designs a data augmentation framework to generate the non-source domain data by combining CGAN~\cite{gan_mirza2014cgan} and infoGAN~\cite{gan_chen2016infogan}. 
A modified version of the Inv-IB loss function $\Lcal_\textrm{IIB}$ is applied to enlarge the representation distance from the source domain to the non-source domains, as shown in Fig.~\ref{fig:source_only_ntl}.
\par
For a similar purpose, Ren \myetal propose M-LOCK~\cite{new_ren2022cybersecurity}, a model locking scheme where the watermarked model provides poor accuracy for the unauthorized data $\X_u$ without the trigger pattern and only make correct predictions on the authorized data $\X_\Delta$. M-LOCK trains the watermarked model using the authorized data $\X_\Delta$ with the correct labels $\Ybf$ and the unauthorized data $\X_u$ with the modified labels $\Ybf_u$, with the KL loss function as
\vspace{-0.06in}
\begin{equation*} 
        \Lcal=\textrm{KL}(\phi(\X_\Delta),\Ybf)+\alpha \cdot \textrm{KL}(\phi(\X_u),\Ybf_u), 
      \vspace{-0.06in}
\end{equation*} 
where $\Ybf_u$ could be the specific label, random uncertain labels or uniform probability vectors.

\textbf{Model-only or model-trigger joint optimization}. Jointly optimizing the original model and the trigger samples may enhance the IP robustness. 
Lao \myetal~\cite{black_yang2021robust} propose a DNN watermarking framework with better IP robustness based on bi-level optimization. As shown in Fig.~\ref{fig:trigger_sample_embedding}, it includes two alternative loops: i) the inner loops that iteratively generate the boundary exemplars ${\bm S}$ of trigger samples $\X_w$, \ie the “worst cases” of $\X_w$, by gradient ascent as 
\vspace{-0.06in}
\begin{equation*}\label{eq:inner_loop}
\begin{split}
    M_{in}&\ \textrm{steps}\\[-2pt]
    {\bm S}_{i+1}&\!\!\leftarrow\!\!{\bm S}_{i,M_{in}}
\end{split}\!\!\left\{
\begin{split}
    \thetabm_{i,j+1} &\!\leftarrow\! {\arg\min}_{\thetabm_{i,j}} \Lcal_{\rm CE}(\phi({\bm S}_{i,j};\thetabm_{i,j}),\Ybf_w),\\[-1pt]
    {\bm S}_{i,j+1} &\!\leftarrow\! {\bm S}_{i,j}\!+\!\beta \nabla_{\X_w} \Lcal_{\rm CE}(\phi(\X_w;\thetabm_{i,j+1}),\Ybf_w),\\[-1pt]
    \thetabm_{i,j} &\!\leftarrow\! \thetabm_{i,j+1},\ {\bm S}_{i,j} \leftarrow {\bm S}_{i,j+1};
\end{split}
\right.
    \vspace{-0.06in}
\end{equation*}
and ii) the outer loops that optimize model parameters $\thetabm$ using the generated examplers ${\bm S_{i+1}}$  to embed trigger samples $\X_w$ as 
\vspace{-0.06in}
\begin{equation*}\label{eq:outer_loop1}
    {\min}_{\thetabm_{i}} \alpha \Lcal_{\rm CE}(\phi(\X;\thetabm_{i}),\Ybf) + (1\!-\!\alpha) \Lcal_{\rm CE}(\phi({\bm S}_{i+1};\thetabm_{i}),\Ybf_w).
    \vspace{-0.06in}
\end{equation*}
Inspired by DNN fault attacks, \cite{black_yang2021robust} embeds the trigger samples by adjusting a few weights unimportant to task samples but important to trigger samples in the outer loops. The degree of the weight importance is measured by the amplitude of weight gradients.

\textbf{Embedding watermarks directly or indirectly}. 
Watermarks can be embedded by directly modifying the original model~\cite{black_adi2018turning} and most trigger-based watermarks apply this way. Moreover, watermarks can be embedded indirectly without direct access to the original model. 
Feasible ways include watermarking input data samples or their labels~\cite{black_li2022untargetedbackdoorwatermark, black_sun2022coprotector},  watermarking local gradients at clients of federated learning systems~\cite{fedwm_li2022fedipr,new_shao2022fedtracker,new_wu2022cits}, watermarking returned results of remote MLaaS APIs for user requests~\cite{black_szyller2019dawn}, etc.
\nop{Embedding backdoor trigger samples would introduce extra security risks into the original model. Fortunately, it is not necessary to modify the original model to protect remote MLaaS APIs from model extraction attacks.}

Szyller~\myetal propose DAWN~\cite{black_szyller2019dawn} that watermarks remote MLaas APIs by modifying the prediction responses of a tiny part of query samples (as trigger samples) rather than the original model. In this way, once the attackers steal the model of the remote MLaaS API, the API owner can verify the ownership by the selected trigger samples. DAWN requires that the prediction response from all clients must be the same for a query. Therefore, DAWN selects the query samples as trigger samples based on the query samples' hash codes. Then, DAWN applies the Fisher-Yates shuffle algorithm to modify prediction responses. It can not only determine the modified prediction response by a deterministic function related to the secret key and query samples' hash codes but also make the adversary cannot infer the true label through a large number of queries.

\subsection{Output-embedded Watermarking}\label{sec:Protect_Invasive_Output}
Watermarks can also be embedded/extracted into/from normal outputs generated by original models~\cite{none_zhang2021deep,none_wu2020watermarking,new_fei2022supervised,none_yu2021artificial,nlp_kirchenbauer2023watermark,nlp_he2022dawnnlp,nlp_xiang2021protectingNLG,nlp_abdelnabi2021awl,new_zhao2022distillationresistantwatermarkingNLP,new_zhao2023protecting,new_he2022cater}, which generally relies on predefined rules or trained extractor DNN models. It offers the following three advantages: 
i) The verification stage of output-embedded watermarking does not require specific triggers or white-box model access. It can even be performed in no-box scenarios where verifiers have no access to the original model and can only extract watermarks from suspect models' outputs spreading on the Internet.  
ii) Its application scenarios cover the generative processing tasks on across various domains like images, video, audio, text, etc. The typical tasks include but are not limited to image generation, style transfer, vision super-resolution, medical image processing, machine translation, code analysis, and question-answering systems. 
Such scenarios have already permeated every aspect of human life and will continue to play an increasingly crucial role in technical advancements
iii) The existing theories and techniques for traditional data watermarking can be applied to output-embedded watermarking, like various theoretically-completed signal processing methods or advanced deep learning methods. 
\begin{figure}[htbp]
	\centering
	\includegraphics[width=.47\textwidth]{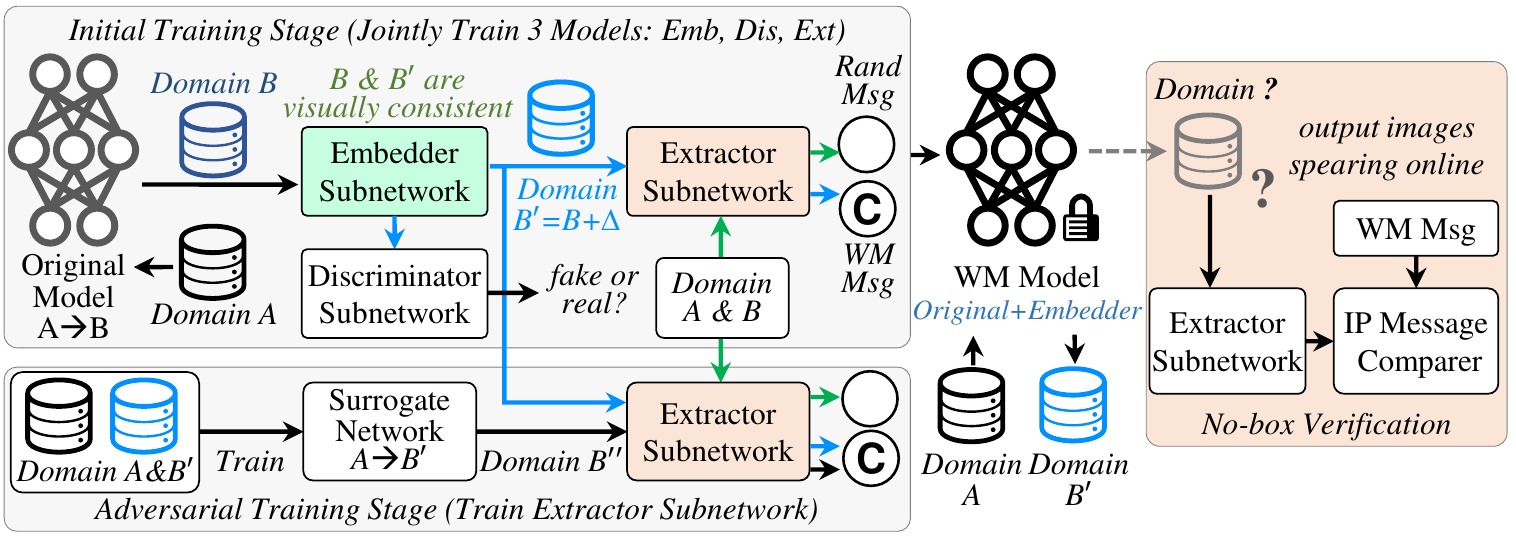}
 \vspace{-0.1in}
	\caption{A representative method of output-embedded DNN watermarking~\cite{none_zhang2021deep}.}
	\label{fig:output_ebmedded}
 \vspace{-0.2in}
\end{figure}

Literature~\cite{none_zhang2021deep} proposes a classical watermark scheme for low-level computer vision or image processing tasks. 
Unlike traditional image watermarking, \cite{none_zhang2021deep} adds a task-agnostic embedder subnetwork after the original model's outputs to embed watermarks into the generated outputs.
As shown in Fig.~\ref{fig:output_ebmedded}, considering an original image processing model from domain A to B, the embedder subnetwork is responsible for embedding a unified and invisible watermark into domain B and outputs the watermarked data domain B'. 
B' and B should be visually consistent, which is achieved by minimizing their feature-level and pixel-level differences and an extra discriminator subnetwork that distinguishes between domains B' and B. 
The embedder subnetwork is jointly trained with the extractor subnetwork. 
The extractor subnetwork outputs the watermark messages when given the watermarked domain B', and random messages otherwise.
To ensure its generalization ability on pirated versions of the original model, an adversarial training stage is designed in \cite{none_zhang2021deep} to train a surrogate network that mimics pirated models. The surrogate network transforms domain A to B'. Then, the extractor subnetwork is trained to output watermark messages when given domains B' and B''.

\textit{Output-embedded watermarking can be applied to detect illegal dataset usage.} 
Yu~\myetal~\cite{none_yu2021artificial} propose to embed watermarks into face datasets for rooting Deepfake attribution. Its intuition is that, the embedded watermarks in training samples have the transferability to generative models trained on these watermarked samples and will appear in the generated deepfakes. 
It first trains an encoder-decoder model $\{\Ecal, \Dcal\}$ to generate watermarked faces samples given an IP message $\bbm$, using the loss function as 
\vspace{-0.06in}
\begin{equation*}
    \min_{\Ecal,\Dcal} {\sum}_{\xbm\in\X} \left(\Lcal_{\rm CE}(\Dcal(\Ecal(\xbm,\bbm)),\bbm)+ \|\Ecal(\xbm,\bbm)-\xbm\|_2^2\right), 
    \vspace{-0.06in}
\end{equation*}
where the encoder $\Ecal$ inputs the original face $\xbm$ and the message $\bbm$ and outputs the watermarked face $\Ecal(\X,\bbm)$. Then, if a face generator is trained on the watermarked faces $\Ecal(\X,\bbm)$,  the generated deepfakes $G(\zbm)$ from the watermarked face generator $G$ will contain the IP message $\bbm$ that can be detected by the decoder $\Dcal$, \ie $\Dcal(G(\zbm))\rightarrow\bbm$.

\par\noindent\textbf{Remark: Combined Usage of DNN Watermarking}. 
Due to the strong memory capability of DNNs, it is feasible to integrate various watermarks into one original DNN model to meet the demands of multiple potential scenes.
The purpose of combined usage can be to adapt to various verification scenarios~\cite{white_rouhani2018deepsigns,black_jia2022srdw,passport_fan2021deepip,passport_ong2021iprgan,lottery_chen2021you} like white-box and black-box verification, or to endow the original model with various Deep IP functions, like copyright, integrity and access control~\cite{black_jia2022srdw,passport_fan2021deepip,lottery_chen2021you}. 
The potential conflict between different watermarks, and the impact of embedding multiple watermarks on the performance (\emph{e.g.}, model fidelity and watermark capacity), warrant further exploration. Szyller \myetal \cite{new_szyller2022conflicting} have built a framework to analyze conflicting interactions between multiple protection mechanisms to determine the effectiveness of combining protection mechanisms for different functions.
It is based on the performance constraints of each protection function and the original objective task.
On this basis, \cite{new_szyller2022conflicting} examines the interactions of pairwise conflicts and explores the feasibility of avoiding the conflicts. It covers four protection mechanisms: model ownership protection, data ownership protection, adversarial robustness, and differential privacy. 







\subsection{Watermarking for Specific Applied Scenarios}\label{sec:Protect_Invasive_Scenarios}
\subsubsection{Watermarking self-supervised learning encoders}  
\begin{figure}[htbp]
 \vspace{-0.1in}
	\centering
\includegraphics[width=.48\textwidth]{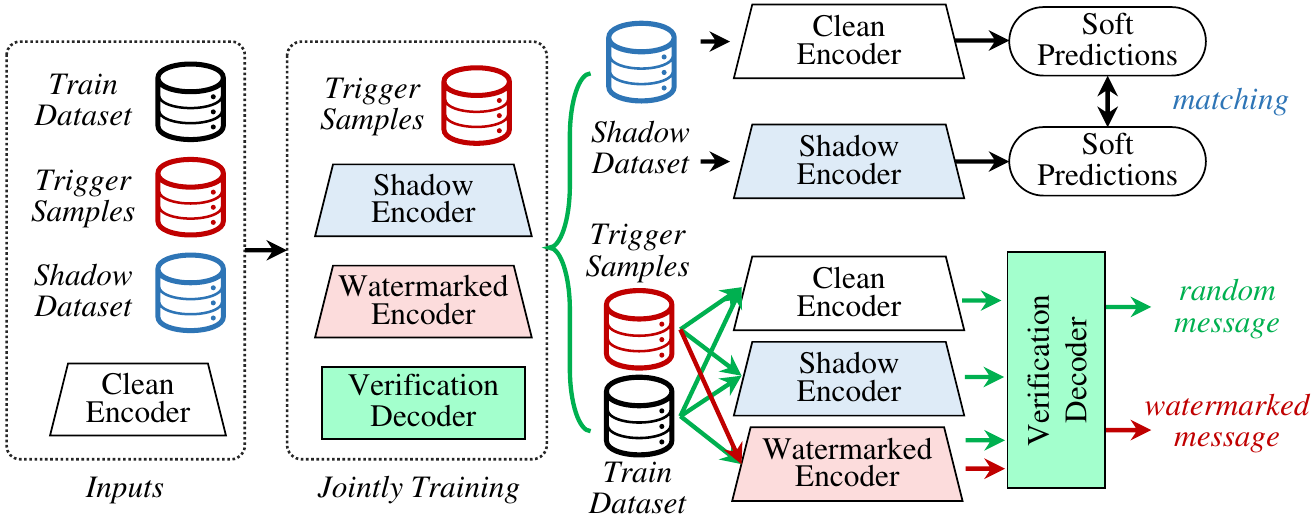}
 \vspace{-0.1in}
	\caption{\textcolor{black}{Watermarking self-supervised learning encoders~\cite{selfsuper_cong2022sslguard}.}}
	\label{fig:wm_ssl}
 \vspace{-0.1in}
\end{figure}

Self-supervised learning (SSL) models trained on unlabeled datasets have been utilized as feature extractors for various downstream tasks. Due to their widespread application scenarios and the huge training cost in terms of data and computation resources, SSL models have become valuable intellectual properties. However, they are facing serious threats from strong model stealing attacks~\cite{selfsuper_dziedzic2022difficulty}. 
To this end, some works focus on protecting SSL encoders~\cite{selfsuper_cong2022sslguard,new_zhang2022awencoder}. SSLGuard~\cite{selfsuper_cong2022sslguard} is the pioneering approach for watermarking SSL pre-trained encoders. It addresses the challenges of embedding watermarks in SSL encoders: i) unlike classification models, SSL encoders cannot directly access predicted labels of trigger samples; and ii) watermarks in SSL encoders should be preserved in various downstream task models. 
For the first challenge, SSLGuard embeds the watermark into the latent codes of trigger samples by finetuning SSL encoders and trains a verification decoder to extract IP messages. The decoder will extract the correct IP messages only when provided with the latent codes of trigger samples from the watermarked/pirated encoders; otherwise (given other samples' latent codes or clean/irrelevant encoders), random messages will be obtained. For the second challenge, SSLGuard introduces a trainable shadow encoder that simulates downstream-task encoders by matching the latent space between the shadow and the watermarked encoders. As shown in Fig.~\ref{fig:wm_ssl}, SSLGuard inputs the clean encoder, the training dataset, candidate trigger samples, and the shadow dataset, and then jointly trains the watermarked encoder, the shadow encoder, the trigger pattern, and the verification decoder.
\subsubsection{Watermarking against transfer learning}  
\begin{figure}[htbp]
\vspace{-0.1in}
	\centering
\includegraphics[width=.47\textwidth]{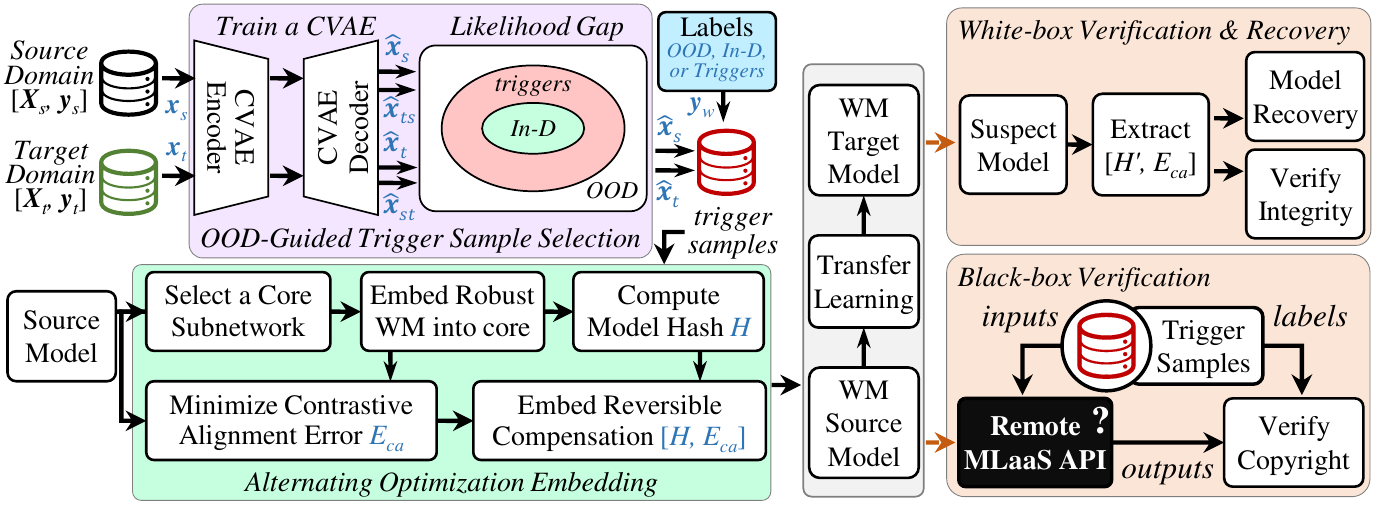}
 \vspace{-0.1in}
	\caption{\textcolor{black}{Watermarking against transfer learning~\cite{black_jia2022srdw}.}}
	\label{fig:wm_transfer}
 \vspace{-0.1in}
\end{figure}

Transfer learning is a deep learning technique that allows pre-trained DNN models to transfer source-domain features to a target domain. 
 It enables developers with limited data and computation resources to obtain high-performance models for new tasks without starting from scratch. 
However, the process of knowledge transfer, such as knowledge distillation or cross-domain distribution adaptation, would most likely corrupt the embedded watermarks in the source-domain model. From another view, preserving the watermarks will limit the generalization ability of the transferred model for the target task, resulting in inadequate knowledge transfer~\cite{black_jia2022srdw,ac_aprilpyone2021transfer}. 
To enhance the IP robustness against transfer learning, 
SRDW~\cite{black_jia2022srdw} proposes an out-of-distribution (OOD) guidance method to generate trigger samples and explore a core subnetwork that performs well on both task and trigger samples for watermark embedding, which is realized by an alternating optimization method based on module risk minimization, as shown in Fig.~\ref{fig:wm_transfer}. 
The intuition is that the models trained on a well-selected subnetwork will exhibit superior generalization performance on OOD samples. Thus, the watermarks will be more robust and covert if a reasonable degree of differentiation (DOD) between task and trigger samples is kept: trigger samples that are too far from the task domain are not covert while too close trigger samples are not robust and damage the model fidelity.
Moreover, for white-box integrity verification and high fidelity, SRDW achieves reversible watermarks by embedding the compensation information for model recovery along with the model hash code into the edge of the core subnetwork through lossless compression.
\subsubsection{Watermarking federated learning (FL) models} 
\begin{figure}[htbp]
	\centering
\includegraphics[width=.46\textwidth]{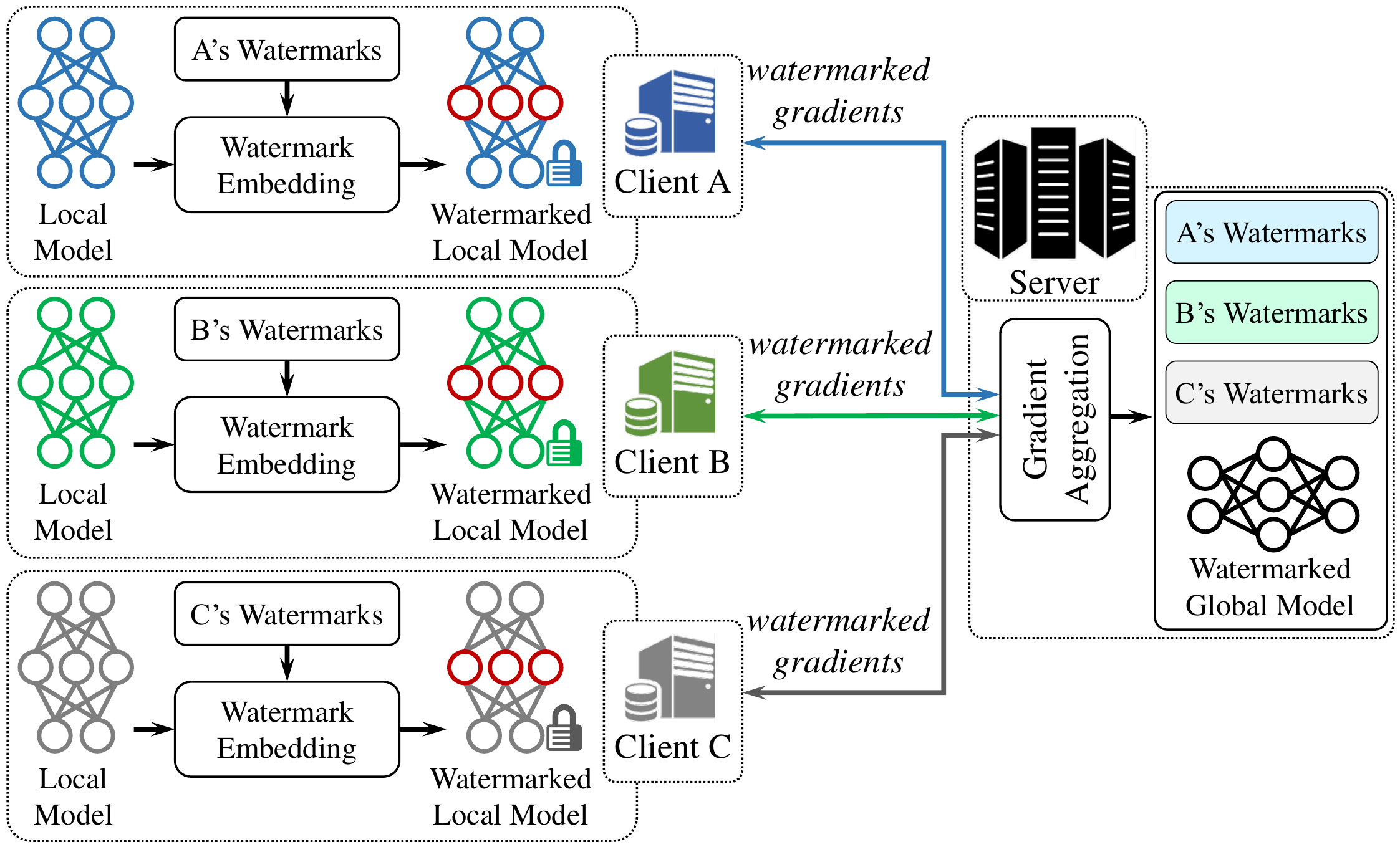}
 \vspace{-0.1in}
	\caption{\textcolor{black}{Watermarking in a federated learning system~\cite{fedwm_li2022fedipr}.}}
	\label{fig:wm_fedipr}
 \vspace{-0.1in}
\end{figure}

FL models, which are collaboratively trained by multi-parties with their own data and computation resources, are threatened by risks like illegal re-distribution or unauthorized/malicious uses during model training and deployment. 
In literature~\cite{federate_yang2023federated}, Yang \myetal analyzes the challenges and future research directions to protect FL-model IPs. 
It mainly has the two challenges: i) how to avoid the conflicts of multi-party watermarks; ii) how to ensure the watermark robustness to privacy-preserving federated learning strategies (like differential privacy, defensive aggregation, and client selection) and malicious attacks (like freerider attacks, fine-tuning/pruning attacks and trigger forgery attacks). 
Some works have tried to watermark FL models~\cite{fedwm_li2022fedipr,new_shao2022fedtracker,new_tekgul2021waffle,federate_yang2023federated,new_li2022watermarking,new_li2021regulating,new_liu2021secure}. 
WAFFLE~\cite{new_tekgul2021waffle} assumes that watermarks are embedded/verified on a trustworthy central server and clients are not owners of collaboratively-trained FL models. It finetunes the aggregated model at each aggregation round to embed backdoor watermarks.
In contrast to WAFFLE's hypothesis, Yang \myetal propose FedIPR~\cite{fedwm_li2022fedipr} that allows each party in an FL system to embed its own parameter-based~\cite{white_uchida2017embedding,white_chen2018deepmarks,white_rouhani2018deepsigns, passport_fan2019rethinking,passport_fan2021deepip} or backdoor-based~\cite{black_adi2018turning, black_le2020adversarial} watermarks, and independently verify the copyright of FL models, as shown in Fig.~\ref{fig:wm_fedipr}. 
It theoretically proves the lower bound $\eta_f$ of the detection rate of feature-based watermarks:
\vspace{-0.04in}
\begin{equation*}
\eta_f=\min\{(KN+M)/(2KN),1\}
\vspace{-0.04in}
\end{equation*}
where $K$ is the number of clients, $N$ is the bit number of each IP message, and $M$ is the number of watermarked parameters. FedIPR also gives the optimal and maximal bit-lengths of IP messages. 
On this basis, Yang \myetal propose FedTracker~\cite{new_shao2022fedtracker} to provide the trace evidence for illegal model re-distribution by unruly clients. It uses a combination of server-side global backdoor watermarks and client-specific local parameter watermarks, where the former is used to verify the ownership of the global model, and the latter is used to trace illegally re-distributed models back to unruly clients. 

CITS-MEW~\cite{new_wu2022cits} follows the assumption of FedIPR: each party in an FL system embeds its own watermark into the global model. It explores two problems: i) the locally embedded watermarks are easy to fail in the aggregation process; ii) the task performance of the global model will be seriously degraded when each party embeds its local watermarks. To this end, CITS-MEW proposes a multi-party entangled watermarking scheme that consists of local watermark enhancement and global entanglement aggregation. The former is to effectively enhance the strength of local watermarks, while the latter is to achieve the entanglement of all local watermarks and maintain the accuracy of the aggregated model. 

\subsubsection{Watermarking Generative Models}
\begin{figure}[htbp]
	\centering
	\includegraphics[width=.47\textwidth]{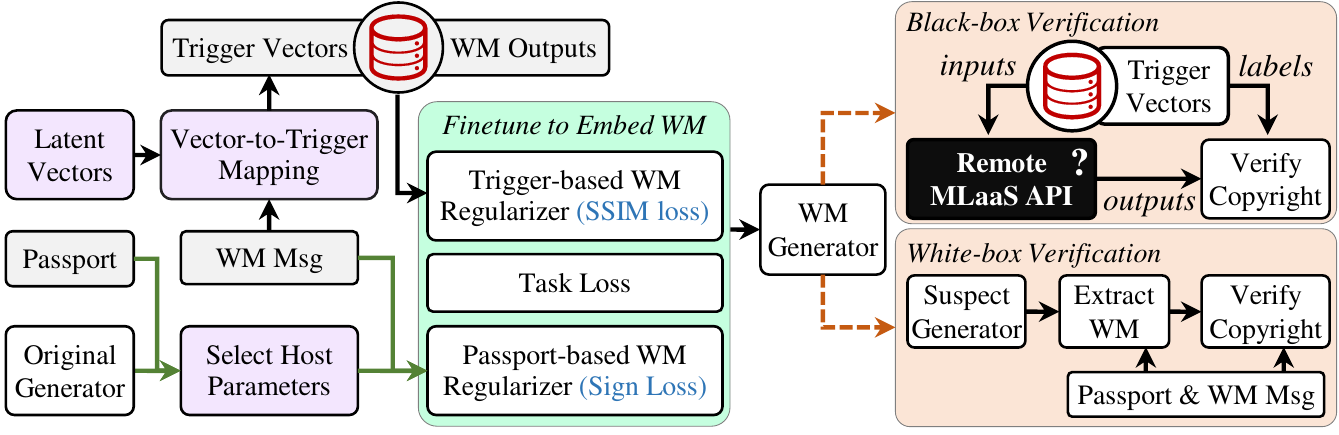}
  \vspace{-0.1in}
	\caption{\textcolor{black}{Watermarking generative models~\cite{passport_ong2021iprgan}.}}
	\label{fig:ganipr}
  \vspace{-0.1in}
\end{figure}
Output-embedded watermarking methods can be used to protect generative models like GANs.  However, since they inject watermarks into the generated images, the quality of generated images is more or less compromised. Some schemes~\cite{passport_ong2021iprgan,ac_qiao2023GANwatermark, new_ruan2023intellectual,black_quan2020watermarking} apply the idea of parameter-based or trigger-based watermarking. 
GAN-IPR~\cite{passport_ong2021iprgan} extents Deep IP Protection to Generative Adversarial Networks (GANs) for both white-box and black-box verification, as shown in Fig.~\ref{fig:ganipr}. With the former, it embeds the IP message $\bbm$ into the scale factors ${\bm \gamma}$ of normalization layers via the sign loss same as Equ.~(\ref{eq:sign_loss}). For the latter, it transforms the latent vector $\zbm$ (\ie the generator $G$'s input) to the trigger latent vector $\xbm_w$ as 
\vspace{-0.06in}
\begin{equation*}
    \xbm_w = \zbm\odot\bbm+c\cdot(1-\bbm).
    \vspace{-0.06in}
\end{equation*}
It masks some elements of $\zbm$ to a constant value $c$. Then, the reconstructive loss $\Lcal_w$ encourages the generator's output $G(\xbm_w)$ close to $\ybm_w$, the real image attached with an owner-defined patch.
\vspace{-0.06in}
\begin{equation*}
    \Lcal_w = 1- \textrm{SSIM}(G(\xbm_w),\ybm_w),
    \vspace{-0.06in}
\end{equation*}
where $\textrm{SSIM}$ is the structural similarity to measure the perceived quality between two images. It can defend against IP ambiguity attacks. 
\cite{ac_qiao2023GANwatermark} and \cite{new_ruan2023intellectual} follow a similar framework to \cite{passport_ong2021iprgan}. 
\cite{ac_qiao2023GANwatermark} selects some verification images and uses a secret key to construct watermark labels for these images. The watermarked generator is trained to output the verification images given the watermark labels concatenated with random vectors as inputs. And \cite{new_ruan2023intellectual} proposes to protect deep semantic segmentation models.

\subsubsection{Watermarking for AI applications} 
Most existing watermarking schemes focus on basic classification tasks while some try to watermark generative models or pretrained encoder models. However, deep models will yield economic benefits in medical, industrial, and various data-processing tasks. Watermarking also requires scalability for various AI applications, such as speech recognition systems~\cite{nlp_chen2020specmark}. 
\nop{SpecMark~\cite{nlp_chen2020specmark} is a Deep IP framework for Speech Recognition Systems. It embeds watermark messages into the spectrum of model parameters. } 
Sofiane \myetal~\cite{new_sofiane2021yes} extend trigger-based watermarks to machine learning models beyond classification tasks, including machine translation, regression, binary image classification and reinforcement learning. These tasks can be categorized into classification (\emph{e.g.}, machine translation, image classification, reinforcement learning in discrete action spaces) and regression (\emph{e.g.}, reinforcement learning in continuous action spaces). \cite{new_sofiane2021yes} applies a simple similarity comparison for verification and designs corresponding methods to select thresholds for the two types. It also proposes methods for trigger generation across vectors, images, and texts. 
\par
Sun~\myetal focus on the ethical and security risks of AI4coding models and propose CoProtector to protect open-source code repositories from unauthorized training usage~\cite{black_sun2022coprotector}. CoProtector allows deep models for code analysis to be trained on only code repositories authorized by announcements and adds some poison data into a part of unauthorized code repositories. Model developers who comply with the announcement can train clean and non-poisoned models on authorized code repositories. If model developers train their models on unauthorized code repositories, the performance of trained models will be significantly degraded and the models will be watermarked. For untargeted poisoning used to degrade model performance, CoProtector proposes four poisoning methods including code corrupting, code splicing, code renaming and comment semantic reverse. CoProtector explores three types of AI-for-coding tasks: code-only tasks like code completion, comment-to-code tasks like code generation, and code-to-comment tasks like code summarization; then, it proposes a unified method to construct trigger sets, \ie placing three unique tokens into an instance where two are from code and one is from comment. CoProtector designs a collaborative protection method to control the percentage of poison data in the current open-source community. A toolkit is also designed to help clients protect their code repositories. 

\section{Deep IP Protection: Non-Invasive Solutions}\label{sec:Protect_Non_Invasive}
\subsection{Overview}\label{sec:Protect_Non_Invasive_Overview}
\par
DNN Fingerprinting is a non-invasive way for Deep IP protection. It extracts the inherent properties of trained DNN models as IP messages, like decision boundaries and adversarial robustness, for comparing model similarities. DNN Models could be similar for a number of reasons, such as the train set, model structures, initial weights, the selected batch order during model iteration, selected optimizers, learning rates, etc. Due to the complexity of the search space for model weights and the stochasticity during model iteration, the similarity between a DNN model and its finetuned/extracted version is significantly higher than that between individually trained models even trained on the same dataset. This similarity margin can naturally be utilized to discriminate pirated models. As a non-invasive technique, fingerprinting achieves optimal model fidelity. Existing fingerprinting methods can be divided into two categories: 
\par
\textit{1) Model similarity can be ascertained through a comparison of model weights or their corresponding hash values, which serve as indicators of weight similarity.}
However, these methods require white-box access to suspect models for verification, and the suspect model must have the same structure as the protected model. Moreover, this way is vulnerable to slight weight perturbations due to the strong uninterpretability of model weights. 
\par
\textit{2) More works focus on constructing DNN fingerprints by analyzing model behaviors on preset test cases.} \textbf{Test cases} are typically adversarial samples or optimized by gradient descent/ascent from seed samples, which serve as inputs to a DNN model. One or multiple \textbf{test metrics} should be carefully designed to depict model behaviors.
 Given the \textit{case-metric} pair,  \textbf{fingerprint comparison} is used to determine whether the suspect model is a pirated copy of the protected model via similarity thresholds or meta-classifiers. 
\par
\begin{figure*}
	\centering
	\includegraphics[width=0.9\textwidth]{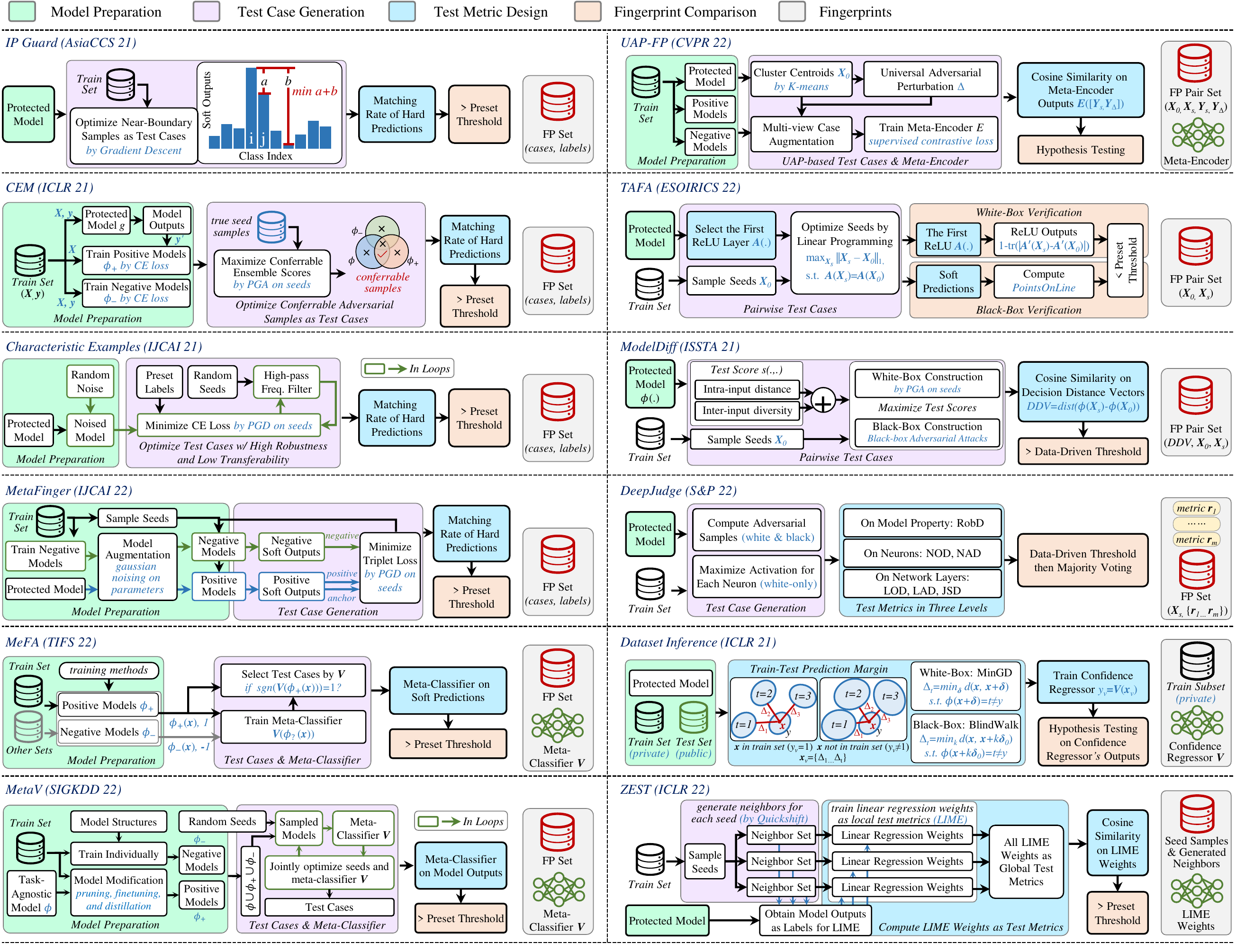}
\vspace{-0.1in}
 \caption{The Illustration of the representative methods for DNN fingerprinting.}
	\label{fig:fingerprinting}
 \vspace{-0.2in}
\end{figure*}

As shown in Table~\ref{tab:fingerprinting} and Fig.~\ref{fig:fingerprinting}, this section summarizes the representative fingerprinting methods from the following views: 
\begin{itemize}[leftmargin=*]
	\setlength{\topsep}{0pt}
	\setlength{\itemsep}{0pt}
	\setlength{\parsep}{0pt}
	\setlength{\parskip}{0pt} 
\item \textit{Data \& Model Preparation}: The prerequisite to construct model fingerprints (FP set in fig.~\ref{fig:fingerprinting}). Besides essential original models, \ie protected models, some positive models and negative models are often generated for better fingerprints, 
where the former is modified/extracted from the original model and the latter is individually trained. Some candidate test cases are often selected from training samples or randomly initialized. 
\item \textit{Test Cases}: A set of input samples exhibiting similar behaviors on the protected model and its pirated versions, yet displaying significant behavioral differences on negative models. 
\item \textit{Test Metrics}: The quantifiable measurements of model behaviors on test cases, such as model predictions and neuron activation. 
\item \textit{Fingerprint Comparison}: The process for verifying whether a suspect model is a pirated copy based on extracted fingerprints. It can be achieved by a simple similarity threshold or a meta-classifier trained on the fingerprints of prepared models, along with optional hypothesis testing. 
\item \textit{Construction Scenarios}: The model access required by fingerprint constructors, generally white-box. Some special scenarios may necessitate black-box or even no-box access for constructors. For example, models may be deployed via a model agent like Atrium\footnote{\url{https://atrium.ai/}} rather than model developers. Note that black-box methods can be directly used in white-box scenarios, and our emphasis here lies on whether a method has been designed for the scenario. 
\item \textit{Verification Scenarios}: The model access required by fingerprint verifiers, generally black-box. 
\item \textit{Target Models}: The types of DNN models for different tasks or learning paradigms, such as classification or generative tasks.
\item \textit{Target Functions}: The Deep IP functions aforementioned
in Section~\ref{sec:background_problem}, such as copyright or integrity verification. 
\end{itemize}

\begin{table*} 
\centering
  \caption{The Comparison of Representative Methods for DNN Fingerprinting}
  \label{tab:fingerprinting}
  \vspace{-0.1in}
    \setlength\tabcolsep{2pt}
    \scriptsize  
    \begin{tabular}{|c|c|c|c|c|c|c|c|c|c|}
    \toprule
    \multicolumn{2}{|c|}{\textbf{Methods}} & \textbf{Year} &\textbf{Test Cases}  & \textbf{Test Metrics} &\makecell{\bf Fingerprint\\[-1pt] \bf Comparison} & \makecell{\bf Construction\\[-1pt] \bf Scenarios} & \makecell{\bf Verification\\[-1pt] \bf Scenarios} &\textbf{Targets} &\makecell{\bf Prepared\\[-1pt] Models }\\
    \midrule
    \multirow{9}{*}{\rotatebox{90}{\bf Weight-based}}&PoL~\cite{fingerprinting_jia2021proof} & 2021 & Training Samples & Model Weights & Training Path Recovery & White-box & White-box & \makecell{Copyright Verification\\[-1pt] on DNN Models} & No\\
    \\[-2.5ex]\cline{2-10}\\[-2ex]
    &\makecell{Zheng \textit{et al}\\[-1pt]\cite{fingerprinting_zheng2022nonrepudiable}} & 2022 & - & \makecell{Random Projection on\\[-1pt]Front-layer Weights} & \makecell{Similarity Threshold\\[-1pt]on Random Projection} & White-box & White-box &\makecell{Non-Repudiable\\[-1pt]Copyright Verification} & No\\
    \\[-2.5ex]\cline{2-10}\\[-2ex]
    &\makecell{PIH \& TLH\\[-1pt]\cite{fingerprinting_xiong2022neural}} & 2022 & - & \makecell{Model Hash from\\[-1pt] Hash Generators} & \makecell{Similarity Threshold\\[-1pt] on Corresponding Bits} & White-box &White-box & \makecell{Tampering Localization\\[-1pt] \& CopyVer on DNNs} & Yes\\
    \\[-2.5ex]\cline{2-10}\\[-2ex]
    &\makecell{Chen~\myetal\\[-1pt]\cite{fingerprinting_chen2022perceptualhash}} & 2022 & - & \makecell{Model Hash on NTS\\[-1pt] of Critical Weights} & \makecell{A threshold on\\[-1pt ]Hamming Distances\\} & White-box &White-box & \makecell{Copyright Verification\\[-1pt] on DNN Models} & Yes\\
    \midrule
    \multirow{48}{*}{\rotatebox{90}{\bf Behavior-based}}&IP Guard~\cite{fingerprinting_cao2021ipguard}                     & 2019 & Near-Boundary Samples & \makecell{Matching Rate on\\[-1pt] Hard Predictions}    & \makecell{A Preset Threshold\\[-1pt]  on Matching Rate}  & White-box & Black-box & \makecell{Copyright Verification\\[-1pt] on Classification Models} & No\\
    \\[-2.5ex]\cline{2-10}\\[-2ex]
    &CEM~\cite{fingerprinting_lukas2019deep}                           & 2019 & \makecell{Conferrable Samples\\[-1pt] from an ensemble model}   & \makecell{Matching Rate on\\[-1pt] Hard Predictions}    & \makecell{A Preset Threshold \\[-1pt] on Matching Rate}  & White-box & Black-box & \makecell{Copyright Verification\\[-1pt] on Classification Models} & Yes\\
   \\[-2.5ex]\cline{2-10}\\[-2ex]
    &AFA~\cite{fingerprinting_zhao2020afa}                             & 2020 & Adversarial Marks\nop{(from target labels to target logits)} & \makecell{Matching Rate on\\[-1pt] Hard Predictions} & \makecell{A Threshold on MR\\[-1pt] from Ghost Networks\nop{~\cite{modelprepare_li2020ghostnetworks}Positive Models}} &White-box &Black-box & \makecell{Copyright Verification\\[-1pt] on Classification Models} & Yes\\ 
    \\[-2.5ex]\cline{2-10}\\[-2ex]
    &\makecell{Characteristic\\[-1pt]Examples~\cite{fingerprinting_wang2021characteristic}} & 2021 & \makecell{Low-Transferability \\[-1pt] Robust C-Examples} & \makecell{Matching Rate on\\[-1pt] Hard Predictions} & \makecell{A Preset Threshold \\[-1pt] on Matching Rate} & White-box &Black-box & \makecell{Copyright Verification\\[-1pt] on Classification Models} & Yes\\ 
    \\[-2.5ex]\cline{2-10}\\[-2ex]
    &\makecell{MetaFinger\\[-1pt]~\cite{fingerprinting_yang2022metafinger}} & 2022 & \makecell{Trigger Samples\\[-1pt] Optimized by Triplet Loss}  & \makecell{Matching Rate on\\[-1pt] Hard Predictions}  &  \makecell{A Preset Threshold \\[-1pt] on Matching Rate} & White-box & Black-box & \makecell{Copyright Verification\\[-1pt] on Classification Models} & Yes\\
    \\[-2.5ex]\cline{2-10}\\[-2ex]
    &MeFA~\cite{fingerprinting_liu2022mefa}                            & 2022 & \makecell{Conferrable Samples\\[-1pt] by fingerprint filtering }   & \makecell{Fingerprint Classifier\\[-1pt] on Soft Predictions}    & \makecell{A Preset Threshold on\\[-1pt]Fingerprint Classifier} & White-box & Black-box & \makecell{Copyright Verification\\[-1pt] on Classification Models} & Yes\\
    \\[-2.5ex]\cline{2-10}\\[-2ex]
    &MetaV~\cite{fingerprinting_pan2022metav}                          & 2022 & \makecell{Adaptive Fingerprints\\[-1pt] on model ensemble} & \makecell{Fingerprint Classifier \\[-1pt] on Model Outputs} & \makecell{A Preset Threshold on\\[-1pt]Fingerprint Classifier} & White-box & Black-box & \makecell{Copyright Verification\\[-1pt] on Task-Agnostic Models} & Yes\\
    \\[-2.5ex]\cline{2-10}\\[-2ex]
    &Modeldiff~\cite{fingerprinting_li2021modeldiff}                   & 2021 & \makecell{Adversarial Sample Pairs \\[-1pt] to maximize test scores} & \makecell{CosSim of Decision\\[-1pt] Distance Vectors (DDV)}& \makecell{Data-driven Thresholds\\[-1pt] on CosSim of DDVs} & White\&Black & Black-box & \makecell{Copyright Verification\\[-1pt] on Classification Models} & No\\
    \\[-2.5ex]\cline{2-10}\\[-2ex]
    &UAP-FP~\cite{fingerprinting_peng2022uapfp}               & 2022 & \makecell{Universal Adversarial\\[-1pt]  Perturbations}  & \makecell{Fingerprint Encoder \\[-1pt] on Soft Predictions}  & \makecell{A Fingerprint Encoder\\[-1pt]\& Two-Sample $t$-Test} & White-box & Black-box & \makecell{Copyright Verification\\[-1pt] on Classification Models} & Yes\\
    \\[-2.5ex]\cline{2-10}\\[-2ex]
    &TAFA~\cite{fingerprinting_pan2021tafa}                            & 2021 & \makecell{Samples Optimized by\\[-1pt] Linear Programming}& \makecell{First ReLU Activation \\[-1pt] \& PointsOnLine Appro. } & \makecell{A Preset Threshold\\[-1pt] on Designed Metrics} & White-box & White\&Black& \makecell{Copyright Verification\\[-1pt] on Task-Agnostic Model} & No\\
   \\[-2.5ex]\cline{2-10}\\[-2ex]
    &\makecell{DeepJudge\\[-1pt]~\cite{fingerprinting_chen2021copy}}                      & 2021 & \makecell{Adversarial Samples \\[-1pt] \& Maximizing Activation} & \makecell{Three-level Metrics:\\[-1pt] Property, Neuron, Layer} & \makecell{ Thresholds from $\varepsilon$-diff \\[-1pt] then Majority Voting} & White-box & White\&Black & \makecell{Copyright Verification\\[-1pt] on Classification Models} & No\\
    \\[-2.5ex]\cline{2-10}\\[-2ex]
    &\makecell{Dong \textit{et al}\\[-1pt]\cite{fingerprinting_dong2021dynamic}}  & 2021 & \makecell{Adversarial Samples\\[-1pt] to inference time} & Inference Time & \makecell{A Preset Threshold \\[-1pt] on EEC AUC scores} & White-box & White-box & \makecell{Copyright Verification\\[-1pt] on Dynamic Models} & No\\
    \\[-2.5ex]\cline{2-10}\\[-2ex]
     &\makecell{He \textit{et al}\\[-1pt]\cite{fingerprinting_he2019sensitive}}  & 2019 & \makecell{Sensitive Samples\\[-1pt] s.t. $\phi(\X_s)\neq\phi_\Delta(\X_s)$} & \makecell{Matching Rate on\\[-1pt] Hard Predictions} & \makecell{Equality Determination} & White-box & Black-box & \makecell{Integrity Verification\\[-1pt] on Classification Models} & No\\
   \\[-2.5ex]\cline{2-10}\\[-2ex]
    &\makecell{PublicCheck\\[-1pt]~\cite{fingerprinting_wang2022publicheck}}            & 2022 & Encysted Samples & \makecell{Matching Rate on\\[-1pt] Hard Predictions} & Equality Determination & Black-box & Black-box & \makecell{Integrity Verification\\[-1pt] on Classification Models} & No\\ 
    \\[-2.5ex]\cline{2-10}\\[-2ex]
    &\makecell{Intrinsic\\[-1pt]Examples~\cite{fingerprinting_wang2021intrinsic}}            & 2022 & \makecell{Adversarial Samples from\\[-1pt] a Min-Max Optimization} & \makecell{Matching Rate on\\[-1pt] Hard Predictions} & \makecell{A Preset Threshold \\[-1pt] on Matching Rate} & White-box & Black-box & \makecell{Functionality Verification\\[-1pt] on Classification Models} & No\\ 
       \\[-2.5ex]\cline{2-10}\\[-2ex]
    &\makecell{Deepfool-FP\\[-1pt]\cite{fingerprinting_wang2021deepfool}}          & 2021 & \makecell{Adversarial Samples by \\[-1pt] the Deepfool method} & \makecell{Matching Rate on\\[-1pt] Hard Predictions} & \makecell{A Confidence Threshold\\[-1pt] on Bayesian Probability} & White-box & Black-box & \makecell{Copyright Verification\\[-1pt] on Classification Models} & Yes\\ 
       \\[-2.5ex]\cline{2-10}\\[-2ex]
    &GAN-FP~\cite{fingerprinting_li2021ganfp}                          & 2021 & \makecell{Covert adversarial samples\\[-1pt] to \{GAN+Classifer\} Model} & \makecell{Fingerprint Classifier\\[-1pt] on Generative Outputs} & \makecell{A Preset Threshold on\\[-1pt]Fingerprint Classifier} & White-box & Black-box & \makecell{Copyright Verification\\[-1pt] on GAN Models} & No\\
        \\[-2.5ex]\cline{2-10}\\[-2ex]
    &SAC~\cite{fingperprinting_guan2022samplecorrelation}& 2022 &\makecell{Wrongly-classified Samples/\\[-1pt]CutMix-augmented Samples}  & \makecell{Correlation Matrices\\[-1pt] of Model Predictions} & \makecell{A Threshold on \\[-1pt] Metric Distances} & White-box & Black-box & \makecell{Copyright Verification\\[-1pt] on Classification Models} & \makecell{SAC-w:Y\\[-1pt]SAC-c:N}\\ 
       \\[-2.5ex]\cline{2-10}\\[-2ex]
    &DI~\cite{fingerprinting_maini2021datasetinference} & 2021 & \makecell{Verifier-private\\[-1pt]Training Samples} & \makecell{Train-Test Margin\\[-1pt] on Model Predictions} & \makecell{Hypothesis Testing on\\[-1pt] Confidence Regressor} & White\&Black & White\&Black & \makecell{Copyright Verification\\[-1pt] on Classification Models}& No\\
        \\[-2.5ex]\cline{2-10}\\[-2ex]
    &SSL-DI~\cite{fingerprinting_dziedzic2022DI4selfsupervise} & 2022 & \makecell{Verifier-private\\[-1pt]Training Samples} &\makecell{Data Density Margin on \\[-1pt] Representation Domain} &\makecell{Hypothesis Testing on\\[-1pt] GMM Density Estimator} & Black-box & Black-box &\makecell{Copyright Verification\\[-1pt] on Self-supervised Models}& No\\
        \\[-2.5ex]\cline{2-10}\\[-2ex]
    &\makecell{TeacherFP\\[-1pt]\cite{fingerprinting_chen2022teacherFP}} & 2022 & \makecell{Synthetic Samples by\\[-1pt] min representation distances} & \makecell{Three belief metrics\\[-1pt] using Hard Predictions} &\makecell{"One-of-the-Best" MR\\[-1pt] from Supporting Sets} &White-box &Black-box &\makecell{Teacher Identification\\[-1pt] for Transfer Learning} & No\\ 
        \\[-2.5ex]\cline{2-10}\\[-2ex]
    &ZEST~\cite{fingerprinting_jia2022Zest} & 2022 & \makecell{Random samples \&\\ Their Constructed Neighbors} & \makecell{Model Weights from\\[-1pt] LIME Approximation \nop{\\[-2pt] Linear Regression Models}} & \makecell{A Distance Threshold \\[-1pt] on LIME Weights} & White\&Black & Black-box & \makecell{Copyright Verification\\[-1pt] on Classification Models} & No\\
        \\[-2.5ex]\cline{2-10}\\[-2ex]
    &FPNet~\cite{new_jeong2022fingerprintnet} & 2022 & - & Spatial Spectrum & A Detector Network & No-box &No-box &\makecell{Generated Image Detection\\[-1pt] for Unseen Models}&No\\
  \bottomrule
\end{tabular}
\vspace{-0.15in}
\end{table*}

\subsection{Comparing Model Weights}\label{sec:Protect_Non_Invasive_Weights}
Due to its requirements on white-box models, as well as low robustness and model scalability, only a few works focus on this idea. There are several feasible approaches:
 \subsubsection{Recover the path of model training.} PoL~\cite{fingerprinting_jia2021proof} is the first weight-based fingerprinting method to our knowledge. It constructs evidence of a model's training by saving selected data batches and meta-data guiding model iteration.
 Correct evidence can recover fragments of the protected model's training process, while forged evidence cannot. This enables copyright verification by directed weight comparison in the recovery path.
 
 \subsubsection{Random projection of model weights.} Zheng \myetal~\cite{fingerprinting_zheng2022nonrepudiable} propose to extract model fingerprints from the front-layer weights using a random projection method similar to Locality Sensitive Hashing (LSH). 
 It enables the provision of non-repudiable and irrevocable ownership proof through a trusted third party.
 
 \subsubsection{Learnable Hash to represent model similarity.} Chen \myetal~\cite{fingerprinting_chen2022perceptualhash} design a perceptual hashing method for detecting copied CNN models. It applies model compression techniques to select critical weights and computes the fixed-length hashcodes of their normal test statistics (NTS). 
 Hamming distances of these hashcodes are then used to measure model similarity. 
Similarly, 
Xiong \myetal~\cite{fingerprinting_xiong2022neural} proposed two DNN-based methods for hashing model weights: Piracy Identification Hash (PIH) and Tampering Localization Hash (TLH). PIH obtains the model hash for copyright verification by inputting pruned weights (including convolution and full-connection layers) into a dual-branch neural network. Pirated models will have similar PIH outputs to the protected model, but individually trained models will not.
TLH is designed for tampering localization, an improved integrity verification to identify the location of tampered weights besides model tampering, by dividing the weights into multiple blocks and concatenating the PIH of these blocks. 
\subsection{Comparing Model Behaviors}\label{sec:Protect_Non_Invasive_Behaviors}
Model behaviors on specific input samples can be designed to reflect the unique model characteristics for Deep IP Protection. The workflow involves four components: Data \& Model Preparation, Test Case Generation, Test Metric Design, and Fingerprint Comparison. Next, we describe the representative schemes from component-wise decomposition and illustrate their required scenarios for IP construction and verification as well as the motivated target tasks.
\subsubsection{Data \& Model Preparation}\label{sec:Protect_Non_Invasive_Behaviors_Pre}
Besides the essential original models and their training datasets, extra models or data samples may be required for better fingerprints, which can aid in test case generation, test metric design, or fingerprint comparison. Generally, defenders create \textbf{positive models} to mimic pirated models by modifying the original models (\emph{e.g.}, parameter noising or fine-pruning), while \textbf{negative models} are created to mimic unrelated models trained individually. Extra data such as out-of-distribution data can be used to train negative models. 
\par
The scope of positive/negative models relies on the protection range of fingerprinting methods, which is determined by IP defenders. 
If models trained independently on the same dataset are considered independent models (with differences in at least one aspect of model structure, training strategy, initial weights, etc.), only the models extracted/modified from the protected model are positive models, and the rest are considered negative models. In this way, positive models are obtained by finetuning/extracting the protected model, for example, using the soft predictions of the protected model as labels to train positive models and using true labels of datasets to train negative models~\cite{fingerprinting_lukas2019deep}. If defenders want to protect datasets, that is, all models trained on the protected datasets are considered pirated models,  defenders can use different model structures and training algorithms to train positive models on the protected datasets, and use extra datasets to train negative models~\cite{fingerprinting_liu2022mefa}. 
\par
{Extra models can be created by model retraining}. 
The training configuration is determined by the defender based on his demands as discussed above. MetaV~\cite{fingerprinting_pan2022metav} uses some model modification methods (\emph{e.g.}, pruning, finetuning and distillation) with different hyperparameters (\emph{e.g.}, pruning ratios) to construct positive models; and the negative models come from three sources: \ding{172} training relatively small-scale DNNs on the same training dataset; \ding{173} finetuning public pretrained models from online sources on the same training dataset; \ding{174} some irrelevant public models.
However, such strategies will cost huge computation resources to train extra models. The solution is to leverage {model augmentation techniques without retraining to efficiently produce various extra models}, 
For example, MetaFinger~\cite{fingerprinting_yang2022metafinger} improves the diversity of prepared models by randomly injecting Gaussian Noises into model parameters. CEM~\cite{fingerprinting_lukas2019deep} and AFA~\cite{fingerprinting_zhao2020afa} leverage ghost networks~\cite{modelprepare_li2020ghostnetworks}, \ie the dropout versions of the protected model, as the positive models. 
%
\par
Model preparation can help test case generation. For instance,  
CEM~\cite{fingerprinting_lukas2019deep} combines the soft predictions of prepared models into an ensemble model to optimize conferrable adversarial samples that have transferability to positive models only but not negative models. 
Likewise, SAC-w~\cite{fingperprinting_guan2022samplecorrelation} selects samples wrongly classified by the protected model and the positive models but correctly classified by the negative models. 
MetaFinger~\cite{fingerprinting_yang2022metafinger} uses a triplet loss to generate test cases that have minimal distances of soft predictions between positive models and the protected model while having maximal distances between negative models and the protected model. 
\par
Model preparation is also applied for test case generation or fingerprint comparison. For example, AFA~\cite{fingerprinting_zhao2020afa} leverages prepared models to determine the similarity threshold for fingerprint comparison. Wang \myetal~\cite{fingerprinting_wang2021deepfool} use prepared models to compute the confidence level of model piracy based on the Bayesian Theorem.
\nop{Dataset Inference~\cite{fingerprinting_maini2021datasetinference, fingerprinting_dziedzic2022DI4selfsupervise} to train a confidence regressor for fingerprint comparison. 
public test samples
private train samples}
MetaV~\cite{fingerprinting_pan2022metav} and MeFA~\cite{fingerprinting_liu2022mefa} use the fingerprints of prepared models to train a meta-classifier to judge whether a given fingerprint comes from a positive/negative model. 
For test case generation, MeFA~\cite{fingerprinting_liu2022mefa} inputs some samples to the positive models cascaded with the meta-classifier, and
selects the samples with the positive output "1"; and MetaV~\cite{fingerprinting_pan2022metav} uses the prepared models to jointly train the test cases and classifier weights. 
UAP-FP~\cite{fingerprinting_peng2022uapfp} trains a fingerprint encoder using supervised contrastive loss to represent the fingerprint similarity. 
Moreover, model preparation is also applied in PIH\&TLH~\cite{fingerprinting_xiong2022neural}, a weight-based fingerprinting method, to train the hash generators. 

\subsubsection{Test Case Generation}\label{sec:Protect_Non_Invasive_Behaviors_Case}
As a critical component of DNN fingerprinting, many research works explore generating conferrable test cases. 
The word "conferrable"  has two meanings: The first is robustness. It refers that the protected model and its pirated versions should have similar model behaviors on the generated test cases. These cases need to defend various attack methods like IP detection, evasion, and removal described in Section \ref{sec:Attack}. The second is uniqueness, \ie suitable transferability. It refers that irrelevant models' behaviors should be distinguished from the protected model's behaviors on the generated test cases.  
\par 
\textbf{Gradient-based sample optimization to generate test cases}. 
It is the common way to generate test cases, like adversarial attacks or their variants. 
The typical optimization techniques include Fast Gradient Sign Method (FGSM)~\cite{adversarial_goodfellow2014FGSM}, Projected Gradient Descent (PGD)~\cite{adversarial_madry2018PGD}, Carlini \& Wagner (C\&W)~\cite{adversarial_carlini2017CW}, etc. 
 Note that test case generation is different from adversarial attacks on optimization objectives, constraints, etc. We will cover this briefly below.

\textit{Some basic examples}. 
IPGuard~\cite{fingerprinting_cao2021ipguard}, the first study for DNN fingerprinting, generates test cases near the decision boundary of the original model. The optimization problem can be formulated as
\vspace{-0.06in}
\begin{equation*}
        \min_{\xbm}{{\rm ReLU}(\phi_i(\xbm)\!-\!\phi_j(\xbm)+k)
        +{\rm ReLU}(\max_{t\neq i,j}\phi_t(\xbm)\!-\!\phi_i(\xbm))}
\vspace{-0.06in}
\end{equation*}
where $i$ and $j$ are random classes, ${\rm ReLU}(x) = max(0, x)$, $\phi_i$ denotes the class-$i$ output logits, $k$ is a hyperparameter to balance between robustness and uniqueness. 
On this basis,
AFA~\cite{fingerprinting_zhao2020afa} proposes adversarial marks, \ie the adversarial samples that 
mimic the logits and the predicted labels of the samples randomly chosen from target classes. 
Wang~\myetal~\cite{fingerprinting_wang2021deepfool} leverage the Deepfool~\cite{adversarial_moosavi2016deepfool} method to extract the adversarial samples with minimal perturbation by projecting seed samples onto the decision boundary.

\textit{Test cases with a suitable transferability}. 
The above methods only consider the decision boundary of the protected model $\phi$. However, the decision boundary of $\phi$ may partially coincide with irrelevant models but not pirated models.
Thus, CEM~\cite{fingerprinting_lukas2019deep} generate conferrable test cases via a conferrable ensemble model $\phi_E$ that is aggregated by some positive models $\Phi_+$ and negative models $\Phi_-$, defined as:
     \vspace{-0.06in}
\begin{equation*}
    \begin{split}
        \textrm{surr}(\Phi_+,\xbm)&=
        \textrm{mean}_{\phi_+\in\Phi_+}\{\textrm{dropout}(\phi_+(\xbm))\},\\[-1pt]
        \textrm{ref}(\Phi_-,\xbm)&=
        \textrm{mean}_{\phi_-\in\Phi_-}\{\textrm{dropout}(\phi_-(\xbm))\},\\[-1pt]
        \phi_E(\xbm;\Phi_+\cup\Phi_-) &= \textrm{softmax}(\textrm{surr}(\Phi_+,\xbm)(1-\textrm{ref}(\Phi_-,\xbm))).\\[-4pt]
    \end{split}
\end{equation*}
It measures the conferrable score of the sample $\xbm$, where  $\textrm{surr}$ and $\textrm{ref}$ depict the prediction confidence of $\xbm$ on $\phi_+$ and $\phi_-$ respectively. Then, test cases $\X_s$ are generated by the optimization problem as 
\vspace{-0.06in}
\begin{equation*}
        \begin{split}
            \min_{\X_s} \ &\Lcal_{CE}(1,{\max}_t[\phi_{E,t}(\X_s)])\!-\!\alpha_1 \Lcal_{CE}(\phi(\X_0), \phi(\X_s)) \\[-1pt]
            &+\alpha_2 \Lcal_{CE}(\phi(\X_s),\textrm{surr}(\Phi_+,\X_s)),\
            \textrm{s.t.}\ \|\X_s\!-\!\X_0\| \leq \epsilon,  \\[-4pt]
        \end{split}
\end{equation*}
where the first item maximizes the conferrable score of the target class $t$; 
the second item maximizes the prediction difference between the seed samples $\X_0$ and their perturbed version $\X_s$. The third item minimizes the prediction difference between the protected model $\phi$ and the positive models $\phi_+$. The samples with high conferrable scores are selected from $\X_s$ as test cases.

\textit{Generating high-quality test cases in a data-free way}.
Test cases can be generated from randomly-initialized seed samples rather than real samples~\cite{fingerprinting_wang2021characteristic,fingerprinting_pan2021tafa,fingerprinting_chen2022teacherFP}. For instance,  Characteristic Examples~\cite{fingerprinting_wang2021characteristic} is designed to generate test cases with high robustness and low transferability in a data-free way.
For high robustness and low transferability, \cite{fingerprinting_wang2021characteristic} injects random noises into the model parameters and uses a high-pass filter to get the high-frequency component of samples at each iteration. 

\nop{However, existing fingerprint methods fingerprint the decision boundary by adversarial examples [Szegedy et al., 2013], which suffer from the following problems. First, the decision boundary is likely to move under model modification, which makes the fingerprint methods less reliable. Second, adversarial examples can be defeated by adversarial defenses such as input modification [Xu et al., 2017] and adversarial training [Zhang et al., 2019], hence it may fail to identify the real stolen models.To solve those problems, in this paper, we present a robust neural fingerprint method MetaFinger which fingerprints the inner decision area where images in this area can only be recognized by models derived from the source model}

\textit{Test scores as objective}.
Modeldiff~\cite{fingerprinting_li2021modeldiff} generates pairwise test cases, \ie the seed samples $\X_0$ and the perturbed version $\X_s$ by maximizing the test scores $s$ on the protected model $\phi$ as $\max_{\X_s} s(\X_0, \X_s)$. The test score $s$ is defined as:
    \vspace{-0.06in}
\begin{equation*}
    \begin{split}
        {\rm divergence}(\X_0,\X_s) =& \mathop{\rm mean}_{i=1,...,|\X_0|}\{\|\phi(\xbm_i)-\phi(\xbm_i')\|_2\}\\[-3pt]
        {\rm diversity}(\X_s) =& \mathop{\rm mean}_{\xbm_i',\xbm_j'\in\X_s}\{\|\phi(\xbm_i')-\phi(\xbm_j')\|_2\}\\[-3pt]
        s(\X_0, \X_s) = {\rm diverge}&{\rm nce}(\X_0,\X_s) + \lambda\cdot {\rm diversity}(\X_s)\\[-5pt]
    \end{split}
\end{equation*}
\nop{where $\X_\Delta$ denotes the adversarial samples of the seed samples$\X_s$.}
where the first line depicts the intra-input distance, \ie the distance of soft predictions between $\X_0$ and $\X_s$; the second line depicts the inter-input diversity, \ie the diversity of the soft predictions of $\X_s$. 
Since gradients cannot be obtained in black-box construction scenarios, Modeldiff designs an iterative algorithm for black-box generation inspired by mutation testing~\cite{jia2010mutualtesting} and black-box adversarial attacks~\cite{adversarial_demir2019deepsmartfuzzer}: 
\ie constructing some neighbors of low-score samples by random perturbation and then stepping into the one with a higher test score. 

\textit{Triplet loss as objective}.
The decision boundary depicted by adversarial samples shows lacking robustness for model modification and adversarial defenses. MetaFinger~\cite{fingerprinting_yang2022metafinger} extracts test cases from the inner-decision area instead of the boundary. Such area is shared by positive models, extracted by minimizing a triplet loss as
\vspace{-0.06in}
\begin{equation*}
        \Lcal(\xbm) \!=\! \!\mathop{\textrm{mean}}_{\forall \phi_p\in\Phi_+}\!\!\{\textrm{KL}(\phi_{a}(\xbm),\phi_p(\xbm))\}
        -\lambda\cdot\!\mathop{\textrm{mean}}_{\forall \phi_n\in\Phi_-}\!\!\{\textrm{KL}(\phi_{a}(\xbm),\phi_n(\xbm))\}
\vspace{-0.06in}
\end{equation*}
where the anchor $\phi_{a}$ and the positive points $\phi_{p}$ are selected from the positive models $\Phi_+$, and the negative points $\phi_{n}$ are selected from the negative models $\Phi_-$. 
The objective is to find the test cases minimizing the KL divergence between $\phi_{a}(\xbm)$ and $\phi_p(\xbm)$ while maximizing the KL divergence between $\phi_{a}(\xbm)$ and $\phi_n(\xbm)$.


\textit{Finding a universal pattern for test cases}. 
Local adversarial samples, \ie giving a specific perturbation for each sample, can only capture local geometric information of decision boundaries and have high randomness. To this end, UAP-FP~\cite{fingerprinting_peng2022uapfp} constructs a Universal Adversarial Perturbation (UAP) $\Delta$ of the protected model that is normal to most of the vectors on the decision boundary. 
Moreover, to profile the decision boundary better, UAP-FP uses K-means to cluster training samples according to their representation vectors; then picks a sample from each cluster into the seed samples $\X_0$. The UAP $\Delta$ is constructed on $\X_0$. UAP-FP also proposes multi-view fingerprint augmentation that combines $K$ nearest neighbors for each $\xbm\in\X_0$ into the extended test cases, to train a meta-encoder to represent fingerprints in a feature space. 
 \nop{The soft predictions are computed as $\{\ybm_s^K,\ybm_\Delta^K\} = \phi(\{\X_s^K,\X_\Delta^K\})$ where $\X_\Delta^K=\X_s^K+\Delta$.}
 \nop{and train a fingerprint encoder $\Ecal$ using a supervised contrastive loss to represent the fingerprint similarity in a latent space.}

 \nop{Most of the existing test metrics are designed for specific tasks like classification, highly relying on concepts the adversarial examples and classification boundary which have no direct counterparts in other typical learning tasks such as regression and generative modeling. }

 \textit{Task-agnostic fingerprinting}. Most methods for test case generation rely on adversarial samples and decision boundaries not explicitly existing in other tasks like regression and generative tasks. For task-agnostic models, TAFA~\cite{fingerprinting_pan2021tafa} generates test cases $\X_s$ according to the first ReLU layer of the protected model, as 
\vspace{-0.06in}
\begin{equation}\label{eq:tafa1}
    \max_{\X_s}\|\X_s-\X_0\|_1,\  {\rm s.t.} \Abm(\X_s)=\Abm(\X_0),\  \X_s\in\Xcal,
    \vspace{-0.06in}
\end{equation}
where $\Abm(\cdot)$ denotes the activation map of the first ReLU layer.
If the first layer of the protected model is a fully-connected layer or a convolutional layer, the constraint of Equ.~(\ref{eq:tafa1}) is linear inequalities. Thus, TAFA reformulates the problem as a linear programming problem to solve efficiently.
From a similar motivation, 
MetaV~\cite{fingerprinting_pan2022metav} jointly optimizes the test cases $\X_s$ and the meta-classifier $\Vcal$ as the following optimization problem:
\vspace{-0.06in}
\begin{equation*}\label{eq:metaV}
        \mathop{\max}_{\X_s,\Vcal} \mathop{\textrm{mean}}_{\forall \phi_{+}\in\Phi_{+}\cup\phi}\{\log \Vcal(\phi_{+}(\X_s))\}-\alpha\mathop{\textrm{mean}}_{\forall \phi_{-}\in\Phi_{-}}\{\log \Vcal(\phi_{-}(\X_s))\},\\[-6pt]
\end{equation*}
where $\Vcal(\phi(\X_s))$ is a binary classifier to output the probability that $phi$ is a positive model.

\textit{Fingerprinting dynamic DNNs}. Since dynamic DNNs exhibit varied time costs across input samples, Dong~\myetal~\cite{fingerprinting_dong2021dynamic} propose a novel view to fingerprint multi-exit models, \ie inference time. \cite{fingerprinting_dong2021dynamic} generates test cases by maximizing the inference time. For a model with $m$ exits $\{f^l|l\!=\!1...m\}$, the min-objective is defined as
\vspace{-0.06in}
\begin{equation*}
\Lcal(\X_s) = {\sum}_{l=1}^{m-1}\textrm{KL}(f^l(\X_s),\ubm)-\textrm{KL}(f^m(\X_s),\ubm),
\vspace{-0.06in}
\end{equation*}
where $\ubm$ is the uniform vector of the same length as $f^l(\X_s)$. 
The loss pushes the internal exits' soft predictions towards the uniform vector $\ubm$ while pushing the last exit's prediction away from $\ubm$. Thus, when inputting the cases $\X_s$, the protected model will output from the last exit but irrelevant models will not.

\textit{Test cases against transfer learning}. TeacherFP~\cite{fingerprinting_chen2022teacherFP} proposes to fingerprint teacher models to judge from which teacher model a student model is learned. Given the probing samples $\X_s$ and the feature extractor $\phi_{T}$, the cases $\widetilde{\X}_s$ are generated by approximating the probing samples' features $\phi_{T}(\X_s)$, as 
\vspace{-0.09in}
\begin{equation*}
{\min}_{\widetilde{\X}_s}\|\phi_{T}(\widetilde{\X}_s)-\phi_{T}(\X_s)\|_2,\ \textrm{s.t.\ }\widetilde{\X}_s\in\Xcal,
\vspace{-0.06in}
\end{equation*}
where $\Xcal$ denotes the feasible solution space. This constrained optimization can be solved by C\&W with a new variable $\W$, as
\vspace{-0.06in}
\begin{equation*} 
{\min}_{\W}\|\phi_{T}(\tanh(\W))-\phi_{T}(\X_s)\|_2,\ \textrm{s.t.}\ \widetilde{\X}_s = \tanh(\W).
\vspace{-0.06in}
\end{equation*}

\par
\textit{Test cases for different verification scenarios}. 
DeepJudge~\cite{fingerprinting_chen2021copy} proposes to generate tests case for the white-box and black-box verification separately. Adversarial attack methods, like FGSM, PGD, and C\&W, can be directly used for black-box verification. For white-box verification, DeepJudge generates a test case for each neuron covered by the designed test metrics.
Given a seed sample and a selected neuron, DeepJudge adjusts the sample to maximize the neuron's activation by PGD until the activation value surpasses a threshold that is pre-computed using training samples. 
\par
\textbf{Selecting Test Cases}.
Test cases can be selected from a given dataset with simple or even no extra processing. For example,  
SAC~\cite{fingperprinting_guan2022samplecorrelation} proposes two methods: i) SAC-w selects the samples wrongly classified by the protected or pirated models but correctly classified by irrelevant models, and ii) SAC-c applies CutMix, a data augmentation method, on randomly selected samples. 
MeFA~\cite{fingerprinting_liu2022mefa} uses a meta-classifier $\Vcal$ to select some in-distribution data samples as test cases. Here, the meta-classifier $\Vcal$ is trained on the soft predictions of prepared models. If the result of $\Vcal(\phi_{+}(\xbm))$ is "1" (in-distribution), the sample $\xbm$ will be selected. 

\textbf{Fragile Test Cases}. Similar to model watermarking, model fingerprinting can also be used to construct fragile test cases for integrity or functional verification. Here are some examples:
\par
\textit{(a) Sensitive-sample fingerprinting for Integrity Verification}~\cite{fingerprinting_he2019sensitive}.
Here, the test cases $\X_s$ are generated by maximizing the difference of their model predictions on the protected model $\phi$ and its modified version $\phi_+$ as $\max_{\X_s}\|\phi(\X_s)-\phi_+(\X_s)\|$, which is equivalent to minimizing model gradients as
    \vspace{-0.06in}
\begin{equation*}
        {\min}_{\X_s} \left\|\nabla_\thetabm{\phi(\X_s;\thetabm)}\right\|_F^2,\ {\rm s.t.\ } \|\X_s-\X_0\|<\epsilon,
        \vspace{-0.06in}
\end{equation*}
where $\|\!\cdot\!\|_F^2$ denotes the Frobenius norm of a matrix, and the constraint $\|\X_s\!-\!\X_0\|\!<\!\epsilon$ is to make $\X_s$ look like the natural seed samples $\X_0$ to defend against IP evasion attacks. The Maximum Active Neuron Cover (MANC) algorithm is designed to select some test cases with the maximal number of activated neurons from $\X_s$.

\nop{Maximum Active Neuron Cover (MANC) sample selection algorithm to select a small number of samples from $\X_s$, to avoid inactive neurons.
Our criterion is to minimize the number of neurons not
being activated by any Sensitive-Sample, or equivalently,
maximize the number of neurons being activated
at least once by the selected samples. We call the resultant
set of Sensitive-Samples with their corresponding
model outputs, the fingerprint of the DNN model.}

\textit{(b) Public integrity verification in double black-box scenarios}.
To this end, PublicCheck~\cite{fingerprinting_wang2022publicheck} designs encysted samples generated by attribute manipulation along a semantic feature axis.
First, a disentangled auto-encoder model $\phi$ is trained on partial training samples of the protected model for attribute-level and abstraction-level disentanglement.
Then, it uses the model $\phi$ to manipulate the seed samples $\X_0$ in the latent space until causing prediction changes, records the latent vectors $\Z_s$, and randomly samples the augmented latent vectors $\Z'_s$ around $\Z_s$. The augmented candidates $\X_s$ are reconstructed using the decoder $\phi_{DE}$ as $\X_s \!= \!\phi_{DE}(\Z'_s)$. 
Against IP evasion attacks, the encysted samples are selected from $\X_s$ via a smoothness metric to ensure pixel-smooth patterns $\X_s\!-\!\X_0$. 
\par
\textit{(c) Functionality Verification}~\cite{fingerprinting_wang2021intrinsic}. 
It is a variety of integrity verification to detect adversarial third-party attacks like transfer learning and backdoor attacks while enabling model compression. Here, the test cases $\X_s$ are randomly initialized and updated by a min-max optimization as: 
    \vspace{-0.06in}
\begin{equation*}
    \min_{\X_s}\max_{\thetabm_\Delta} \Lcal(\phi_\Delta(\X_s;\thetabm\!+\!\thetabm_\Delta),\Ybf_t),\ {\rm s.t.} 
    \ \|\thetabm_\Delta\| <\epsilon,
        \vspace{-0.06in}
\end{equation*}
where the inner max-optimization constructs the perturbed version $\phi_\Delta$ of the original model, and the outer min-optimization obtains the adversarial samples $\X_s$ of $\phi_\Delta$ to the target label $\Ybf_t$. The two processes are performed alternatively.







\subsubsection{Test Metric Design}\label{sec:Protect_Non_Invasive_Behaviors_Metric}
It is to extract the quantifiable measurements of model behaviors on test cases, which can be extracted from three main levels: \ding{172} model outputs like the matching rate of hard predictions~\cite{fingerprinting_cao2021ipguard, fingerprinting_lukas2019deep} or the similarity/projection of soft predictions~\cite{fingerprinting_peng2022uapfp,fingerprinting_li2021modeldiff,fingerprinting_liu2022mefa,fingperprinting_guan2022samplecorrelation,fingerprinting_maini2021datasetinference,fingerprinting_dziedzic2022DI4selfsupervise}; \ding{173} inner model components like neuron-wise or layer-wise activation~\cite{fingerprinting_pan2021tafa, fingerprinting_chen2021copy}; \ding{174} model properties like adversarial robustness~\cite{fingerprinting_chen2021copy} or model interpretation results (\emph{e.g.}, LIME weights~\cite{fingerprinting_jia2022Zest}); 

\textbf{Output-level Test Metrics}. Most of the current works ~\cite{fingerprinting_cao2021ipguard,fingerprinting_lukas2019deep,fingerprinting_pan2022metav,fingerprinting_peng2022uapfp,fingerprinting_liu2022mefa,fingerprinting_yang2022metafinger,fingerprinting_wang2021characteristic,fingerprinting_zhao2020afa,fingerprinting_wang2022publicheck, fingerprinting_wang2021deepfool} use model outputs like soft/hard predictions $\phi(\X_s)$ (or $\phi(\{\X_0,\X_s\})$) on test cases $\X_s$ (or pairwise test cases $\{\X_0,\X_s\}$) to construct test metrics. 
Soft predictions have more information about model behavior and can support more ways for fingerprint comparison, but sometimes may be unavailable in the black-box scenarios. If necessary, label smoothing can be applied to transform hard predictions to a soft probability vector. 
\par
\textit{Matching Rate of Hard Predictions}~\cite{fingerprinting_cao2021ipguard,fingerprinting_lukas2019deep,fingerprinting_yang2022metafinger,fingerprinting_wang2021characteristic,fingerprinting_zhao2020afa,fingerprinting_wang2022publicheck, fingerprinting_wang2021deepfool}. One of the most commonly used test metrics is the fraction of test cases that yield identical hard predictions between the suspect model $\hat\phi$ and the protected model $\phi$, which is formulated as 
\vspace{-0.06in}
\begin{equation}\label{eq:mr}
{\rm MR}(\phi,\hat\phi)\!=\!|\X_{m}|/|\X_s|,\X_{m}\!=\!\{\xbm|\phi(\xbm)\!=\!\hat{\phi}(\xbm),\forall\xbm\!\in\!\X_s\},
\vspace{-0.06in}
\end{equation}
where $\X_s$ denotes the test cases and $\X_m$ denotes the matching set. 
\nop{An MR measures the proportion of the matching set, \ie the test cases that have the same predicted labels between the protected model $g$ and the suspected model $\hat{g}$.}
\par
Generally, if the test cases are good enough, such test metrics have enough QoI to judge piracy/irrelevant models. It requires test cases not too close to the decision boundary and the class centroid. The former leads to a high false negative rate and the vulnerability against IP removal, while the latter may misjudge irrelevant models as piracy models. However, it is hard to satisfy. Moreover, for some tasks like self-supervised learning or transfer learning, fingerprints should be constructed in a representation domain rather than model predictions. 
Thus, some works develop more possibilities of test metric design for a better QoI, generally by processing the model predictions or neuron activation on test cases. 
\par
\textit{Metrics on Soft Predictions}. 
\par
To our knowledge, Modeldiff~\cite{fingerprinting_li2021modeldiff} propose the first test metrics on soft predictions, named Decision Distance Vectors (DDV). A DDV captures the decision pattern of a model on pairwise test cases $\{\X_0,\X_s\}$. $\X_0$ and $\X_s$ denote the seed samples and their perturbed version, respectively. The DDV of the model $\phi$ is defined as
\vspace{-0.04in}
\begin{equation}\label{eq:ddv}
    DDV_g=\{dist(\phi(\xbm),\phi(\xbm'))|\forall \xbm\in\X_s\},
    \vspace{-0.04in}
\end{equation}
where $\xbm'\in\X_\Delta$ is the perturbed version of $\xbm$, and $dist$ is the cosine distance if $\phi(\xbm)$ is a 1-D array. 

\nop{The distance metric dist is the Cosine distance here if f(x) is a 1-D array (e.g. when f is a classifier) since Cosine distance is good at comparing different scales of vectors. A DDV basically captures the decision pattern of a model on the test inputs. If two models are similar, they would have similar patterns when measuring the distance between each pair of test inputs. The concept is analogous to testing two people with the same quiz questions, they would give similar answers if they have common knowledge.}

The existing fingerprinting schemes based on transferable adversarial samples are sensitive to transfer learning scenarios and adversarial defense strategies. Thus, SAC~\cite{fingperprinting_guan2022samplecorrelation} designs the
pairwise sample correlation as the test metric. 
Given the test cases $\X_s$ and their model predictions $\mathcal{O}=\{\obm_i|\obm_i=\phi(\xbm_i), \forall \xbm_i\in\X_s\}$, SAC computes a model-specific correlation matrix $\C\in\Rbb^{|\X_s|\cdot|\X_s|}$ as 
\vspace{-0.06in}
\begin{equation}\label{eq:sac}
       \C=\{c_{i,j}|c_{i,j} = corr(\obm_i, \obm_j), i,j=1...|\X_s|\},  
    \vspace{-0.06in}
\end{equation}
where $corr$ could be cosine similarity or Gaussian RBF, as 
\vspace{-0.06in}
\begin{equation*}
    \begin{split}
       corr(\obm_i, \obm_j) &= cos(\obm_i, \obm_j) = 
       {\obm_i^T \obm_j}/({\|\obm_i\|\|\obm_j\|}) \ \ \textrm{or}\\[-1pt]
       corr(\obm_i, \obm_j) &= RBF(\obm_i, \obm_j) = \exp(-
       {0.5\|\obm_i-\obm_j\|^2_2}/{\delta^2}).\\[-3pt]
    \end{split}
    \vspace{-0.06in}
\end{equation*}
Note that $\obm_i$ is a probability vector. For those scenarios that can only obtain the suspect model's hard predictions, label smoothing is applied to transform the hard predictions into a probability vector, as 
       $\obm =(1-\epsilon)\eta(y)+\epsilon \ubm,$
where $\ubm$ is a uniform vector and $\eta(y)$ is a one-hot encoding for the hard prediction $y$.

\par
Besides matching rates (MR), TeacherFP~\cite{fingerprinting_chen2022teacherFP} proposes two extra belief metrics to select which model the suspect model $\hat{\phi}$ learns from, given $m$ candidate teacher models $\Phi = \{\phi_i|i=1...m\}$: Eccentricity (Ecc), and Empirical Entropy (EE), defined as
\vspace{-0.06in}
\begin{equation*}
    \begin{split}
        {\rm Ecc}&= ({{\max}_1(\vbm)\!-\!{\max}_2(\vbm)})/{\textrm{std}(\vbm)},\ \vbm\!=\!\{{\rm MR}({\phi},{\hat\phi})|\forall \phi\in\Phi\},\\
         {\rm EE} &= -{\sum}_{y\in\Ycal} \hat{p}_y\log \hat{p}_y,\ \hat{p}_y=\textrm{Prob}(\phi^*(\xbm)=y),\\[-4pt]
    \end{split}
    \vspace{-0.06in}
\end{equation*}
where $\phi^*={\arg\max}_{\phi}{{\rm MR}}({\phi},{\hat\phi})$ is the model with the maximal MR in  candidate teacher models $\Phi$, ${\max}_k$ is to find the $k$-th largest value in the set, and $\textrm{std}(\vbm)$ refers to the standard deviation of $\vbm$.
Ecc measures the prominence degree of the extrema in a set and EE depicts how much information is given to make the decision. 

\textbf{Component-level Test Metrics}.

\nop{Most of the existing test metrics are designed for specific tasks like classification, highly relying on concepts the adversarial examples and classification boundary which have no direct counterparts in other typical learning tasks such as regression and generative modeling. }
Most existing test metrics are designed for specific tasks like classification, highly relying on task-specific model predictions.
Thus, for task-agnostic models independent from specific tasks and training samples, TAFA~\cite{fingerprinting_pan2021tafa} design a method derived from the activation of the first ReLU layer $\Abm$. It depicts the distance from the random seeds $\X_0$ to the boundary of linear regions, \ie the adversarial version $\X_s$ of $\X_0$. Based on the first-ReLU activation on a suspect model, TAFA computes the normalized Hamming distance as the confidence score, as
\vspace{-0.06in}
\begin{equation}\label{eq:tafa3}
    {s_{wb}} = 1-\frac{1}{d_1} \mathop{\textrm{mean}}_{\forall \xbm_0\in\X_0}
    \{tr(|\Abm(\xbm_0)-\Abm(\xbm_s)|)\},
    \vspace{-0.06in}
\end{equation}
where $d_1$ is the dimension of a sample's ReLU activation and $\xbm_\Delta$ is the adversarial version of $\xbm$ derived from Equ.~(\ref{eq:tafa1}).
For black-box verification, \textit{Pointsonline} is used to approximate this metric. 

DeepJudge~\cite{fingerprinting_chen2021copy} designs more comprehensive and fine-grained test metrics on three levels: \ding{172} Neuron-level like Neuron Output Distance (NOD) and Neuron Activation Distance (NAD); \ding{173} Layer-level like Layer Activation Distance (LAD) Jensen-Shanon Distance (JSD); and \ding{174} \textit{Property-level} like Robustness Distance (RobD).
\nop{Intuitively, the output of each neuron in a model follows its own statistical distribution, and the neuron outputs in different models should vary. Motivated by this, DEEPJUDGE uses the output status of neurons to capture the difference between the two models and defines the following two neuron-level metrics NOD and NAD.}

\textit{Neuron-wise Test Metrics}. 
The output of each neuron in a model follows its own statistical distribution. 
Given the $i$-th neuron's output of the protected model and the suspect model on the $l$-layer $f^l_i,\hat{f}^l_i$, the neuron-wise metrics, NOD and NAD, are defined as 
\vspace{-0.06in}
\begin{equation*}
    \begin{split}
\textrm{NOD}(f^l_i,\hat{f}^l_i,\X_s)&={\sum}_{\xbm\in\X_s}|f^l_i(\xbm)-\hat{f}^l_i(\xbm)|,\\
\textrm{NAD}(f^l_i,\hat{f}^l_i,\X_s)&={\sum}_{\xbm\in\X_s}|S(f^l_i(\xbm))-S(\hat{f}^l_i(\xbm))|,\\[-4pt]
    \end{split}
    \vspace{-0.06in}
\end{equation*}
where the step function $S(\phi_{l,i}(\xbm))$ returns $1$ if $\phi_{l,i}(\xbm)$ is greater than a threshold, $0$ otherwise. 

\textit{Layer-wise Test Metrics}, like LOD, LAD and JSD, depict the model difference in layer outputs, defined as 
\nop{The layer-wise metrics in DEEPJUDGE take into account the output values of the entire layer in a DNN model. Compared with neuron-level metrics, layer-level metrics provide a full-scale view of the intermediate layer output difference between two models.}
\vspace{-0.06in}
\begin{equation*}
    \begin{split}
\textrm{LOD}(f^{l},\hat{f}^{l},\X_s)&={\sum}_{\xbm\in\X_s}\|f^{l}(\xbm)-\hat{f}^{l}(\xbm)\|_2,   \\
\textrm{LAD}(f^{l},\hat{f}^{l},\X_s)&={\sum}_{i=1}^{|N_l|}\textrm{NAD}(f^l_i,\hat{f}^l_i,\X_s),
    \end{split}
    \vspace{-0.1in}
\end{equation*}
\vspace{-0.08in}
\begin{equation*}
\textrm{JSD}(f^{L},\hat{f}^{L},\X_s)={\sum}_{\xbm\in\X_\Delta}\{\textrm{KL}(f^{L}(\xbm),\qbm)+\textrm{KL}(\hat{f}^{L}(\xbm),\qbm)\}
\vspace{-0.06in}
\end{equation*}
 \nop{neuron $\phi$ $\hat{\phi}$ and activation function $S$, 
 $f^l$ and $\hat{f}^l$ denote the $l$-layer output functions  of the protected model $g$ and the suspected model $\hat{g}$.}
 where $f^l$ and $\hat{f}^l$ denote the $l$-th layer's output of the protected and suspect model, $\qbm\!=\!(f^{L}(\xbm)\!+\!\hat{f}^{L}(\xbm))/2$, and $f^{L}$ denotes the output layer. 
 \nop{JSD quantifies the similarity between two models’ output distributions, and is particularly more powerful against model extraction attacks where the suspect model is extracted based on the probability vectors (distributions) returned by the victim model.}
JSD quantifies the model similarity in output distributions; thus it is robust against model extraction attacks.

\textbf{Property-level Test Metrics}.
The property-level metrics are a set of model properties used to characterize model similarity, like
RobD that depicts models' difference in adversarial robustness:
\vspace{-0.06in}
\begin{equation*}
    RobD(\phi,\hat{\phi},\X_s)= {\sum}_{\xbm\in \X_s}|(\mathbbm{1}(\phi(\xbm)\neq \ybm_s)-\mathbbm{1}(\hat{\phi}(\xbm)\neq \ybm_s)|,
    \vspace{-0.06in}
\end{equation*}
where $\X_s$ denotes test cases, 
$\ybm_s$ denotes the true label for each $\xbm\in\X_s$, and $\mathbbm{1}$ denotes the indicator function. RobD and JSD are designed for black-box verification scenarios while the others, NOD, NAD, LOD and LAD, are designed for white-box verification.

\textit{Train-Test Margin}.
Dataset Inference (DI)~\cite{fingerprinting_maini2021datasetinference} leverages the train-test margin on model predictions as a test metric. The intuition is that, the protected model exhibits more confidence in its training samples rather than in unseen testing samples. It is obtained by gradient descent with the following optimization objective: 
\vspace{-0.06in}
\begin{equation}\label{eq:DI}
    {\arg\min}_{\deltabm_t} d(\xbm, \xbm+\deltabm_t),\ \textrm{s.t.}\ \phi(\xbm+\deltabm_t)=\tbm, \forall \tbm \neq \ybm,  
    \vspace{-0.06in}
\end{equation}
where $\ybm$ is the hard label of a verifier-private training sample $\xbm$, and $d$ is the samples' distance metric. Then, the prediction margin, \ie the minimal distance to the decision boundary with $T$-classes $\{{\delta}_\tbm|t=1...T\}$, is obtained.
For the black-box construction scenarios, the gradients cannot be obtained; thus, DI gives a fixed perturbation $\delta_0$ and minimizes $k_t$ as 
\vspace{-0.06in}
\begin{equation*}
    {\arg\min}_{k_t}d(\xbm, \xbm+k_t\deltabm_0),\ \textrm{s.t.}\ \phi(\xbm+k_t\deltabm_0)=\tbm, \forall \tbm \neq \ybm.  
    \vspace{-0.06in}
\end{equation*}
The theoretical proof shows that DI succeeds nearly every time since DI aggregates the signal over multiple samples. 
On this basis, Adam~\myetal propose SSL-DI)~\cite{fingerprinting_dziedzic2022DI4selfsupervise}, an extended DI for Self-Supervised Learning models. 
DI is an attractive way to protect SSL models because it is costly to retrain large SSL models~\cite{ssl_he2022mae} and the performance will affect all downstream tasks. However, the original DI method~\cite{fingerprinting_maini2021datasetinference} relies on computing the distances from samples and decision boundaries: \ding{172} SSL encoders trained on unlabeled data have no explicit decision boundaries; \ding{173} SSL models do not have enough overfitting to show a clear train-test margin on predictions. Thus, SSL-DI uses the data density on the representation domain as the margin. Its key principle is that the log-likelihood of output representation on training samples is larger than on testing samples for the positive models, however, this is not the case for the negative models. The log-likelihood is depicted by density estimation using Gaussian Mixture Models.

\textit{LIME weights}. The performance on robustness and uniqueness of most test metrics depends heavily on excellent test cases that are hard to achieve, while the case-free methods based on weight comparison are hard to support cross-architecture verification and black-box scenarios. To this end,
ZEST~\cite{fingerprinting_jia2022Zest} introduces the Local Interpretable Model-Agnostic Explanations (LIME) algorithm for test metrics. 
First, ZEST selects some training samples, and for each sample $\xbm_i$, a set of its nearby samples $\X_{i}^\Delta$ is constructed using the Quickshift method~\cite{sample_vedaldi2008quickshift}. Then, for each sample set $\X_{i}^\Delta$, ZEST trains a local linear regression model $h_i:\X_{i}^\Delta\rightarrow \phi(\X_{i}^\Delta)$ to approximate the model's local behavior on $\X_{i}^\Delta$. Finally, the weights of all the regression models are concatenated as an IP message approximating the global model behavior.


In the future, more possibilities for test metric designs, especially with interpretability or theoretical constraints, need to be explored to cover various emerging tasks and multi-view demands. 
\subsubsection{Fingerprint Comparison}\label{sec:Protect_Non_Invasive_Behaviors_Compare}
Test metric thresholds are commonly used for fingerprint comparison~\cite{fingerprinting_cao2021ipguard,fingerprinting_lukas2019deep,fingerprinting_wang2021characteristic,fingerprinting_yang2022metafinger, fingerprinting_zhao2020afa, fingerprinting_chen2022teacherFP, fingerprinting_wang2022publicheck, fingerprinting_pan2021tafa, fingerprinting_dong2021dynamic, fingperprinting_guan2022samplecorrelation, fingerprinting_wang2021deepfool, fingerprinting_jia2022Zest}. Most methods based on hard predictions use a threshold on the matching rate as Equ.~(\ref{eq:mr}) to judge pirated models. If MR$(\phi,\hat{\phi})$ is larger than a threshold $\epsilon$, the suspect model $\hat\phi$ is regard as piracy. 
On this basis, TeacherFP~\cite{fingerprinting_chen2022teacherFP} develops a "one-of-the-best" strategy for transfer learning.
It first selects the supporting set by removing the test cases with the maximum occurrence from the matching set. Then, given a suspect model and some candidate teacher models, TeacherFP selects the victim model with the maximum MR on the supporting set. 
\par
Simple threshold comparison is also used for elaborate test metrics.
For example, Modeldiff uses a threshold on the cosine similarity of DDVs (Equ.~(\ref{eq:ddv})); TAFA uses a threshold on the distances from test cases to region boundaries as Equ.~(\ref{eq:tafa3}); Dong~\myetal~\cite{fingerprinting_dong2021dynamic} use a threshold on EEC AUC scores; DeepFool-FP~\cite{fingerprinting_wang2021deepfool} uses a confidence threshold on a probability derived from Bayesian Theorem; SAC~\cite{fingperprinting_guan2022samplecorrelation} uses a distance threshold on
sample correction matrices as Equ.~(\ref{eq:sac});
ZEST~\cite{fingerprinting_jia2022Zest} uses a distance threshold on LIME weights. 
DeepJudge~\cite{fingerprinting_chen2021copy} determines a specific threshold for each test metric and designs a majority voting strategy: the model is judged pirated if more than half of the metrics of the suspected model exceed the given thresholds. 
It is worth mentioning that, though a simple voting method can work in DeepJudge, more exploration for multi-metric decisions is valuable in practical scenarios. 
\par
Some advanced methods apply machine learning models for fingerprint comparison. SSL-DI~\cite{fingerprinting_dziedzic2022DI4selfsupervise} trains Gaussian Mixture Models (GMM) on data representations to estimate the data density in the latent space. MeFA~\cite{fingerprinting_pan2022metav}, MetaV~\cite{fingerprinting_pan2022metav}, and DI~\cite{fingerprinting_maini2021datasetinference} train binary fingerprint classifiers $\Vcal$ whose outputs are close to "1" given piracy fingerprints, and "0" given irrelevant fingerprints. Therein, MeFA and MetaV use soft predictions on test cases as inputs, while DI uses the prediction margin as Equ.~(\ref{eq:DI}). Similarly,
UAP-FP~\cite{fingerprinting_peng2022uapfp} uses augmented fingerprints to train an encoder to represent fingerprints into a latent space and compares the fingerprints by the cosine similarity of their representations. To leverage the label information, UAP-FP trains the two-branch structure of SimCLR~\cite{simclr_chen2020simple} via a supervised contrastive loss as
\vspace{-0.06in}
\begin{equation*}
        \Lcal = -\sum_{i\in I}
        \mathop{\textrm{mean}}_{\forall \mu\in\Psi(i)}
        \{\log \frac{\exp(\cos(\zbm_i,\zbm_\mu)/\tau)}{\sum_{\nu\in \{I \setminus i\}}\exp(\cos(\zbm_i,\zbm_\nu)/\tau)}\}
    \vspace{-0.06in}
\end{equation*}
where $\zbm = E(\{\phi(\X_0),\phi(\X_s)\})$ is the output of the encoder $E$; 
$I$ denotes the batch index; 
$\Psi(i)=\{\mu|\mu\in\{I \setminus i\},\tilde{y}_\mu=\tilde{y}_i\}$ denotes the fingerprint indexes with the same label of the $i$-th fingerprint. 

\par
\textit{Preset or Data-driven Threshold}. 
The performance of model fingerprinting is sensitive to the choice of thresholds, even for meta-classifiers. 
The threshold can be preset as a empirical value~\cite{fingerprinting_cao2021ipguard,fingerprinting_lukas2019deep,fingerprinting_wang2021characteristic,fingerprinting_yang2022metafinger}, or be determined in a data-driven way. For example, AFA~\cite{fingerprinting_zhao2020afa} 
uses half of the average MR of some positive models as the threshold; SAC~\cite{fingperprinting_guan2022samplecorrelation} uses the average of the correlation scores of negative/positive models; and 
MeFA~\cite{fingerprinting_pan2022metav} use the average of the outputs from fingerprint classifier on test cases.
Note that the MR threshold is set a 1 for the integrity verification task ~\cite{fingerprinting_wang2022publicheck,fingerprinting_he2019sensitive}.
\par
\textit{Hypothesis Testing} is often used to avoid manual threshold setting. 
DI~\cite{fingerprinting_maini2021datasetinference} forms a two-sample $t$-Test on the distribution on confidence score vectors. First, it extracts the prediction margins ${\bm \delta}$ and $ {\bm \delta}'$ of some training samples on the protected and suspected model as Equ.~(\ref{eq:DI}), respectively; then inputs ${\bm \delta}, {\bm \delta}'$ to the confidence regressor (a fingerprint classifier) to get confidence score vectors $\cbm$ and $\cbm'$. Next, a two sample T-test is formed on $\cbm$ and $\cbm'$. If the $p$-value for the one-sided hypothesis $H_0:\bar{\cbm}'<\bar{\cbm}$ is smaller than 0.05, the suspected model is judged pirated. Note that  $\bar{\cbm}$ is the mean of $\cbm$. 
UAP-FP uses a similar two-sample $t$-Test~\cite{fingerprinting_peng2022uapfp} on $\Omega'$ and $\Omega_-$, \ie the fingerprint similarities from suspected models and irrelevant models. The suspected model is judged pirated if $\Omega'$ is significantly greater than $\Omega_-$. 
SSL-DI~\cite{fingerprinting_dziedzic2022DI4selfsupervise} forms a hypothesis test on the density from GMMs: if the representation density of training samples is significantly greater than that of unseen samples, the test model is stolen.
Based on hypothesis tests, DeepJudge~\cite{fingerprinting_chen2021copy} designs an $\varepsilon$-difference strategy to select a threshold for each metric $\lambda$. It uses one-tailed $t$-test to get the lower bound $LB_\lambda$ based on the statistics of negative models at the 99\% confidence level and give the threshold as $\alpha_\lambda LB_\lambda$, where $\alpha_\lambda$ is a relaxing coefficient.


\nop{
B: Fingerprint (Meta) Classifier/ Encoder/Estimator. ~\cite{fingerprinting_liu2022mefa,fingerprinting_pan2022metav,fingerprinting_peng2022uapfp, fingerprinting_dziedzic2022DI4selfsupervise, fingerprinting_maini2021datasetinference}
MeFA~\cite{fingerprinting_pan2022metav} exception
MetaV~\cite{fingerprinting_pan2022metav} concat
UAP-FP~\cite{fingerprinting_peng2022uapfp} concat encoder.
DI: Confidence Regressor~\cite{fingerprinting_maini2021datasetinference}
SSL-DI~\cite{fingerprinting_dziedzic2022DI4selfsupervise} GMM

C: Hypothesis Testing: Confidence Regressor and Hypothesis Testing~\cite{fingerprinting_maini2021datasetinference}; GMM Density Estimator
and Hypothesis Testing~\cite{fingerprinting_dziedzic2022DI4selfsupervise};
UAP-FP~\cite{fingerprinting_peng2022uapfp}, DeepJudge~\cite{fingerprinting_chen2021copy}
}

\subsubsection{Target Scenarios, Models, \& Functions} \label{sec:Protect_Non_Invasive_Behaviors_Scene}
We describe the following aspects of fingerprinting methods: 
\par
\textit{(a) Construction scenarios}: 
Most works require white-box access to construct fingerprints because they need model gradients to obtain test cases/metrics. Five works can be used for 
 black-box construction: 
Modeldiff~\cite{fingerprinting_li2021modeldiff}, DI~\cite{fingerprinting_maini2021datasetinference}, and PublicCheck~\cite{fingerprinting_wang2022publicheck} are inspired by black-box adversarial attacks. They set a fixed perturbation direction and gradually increase the perturbation step length until changing predicted labels.
DI/SSL-DI~\cite{fingerprinting_maini2021datasetinference,fingerprinting_dziedzic2022DI4selfsupervise} and ZEST~\cite{fingerprinting_jia2022Zest} directly use private training samples as test cases and require only model outputs to compute test metrics.
These methods have been described in detail above and thus we do not repeat them here.
In addition, No-box construction is valid for generative models like FPNet~\cite{new_jeong2022fingerprintnet} without specific test cases. 
\par
\textit{(b) Verification scenarios}: Weight-based methods can be used for only white-box verification while almost all the behavior-based methods work on black-box verification, except the work~\cite{fingerprinting_dong2021dynamic} using white-box inference time. 
Note that TAFA~\cite{fingerprinting_pan2021tafa}, DeepJudge~\cite{fingerprinting_chen2021copy}, and DI~\cite{fingerprinting_maini2021datasetinference} can extract better test metrics with white-box gradients or hidden-layer outputs.
\par
\textit{(c) Target functions}: Most works focus on copyright verification while some works extend DNN fingerprinting to other tasks: 
\cite{fingerprinting_he2019sensitive} and \cite{fingerprinting_wang2022publicheck} 
focus on integrity verification on classification models; \cite{fingerprinting_wang2021intrinsic} proposes robust fingerprints for functionality verification; 
\cite{fingerprinting_xiong2022neural} proposes tampering localization to identify the tampered position of the original model; 
TeacherFP~\cite{fingerprinting_chen2022teacherFP} proposes teacher model fingerprinting attacks against transfer learning to judge which model is a given student model learned from; FPNet~\cite{new_jeong2022fingerprintnet} proposes generated image detection to judge whether an image is generated or real. 
Most works focus on the robustness against IP Removal attacks and the distinguishing ability against pirated and irrelevant models. However, few works focus on potential attacks during IP verification, like IP evasion and ambiguity. Some methods can defend against IP detection attacks in part, like DI/SSL-DI~\cite{fingerprinting_maini2021datasetinference,fingerprinting_dziedzic2022DI4selfsupervise} using normal training samples, or 
\cite{fingerprinting_wang2022publicheck}, \cite{fingerprinting_he2019sensitive}, and \cite{fingerprinting_li2021ganfp}
extracting test cases close to normal samples. Moreover, interpretable fingerprinting schemes with theoretical guarantees, or fine-grained cause localization of model similarity, have not been focused on.
\par
\textit{(d) Target models}: Most works are designed for classification models.  Besides, other learning tasks 
are explored: 
TAFA~\cite{fingerprinting_pan2021tafa} and MetaV~\cite{fingerprinting_pan2022metav} fingerprint task-agnostic models by the first ReLU activation and a fingerprint classifier, respectively;
SSL-DI~\cite{fingerprinting_dziedzic2022DI4selfsupervise} utilizes the train-test margin on representation density to fingerprint encoder models trained from self-supervised learning; Dong \myetal~\cite{fingerprinting_dong2021dynamic}  fingerprints multi-exit models via inference time. GAN-FP~\cite{fingerprinting_li2021ganfp} and FPNet~\cite{new_jeong2022fingerprintnet} fingerprint generated models. 
GAN-FP~\cite{fingerprinting_li2021ganfp} cascades the protected Generative Adversarial Network (GAN) model with a classifier head $\Vcal$ that predicts the class labels of generated outputs. 
It constructs test latent codes with covert generated outputs against IP evasion in three ways: \ding{172} creating adversarial latent codes whose generated outputs fool the classifier $\Vcal$; \ding{173} using a CE loss to finetune $\Vcal$ to embed invisible backdoors (normal generated outputs with modified labels); \ding{174} using a triplet loss and a fine-grained categorization method to embed backdoors. The latter two are similar to DNN watermarking, but they only finetune $\Vcal$ rather than the generator. 
FPNet~\cite{new_jeong2022fingerprintnet} designs a 
generated image detection method for unseen generative models. It treats a generative model as a combination of filters and uses the specific spatial spectrum from the upsampling process of image generation as model fingerprints. A fingerprint generator is designed to simulate the fingerprints of unseen generative models, and a fingerprint detection network is trained to determine whether a given image is generated or real.
Vonderfecht~\myetal~\cite{fingeprinting_vonderfecht2022fingerprints} extend fingerprints for generative models to single-image super-resolution (SISR) models and confirm that detectable fingerprints do exist in SISR models, especially using high upscaling factors or adversarial loss.
\par


\section{Testing Protocols \& Performance}\label{sec:performance}
\subsection{Existing Testing Protocols}
At present, there is no testing protocol for Deep IP defence and attack, which can cover watermarking and fingerprint schemes, span the whole pipeline of IP registration and verification, and test multiple metrics,  like model fidelity, QoI, and EoI, in a fair and comprehensive way.
Some works try to provide a fair performance comparison for contemporary model watermarking methods. For example, Chen \textit{et al}~\cite{survey_chen2018performance} propose an experimental performance comparison on fidelity, robustness and integrity. Six classic schemes are considered, including inner-component-based~\cite{white_uchida2017embedding,white_rouhani2018deepsigns} and trigger-based watermarking~\cite{black_adi2018turning,black_le2020adversarial,white_rouhani2018deepsigns,black_zhang2018protecting}. The robustness is tested against model fine-tuning attacks, parameter pruning attacks and watermark overwriting attack. \cite{black_lee2022evaluating}\footnote{\url{https://github.com/WSP-LAB/wm-eval-zoo}} evaluates the robustness and fidelity of some trigger-based methods. It considers three types of attacks: detection, removal and ambiguity. For IP removal attacks, it proposes adaptive attacks: it consumes that the adversary knows the algorithm but not specific trigger samples, and trains a generative model to mimic trigger samples to repair the predictions. However, these testing protocols are based on certain fixed configurations of watermarking schemes, like the learning rate and the similarity threshold, which is hard to provide a fair comparison. 
Lukas \textit{et al}~\cite{sok_lukas2021sok}\footnote{\url{https://crysp.uwaterloo.ca/research/mlsec/wrt}, \url{https://github.com/dnn-security/Watermark-Robustness-Toolbox}} propose WRT, a watermarking-robust-toolkit, to test DNN watermarking on fidelity, robustness and efficiency. It formulates the embedding and removal of watermarks as a security game. It first constructs a payoff matrix for multiple attack and defence configurations, where an element represents the attack payoff that is zero for unsuccessful attacks and is  equal to the accuracy of the surrogate model otherwise. The attacker and defender choose their configurations based on the payoff matrix to achieve Nash Equilibrium. The attacker will win if the normalized watermark accuracy is lower than 0.5 within 5\% loss on the task accuracy. \cite{sok_lukas2021sok} mainly tests the robustness against IP removal attacks and implements various attacking means including three types: input processing, model modification and model extraction. More advanced attack and defence strategies, more performance metrics and more complete test pipelines need to be studied continuously with the development of Deep IP, towards commercial testing protocols.
\subsection{Performance of representative protection methods}
This survey has summarized some performance metrics of Deep IP Protection in Section~\ref{sec:background_criteria}. Here, we compare the performance of the current open-source DeepIP solutions on these metrics. The results shown in Table ~\ref{tab:performance} are benchmarked against the experimental results in existing papers that focus on performance comparisons~\cite{sok_lukas2021sok,black_lee2022evaluating}. Moreover, we refer to the theoretical analysis and experimental results in the papers of the corresponding schemes. 
We define several performance levels which are 'A', 'B',and 'C' from high to low. Specially, '?' means that the metric of the method is without any experimental test and theoretical analysis. 'N/A' stands for unclear, meaning that existing results do not provide a performance level clear enough. The superscript 'T' indicates that the metric is theoretically guaranteed or partially guaranteed.

\begin{table}
\centering
  \caption{The Performance of Deep IP Protection }
  \label{tab:performance}
    \setlength\tabcolsep{1pt}
    \scriptsize
    \begin{tabular}{|c|c|ccc|c|cccccc|cc|c|}
    \toprule
    
    \multicolumn{2}{|c|}{\multirow{2}{*}{\bf Methods}}
                           &   \multicolumn{3}{c|}{\textbf{Fidelity}}  
                           &   \textbf{Capacity} 
                           &   \multicolumn{6}{c|}{\textbf{Robustness}}  
                           &   \multicolumn{2}{c|}{\textbf{Others}}
                           &   \multicolumn{1}{c|}{\multirow{2}{*}{\makecell{\bf Code\\\bf Links}}}\\
     \multicolumn{2}{|c|}{} & \rotatebox{0}{\it Acc} & \rotatebox{0}{\it Eff} & \rotatebox{0}{\it Risk} 
                            & \rotatebox{0}{\it Upp/Pay} 
                           & \rotatebox{0}{\it Pre} & \rotatebox{0}{\it Mod} & \rotatebox{0}{\it Ext}
                           & \rotatebox{0}{\it Amb} & \rotatebox{0}{\it Eva} & \rotatebox{0}{\it Rmv}
                           & \rotatebox{0}{\it FPR} & \rotatebox{0}{\it Sca} & \\
    \midrule
    \multirow{32}{*}{\rotatebox{90}{\bf Watermarking}}
    &Uchida~\cite{white_uchida2017embedding}   & B & A & - & ?/Multi  & - & B & C & C & C & B- & A  & B & \cite{url_uchida}\\
    &Deepmarks~\cite{white_chen2018deepmarks}          & B & A &  - & ?/Multi & - & B & C  & C & C & B- & A  & B & \cite{url_sok} \\ 
    &Deepsigns~\cite{white_rouhani2018deepsigns}       & B & A &  - & ?/Multi & - & B+ & C  & B & B & B & A  & B & \cite{url_sok} \\
    &Residuals~\cite{white_liu2021residuals}  & A & A &  - & ?/Multi  & - & B+ & C  & B+ & B & B & A  & B & \cite{url_liu} \\ 
    &RIGA~\cite{white_wang2021riga}                    & B & A & - & ?/Multi  & - & B+ & C    & B- & B+ & B & A  & B    &\cite{url_riga}\\ 
    &MOVE~\cite{white_li2022defending}                 & A & A & - & ?/One   & - & B+ & B    & B+ & B+ & B+ & A  & B    & \cite{url_move} \\
    &HufuNet~\cite{new_lv2023robustness}               & B & ? & -  & ?/One & - & B+ & C & B+ & B+ & B & A  & B & \cite{url_hufunet}\\
    \\[-2ex]\cline{2-15}\\[-1.5ex]
    &Adi~\cite{black_adi2018turning}            & B+ & A & B & ?/One & B+ & B & C   & C & B- & B-  & A  & B & \cite{url_sok}\\
    &Content~\cite{black_zhang2018protecting}          & B+ & A  & C  & ?/One & B+ & B & B-   & C & B- & B & A  & B & \cite{url_sok}\\
    &Noise~\cite{black_zhang2018protecting}            & B+ & A  & C & ?/One & B+ & B+ & B-     & C & B- & B  & A  & B & \cite{url_sok}\\
    &Unralated~\cite{black_zhang2018protecting}        & B+ & A  & B- & ?/One & B+ & B & C     & C & B- & B-  & A  & B & \cite{url_sok}\\
    &Blackmarks~\cite{black_chen2019blackmarks}        & B & A  & C & ?/Multi & B+ & B & B-     & \multicolumn{3}{c|}{N/A} & A  & B  & \cite{url_sok}\\
    &DAWN~\cite{black_szyller2019dawn}                 & B- & A  & B- & ?/One & B+ & B- & B+     & C & C & B- & A  & B  & \cite{url_sok}\\
    &Namba~\cite{black_namba2019robust}        & B  & A & B- & ?/One & B+& B & C    & ? & B+ & B & A  & B  &  \cite{url_namba}\\
    &Certified~\cite{black_bansal2022certified}     & B  & A & B- & ?/One & B+ & B+ & B-     & \multicolumn{3}{c|}{N/A} & A  & B &  \cite{url_certified}\\
    &UBW-P~\cite{black_li2022untargetedbackdoorwatermark}& B+ & A  & A & ?/One & B+ & B & B-    & \multicolumn{3}{c|}{N/A} & A  & B  & \cite{url_ubw}\\
    &Blind-WM~\cite{black_li2019prove} & B+ & A & B- & ?/One & B+ & B+ & B-    & B- & B+ & B & A  & B  & \cite{url_zheng}\\
    &EWE~\cite{black_jia2021entangled}          & B & A  & B- & ?/One & B+ & B+ & B+     & B- & B- & B & A  & B  & \cite{url_jia}\\ 
    &Merrer~\cite{black_le2020adversarial} & A & A & A & ?/One  & B+ & B+ & C    & B- & B- & B & B & B & \cite{url_merrer}\\ 
    &NTL~\cite{black_wang2022ntl}                      & B+ & A & B- & ?/One & B+ & B+ & B+     & \multicolumn{3}{c|}{N/A} & A  & B  & \cite{url_ntl} \\
    &UBW-C~\cite{black_li2022untargetedbackdoorwatermark}& B & -  & A & ?/One &  \multicolumn{6}{c|}{N/A}     & A  & B  &\cite{url_ubw}\\
    &CosWM~\cite{black_charette2022cosine}  & B & A  & ? & ?/One & B+ & B+ & B+   & \multicolumn{3}{c|}{N/A} & A  & B  & \cite{url_coswm}\\
    \\[-2ex]\cline{2-15}\\[-1.5ex]
    &Passport~\cite{passport_fan2021deepip}            & B- & B+ & - & ?/Multi & -  & B+ & B- & B+ & B & B+  & A  & B  & \cite{url_passport}\\
    &Pass-Norm~\cite{passport_zhang2020passport}   & B & B+ & - & ?/Multi & -  & B+ & B-     & B+ & B & B+  & A  & B  & \cite{url_passportnorm}\\
    &IPR-GAN~\cite{passport_ong2021iprgan} & B- & B+  & - & ?/Multi & - & B+ & B-  & B+ & B & B+ & A  & B  & \cite{url_iprgan}\\
    &IPR-RNN~\cite{passport_lim2022rnn} & B & B+ & - & ?/Multi & - & B+ & B-  & B+ & B & B+ & A  & B  & \cite{url_iprrnn} \\
    &Lottery~\cite{lottery_chen2021you}    & B+ & B+ & B & ?/Multi & - & B & -  & \multicolumn{3}{c|}{N/A} & A  & B & \cite{url_lottery} \\
    \\[-2ex]\cline{2-15}\\[-1.5ex]
    &WAFFLE~\cite{new_tekgul2021waffle} & \multicolumn{3}{c|}{N/A} & ?/One & \multicolumn{3}{c}{N/A}   & \multicolumn{3}{c|}{N/A} &  \multicolumn{2}{c|}{N/A} &\cite{url_waffle} \\
    &FedIPR~\cite{fedwm_li2022fedipr} & \multicolumn{3}{c|}{N/A} & T/Multi & \multicolumn{3}{c}{N/A}     & \multicolumn{3}{c|}{N/A} & \multicolumn{2}{c|}{N/A}
    &\cite{url_fedipr}\\
    &SSLGuard~\cite{selfsuper_cong2022sslguard} &\multicolumn{3}{c|}{N/A} & ?/One & \multicolumn{3}{c}{N/A}  & \multicolumn{3}{c|}{N/A} &  A  & A  & \cite{url_sslguard}\\[1pt]
    &CoProtector~\cite{black_sun2022coprotector} & \multicolumn{3}{c|}{N/A} & ?/One & \multicolumn{3}{c}{N/A}  & \multicolumn{3}{c|}{N/A} &  \multicolumn{2}{c|}{N/A} & \cite{url_coprotector}\\
    \midrule
    \multirow{14}{*}{\rotatebox{90}{\bf Fingerprinting}}
    &PoL~\cite{fingerprinting_jia2021proof}            & \multicolumn{3}{c|}{${\rm A^T}$} & ?/One & \multicolumn{3}{c}{N/A}  & \multicolumn{3}{c|}{N/A} & A & -  & \cite{url_pol} \\
    &IP Guard~\cite{fingerprinting_cao2021ipguard}     & \multicolumn{3}{c|}{${\rm A^T}$} & ?/One & \multicolumn{3}{c}{N/A}     & \multicolumn{3}{c|}{N/A} & \multicolumn{2}{c|}{N/A} & \cite{url_deepjudge} \\
    &CEM~\cite{fingerprinting_lukas2019deep}           & \multicolumn{3}{c|}{${\rm A^T}$} & ?/One & \multicolumn{3}{c}{N/A}    & \multicolumn{3}{c|}{N/A} & \multicolumn{2}{c|}{N/A} & \cite{url_cem}\\
    &MeFA~\cite{fingerprinting_liu2022mefa}            & \multicolumn{3}{c|}{${\rm A^T}$} & ?/One & \multicolumn{3}{c}{N/A}    & \multicolumn{3}{c|}{N/A} & \multicolumn{2}{c|}{N/A}  & \cite{url_mefa} \\
    &MetaFinger~\cite{fingerprinting_yang2022metafinger} & \multicolumn{3}{c|}{${\rm A^T}$} & ?/One & \multicolumn{3}{c}{N/A}    & \multicolumn{3}{c|}{N/A} & \multicolumn{2}{c|}{N/A} & \cite{url_metafinger}\\
    &Modeldiff~\cite{fingerprinting_li2021modeldiff}   & \multicolumn{3}{c|}{${\rm A^T}$} & ?/One & \multicolumn{3}{c}{N/A}     & \multicolumn{3}{c|}{N/A} & \multicolumn{2}{c|}{N/A} & \cite{url_modeldiff}\\
    &SAC~\cite{fingperprinting_guan2022samplecorrelation} & \multicolumn{3}{c|}{${\rm A^T}$} & ?/One & \multicolumn{3}{c}{N/A}     & \multicolumn{3}{c|}{N/A} & \multicolumn{2}{c|}{N/A} & \cite{url_sac}\\
    &DeepJudge~\cite{fingerprinting_chen2021copy}      & \multicolumn{3}{c|}{${\rm A^T}$} & ?/One  & \multicolumn{3}{c}{N/A}     & \multicolumn{3}{c|}{N/A} & \multicolumn{2}{c|}{N/A} & \cite{url_deepjudge} \\
    &ZEST~\cite{fingerprinting_jia2022Zest}            & \multicolumn{3}{c|}{${\rm A^T}$} & ?/One & \multicolumn{3}{c}{N/A}   & \multicolumn{3}{c|}{N/A} & \multicolumn{2}{c|}{N/A} &  \cite{url_zest}\\
    &TeacherFP~\cite{fingerprinting_chen2022teacherFP} &\multicolumn{3}{c|}{${\rm A^T}$}  & ?/One & \multicolumn{3}{c}{N/A}    & \multicolumn{3}{c|}{N/A} & \multicolumn{2}{c|}{N/A} & \cite{url_teacherfp}\\
    &DI~\cite{fingerprinting_maini2021datasetinference}  & \multicolumn{3}{|c|}{${\rm A^T}$} & ?/One & \multicolumn{3}{c}{N/A}     & \multicolumn{3}{c|}{N/A} & \multicolumn{2}{c|}{N/A} & \cite{url_di}\\
    &SSL-DI~\cite{fingerprinting_dziedzic2022DI4selfsupervise}  & \multicolumn{3}{c|}{${\rm A^T}$} & ?/One & \multicolumn{3}{c}{N/A}     & \multicolumn{3}{c|}{N/A} & \multicolumn{2}{c|}{N/A} & \cite{url_ssldi}\\
  \bottomrule
\end{tabular}\\[2pt]
Acc: accuracy, Eff: ffficiency, Risk: extra risks; Pre: input preprocessing, Mod: model modification,  Ext: model extraction, Upp: theoretical upper mound; Pay: the bit number of payloads; Amb: IP ambiguity, Eva: IP detection \& evasion, Rmv: adaptive IP removal; FPR: false positive rate, Sca: the scalability to various models and tasks. 
\vspace{-0.15in}
\end{table}

According to Table ~\ref{tab:performance}, we highlight the following conclusions: 
\begin{itemize}[leftmargin=*]
	\setlength{\topsep}{0pt}
	\setlength{\itemsep}{0pt}
	\setlength{\parsep}{0pt}
	\setlength{\parskip}{0pt} 
 \item The existing benchmark tests have proved that there is no Deep IP solution that defends against various attack strategies like IP evasion, remove, and ambiguity attacks. It still have a long way towards the guaranteed robustness of Deep IP solutions. 
 \item Until now, most of Deep IP solutions have not been convincingly evaluated to prove their security performance. Especially, there is no testing protocol for the performance of fingerprinting schemes. We hereby advocate more testing protocol and benchmark toolkits to evaluate the performance of Deep IP solutions.
 \item Only FedIPR~\cite{fedwm_li2022fedipr} has tried to provide a theoretical analysis about watermark capacities and detection rate. It still remains open to exploring capacity bounds of IP identifiers. 
 \item Fidelity and robustness are generally traded off against each other. Fingerprinting methods have optimal fidelity since it requires no model modification, but the robustness is not yet guaranteed. Watermarking methods with higher robustness may have lower fidelity. No empirical or theoretical studies have yet explored the quantitative relationship between fidelity and robustness. 
 \item The task-level scalability of inner-component-based solutions is generally higher than that of input/output-based solutions, because the former usually does not depend on the specific model outputs. However, the former has few scalability across model structures; the internal model components are quite different after cross-structure model extraction. 
\end{itemize}
\section{Conclusion and Outlook} \label{sec:future}
\textcolor{black}{As a comprehensive survey of contemporary DNN IP protection, this article started by discussing the pressing need, the problem, requirements, challenges, and performance evaluation criteria  of DNN IP protection. Later on, we first summarized the general frameworks and various threats of two commonly used DNN IP protection techniques: watermarking and fingerprinting, and then presented a novel  taxonomy to group the representative methods clearly. Merits and demerits of the identified categories and their
underlying connections are also analyzed. Finally,  a performance comparison of representative methods against potential threats was provided. Although promising progress has been made, robust and practical deep IP protection in the wild is still in its infancy, and many open challenges remain. Therefore, we identify a number of promising future directions for DNN IP protection.}

\textcolor{black}{\textbf{(1) Theoretical Study for DNN IP Protection}. Different from traditional watermarking of static multimedia,
DNN watermarking has many new opportunities due to the following characteristics of DNNs: 
(\emph{a}) a large, complex parameterized function consisting of multiple processing layers, (\emph{b}) learning with a large-scale dataset, (\emph{c}) learning representations of data with multiple levels of abstraction, (\emph{d}) learning to generate output by minimizing a loss function for the task it is thought to accomplish, (\emph{e}) unique behaviors can be induced during learning, and (\emph{f}) the issue of catastrophic forgetting. As reviewed in this work, there are various hosts to embed the watermark, including (i) some intermediate weight layers, (ii) the input data, (iii) the unique behavior of the DNN, (iv) some intermediate data representations, and (v) the output.  It should be noted that a DNN model needs to accomplish some difficult tasks, whereas the embedding of the IP identification information naturally conflicts with the high fidelity of the model.  Therefore, to advance this field, we believe that the development of a rigorous theory for DNN IP protection (watermarking or fingerprinting) should be strongly beneficial. However, current research in this field is rather scattered, and very few works focus on the theoretical study. Certainly,  the development of a rigorous theory for  DNN IP protection is not easy, and the blackbox issue of DNNs further increases the difficulty. Towards such a goal, several  questions deserve future attention. (i) How much IP identification information such as triggering inputs can we inject without impacting the capability of the network to solve the original problem? (ii)  How to design high-capacity static watermarks with certain robustness against finetuning, network pruning, etc.? (iii) How to develop certifiable watermarking robustness against various attacks with provable guarantees~\cite{black_bansal2022certified}
like certified adversarial robustness~\cite{adversarial_cohen2019certified}? 
 (iv) How to build the connection between the theory of adversarial robustness and the theory of DNN watermarking and fingerprinting?}

\textcolor{black}{ \textbf{(2) Better Protection Frameworks.} At present, the field of Deep IP protection is at the early stage of development and far from mature. None of the solutions has been tested for a long time in a real application environment. Therefore, developing novel DNN IP protection frameworks that can simultaneously ensure fidelity, robustness, and capacity is of great interest.} Feasible research directions include but are not limited to (i) adaptation improvement of advanced techniques in adversarial attacks and backdoor attacks; (ii) introducing the interpretable methods for DNNs like attribution and hidden-semantic analysis~\cite{survey_zhang2021interpretability}; (iii) exploring the mathematics of the generalization ability of DNNs in depth like optimal transport or information geometry. 
Moreover, more vulnerabilities in the construction and verification process must be identified to enable defenders to propose targeted defense strategies that mitigate potential threats. The interplay between defense and attack methods will foster mutual development, thereby promoting the maturity of Deep IP protection.   

\textbf{(3) {Efficient Protection Frameworks}}.
The efficiency of IP construction and verification is a crucial requirement for Deep IP Protection, as constructing IP identifiers often involves computationally complex gradient optimization on models or samples at a scale. Feasible research directions include but are not limited to: (i) Exploring methods to make IP identifiers sensitive to construction while insensitive to removal, for example, to guarantee the quality of IP identifiers without requiring model retraining, extensive model finetuning or gradient optimization. (ii) Exploring methods to reduce the verification overhead, for example, utilizing surrogate models for optimizing high-quality query samples and reducing the number of queries submitted for remote verification. (iii) Enhancing the cross-model and cross-task scalability of IP identifiers, such as a single IP construction to meet the IP protection needs of numerous downstream tasks.


\par\textbf{\textcolor{black}{(4) IP Protection for Large Foundation Models.}} AI applications based on large foundation models have covered various aspects of human daily lives, like question answering (ChatGPT\footnote{\url{https://openai.com/product}}) and voice assistants. Since building a high-performance large model costs a lot on infrastructure construction, data collection and labelling, and model training, it is an urgent need to develop a complete large model IP tool  than can provide strong attack resistance, acceptable overhead and imperceptible performance degradation. The following challenges warrant consideration: (i) Large models exhibit greater complexity and uncertainty, with behavioral boundaries that are difficult to explore and interpret.  Embedding watermarks will significantly degrade the performance and introduce agnostic security risks, while extracting fingerprints from large models is more challenging and offers few guarantees in the face of various attacks or task adaptations;
(ii) Large foundation models (GPT-4\cite{large_openai2023gpt4} and DALL-E 2~\cite{large_ramesh2022hierarchical}) typically support cross-modal data including text and images, and incorporate new model structures or learning techniques unexplored in Deep IPs like Denoising Diffusion Probabilistic Models and Transformer; iii) Once a high-performance large model is released, it will scale to various downstream tasks. Criminals will also use these powerful tools for malicious purposes. Therefore, in addition to ensuring attack resistance, Deep IP tools should have functions to prevent or trace model abuse.  

\textcolor{black}{\textbf{(5) A Unified Evaluation Benchmark.} In the past few years, many approaches have been proposed for DNN IP protection, however, none of them adopts the same criteria for evaluating competing requirements like fidelity, robustness, capacity, etc. This is certainly not beneficial for a fair comparison of various methods and their deployment
in practical applications, yet it slows down progress in this field.  Therefore, a unified, systematic, empirical evaluation benchmark is essential for advancing DNN IP protection. This issue has been realized by some researchers~\cite{sok_lukas2021sok} who evaluated the robustness against
a number of removal attacks of some recent DNN IP protection methods. To this end, a wide spectrum of evaluation criteria should be carefully established, such as a unified and large dataset, a set of quantitative metrics, a set of fundamental functional metrics including fidelity, robustness, capacity, efficiency, etc., a set of common attacks, and various application scenarios.}

\textbf{(6) \textcolor{black}{Exploring More Functions.}} As this survey has shown, most existing DNN watermarking and fingerprinting schemes focus on copyright verification. Several works focus on other functions like integrity verification, user management or applicability authorization. However, more functions are worth considering. For example, semi-fragile watermarks or fingerprints can be researched to allow legitimate users to modify the model within limits to support their own personalized applications. Here, more functions can be explored from the ideas of traditional data watermarking or the properties of DNNs like high-dimensional nonlinear  decision boundary and generalization domain. 


\textcolor{black}{\textbf{(7) Beyond Image Classification.} Most existing DNN IP protection approaches focus on discriminative DNN models for the fundamental classification task. Very few works focus on more complex tasks like object detection, image segmentation, image captioning, and generative models like GANs and diffusion models  for the image generation task, as well as other learning paradigms like reinforcement learning, selfsupervised learning, transfer learning, federated learning, and multimodal learning.} 

\textbf{(8) \textcolor{black}{Secure Pipelines}}. 
 The pipeline of Deep IP protection from design to deployment to forensics, includes not only IP construction and verification, but also more components, like discovering infringements, credible notarization mechanism, and secure IP management platforms: (a) \textit{Active defense techniques}: Deep IP Protectors can prevent model stealing by identifying the specific query patterns of thieves, from the patterns of query samples and network states. (b) \textit{Low-cost automated discovery for potential infringement facts}. Protectors need to accurately locate MLaaS providers using pirated models. For black-box verification, it is crucial to balance verification cost with verification accuracy, as there are many MLaaS APIs online and most providers operate on a pay-per-request business model. In no-box scenarios, it requires a mechanism to perceive and acquire suspect data and its source model. 
 (c) \textit{Credible notarization mechanism and platforms.} If IP identifiers are verified by model owners, the copyright conflict problem is hard to tackle as any party can create his own IP identifier of the protected model, confusing the verification results. Thus, a credible verification platform is necessary to build a database of IP identifiers. This platform can be decentralized or owned by a trusted third party like a judicial authority. Moreover, the construction order detection mechanism for multiple IP identifiers is required to trace the model distribution chain, using catastrophic forgetting of DNNs. 

\textbf{(9) \textcolor{black}{Beyond Techniques: Standardization, Law and Policy}}. In addition to technical advancements, it is also crucial to establish legal frameworks for Deep IP protection. Firstly, the industry should develop a standardized workflow for Deep IP protection, and only models that comply with the standards will be protected. The standards should be secure enough and as potential vulnerabilities arise. 
The standards require a deep IP scheme not relying on algorithm concealment. However, the existing schemes will not work if the algorithm details are leaked, which has been proven unreliable in the long-term practice in the field of information security. The Kerckhoffs principle should be followed: : even if the adversary is aware of the algorithm details, it should remain arduous for them to remove, modify, or forge the IP identifiers while ensuring model performance. Moreover, judicial institutions and government functional departments should conduct research and implement  effective laws and policies to deter the theft, piracy, abuse and other malicious behaviors, and promote the healthy development of deep learning technology and its related industries.
\nop{
\par\textbf{(10) Deep IP Schemes not relying on Algorithmic Concealment}. Most existing Deep IP schemes are based on algorithmic concealment to realize the discriminability of IP signatures, which assumes that the adversary does not know the specific details of the algorithm. However, schemes based on this assumption have been proven to be unreliable in the long-term practice in the field of information security. The Kerckhoffs principle should be followed to develop a Deep IP solution that relies only on key security:  Even if the Deep IP algorithm is known by the adversary, it is difficult for the adversary to remove or modify the constructed IP signatures of the target model on the premise of ensuring the model performance so as to avoid IP verification or falsely ownership claiming.} 


\footnotesize
\bibliographystyle{IEEEtran}
\bibliography{deepip} 
\vspace{-0.7in}
\begin{IEEEbiography}[{\includegraphics[width=1in,height=1.25in,clip,keepaspectratio]{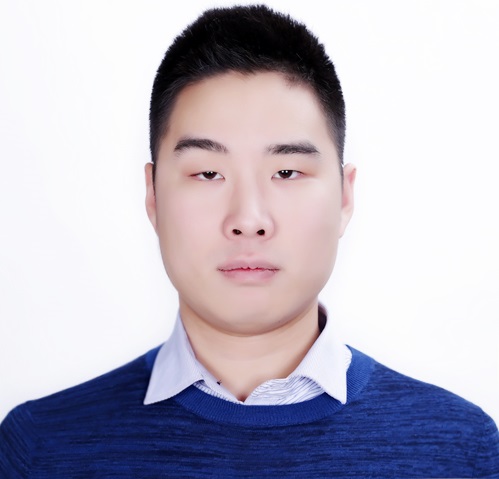}}]{Yuchen Sun} received the B.S. degree in Telecommunication Engineering from the Huazhong University of Science and Technology (HUST), Wuhan, China, in 2018. He has been with the School of System Engineering, National University of Defense Technology (NUDT), Changsha, since 2018, where he is currently a Ph.D. candidate. His research interests include Trustworthy Artificial Intelligence, Distributed Systems, and Wireless Indoor Localization. 
\end{IEEEbiography}
\vspace{-0.7in}
\begin{IEEEbiography}[{\includegraphics[width=1in,height=1.25in,clip,keepaspectratio]{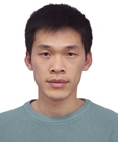}}]{Tianpeng Liu}
Tianpeng Liu received the B.Eng. and M.Eng and Ph.D. degrees from the National University of Defense Technology (NUDT), Changsha, China, in 2008, 2011, and 2016 respectively. He is currently an associate professor at the College of Electronic Science and Technology, NUDT. His primary research interests are radar signal processing, automatic target recognition, and cross-eye jamming.
\end{IEEEbiography}
\vspace{-0.7in}
\begin{IEEEbiography}[{\includegraphics[width=1in,height=1.25in,clip,keepaspectratio]{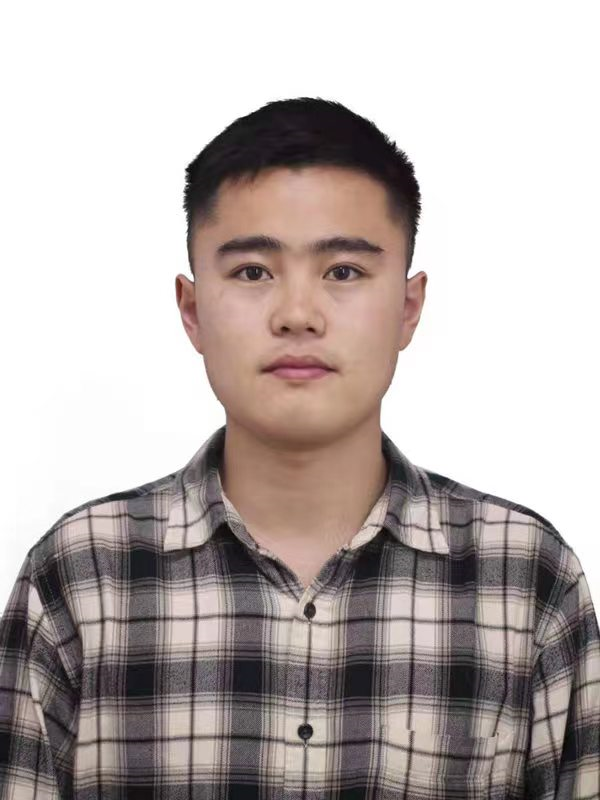}}]{Panhe Hu}
Panhe Hu was born in Shandong, China, in 1991. He received his BS degree from the Xidian University, Xi’an, China, in 2013, and his PhD degree from the National University of Defense Technology (NUDT), Changsha, China, in 2019. He is currently an associate professor with the College of Electronic Science and Technology, NUDT. His current research interests include radar signal design, array signal processing and deep learning.
\end{IEEEbiography}
\begin{IEEEbiography}[{\includegraphics[width=1in,height=1.25in,clip,keepaspectratio]{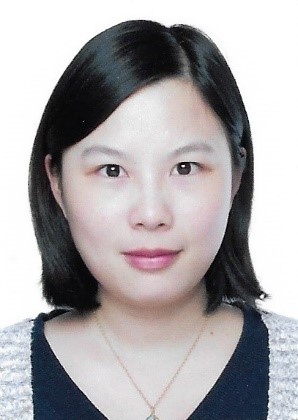}}]{Qing Liao}
received her Ph.D. degree in computer science and engineering in 2016 supervised by Prof. Qian Zhang from the Department of Computer Science and Engineering of the Hong Kong University of Science and Technology. She is currently a professor at School of Computer Science and Technology, Harbin Institute of Technology (Shenzhen), China. Her research interests include artificial intelligence, data mining and information security.
\end{IEEEbiography}
\vspace{-0.6in}
\begin{IEEEbiography}[{\includegraphics[width=1in,height=1.25in,clip,keepaspectratio]{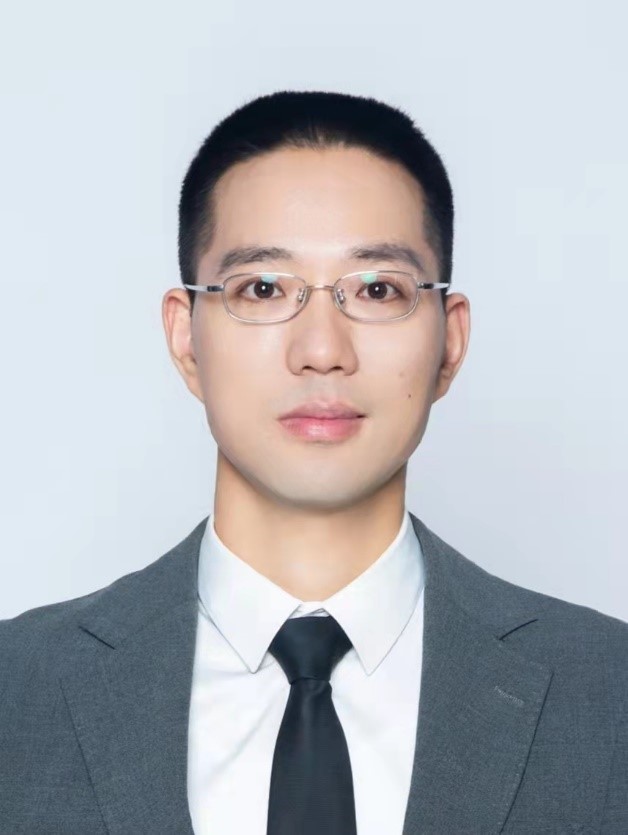}}]{Shaojing Fu}
Shaojing Fu received the Ph.D. degree in applied mathematics from the National University of Defense Technology in 2010. He spent a year as a Joint Doctoral Student at the University of Tokyo for one year. He is currently a Professor at the College of Computer, National University of Defense Technology. His research interests include cryptography theory and application, security in the cloud, and mobile computing. 
\end{IEEEbiography}
\vspace{-0.6in}
\begin{IEEEbiography}[{\includegraphics[width=1in,height=1.25in,clip,keepaspectratio]{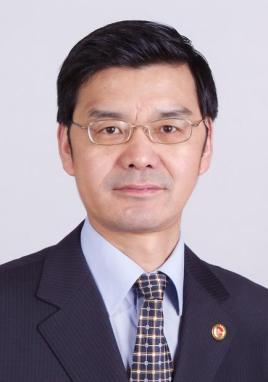}}]{Nenghai Yu}
Nenghai Yu received the B.S. degree from the Nanjing University of Posts and Telecommunications, Nanjing, China, in 1987, the M.E. degree from Tsinghua University, Beijing, China, in 1992, and the Ph.D. degree from the University of Science and Technology of China (USTC), Hefei, China, in 2004. Since 1992, he has been a Faculty Member with the Department of Electronic Engineering and Information Science, USTC, where he is currently a Professor. He is the Executive Director of the Department of Electronic Engineering and Information Science, USTC, where he is the Director of the Information Processing Center. He has authored or coauthored over 130 papers in journals and international conferences. His current research interests include multimedia security, multimedia information retrieval, video processing, and information hiding.
\end{IEEEbiography}
\vspace{-0.6in}
\begin{IEEEbiography}[{\includegraphics[width=1in,height=1.25in,clip,keepaspectratio]{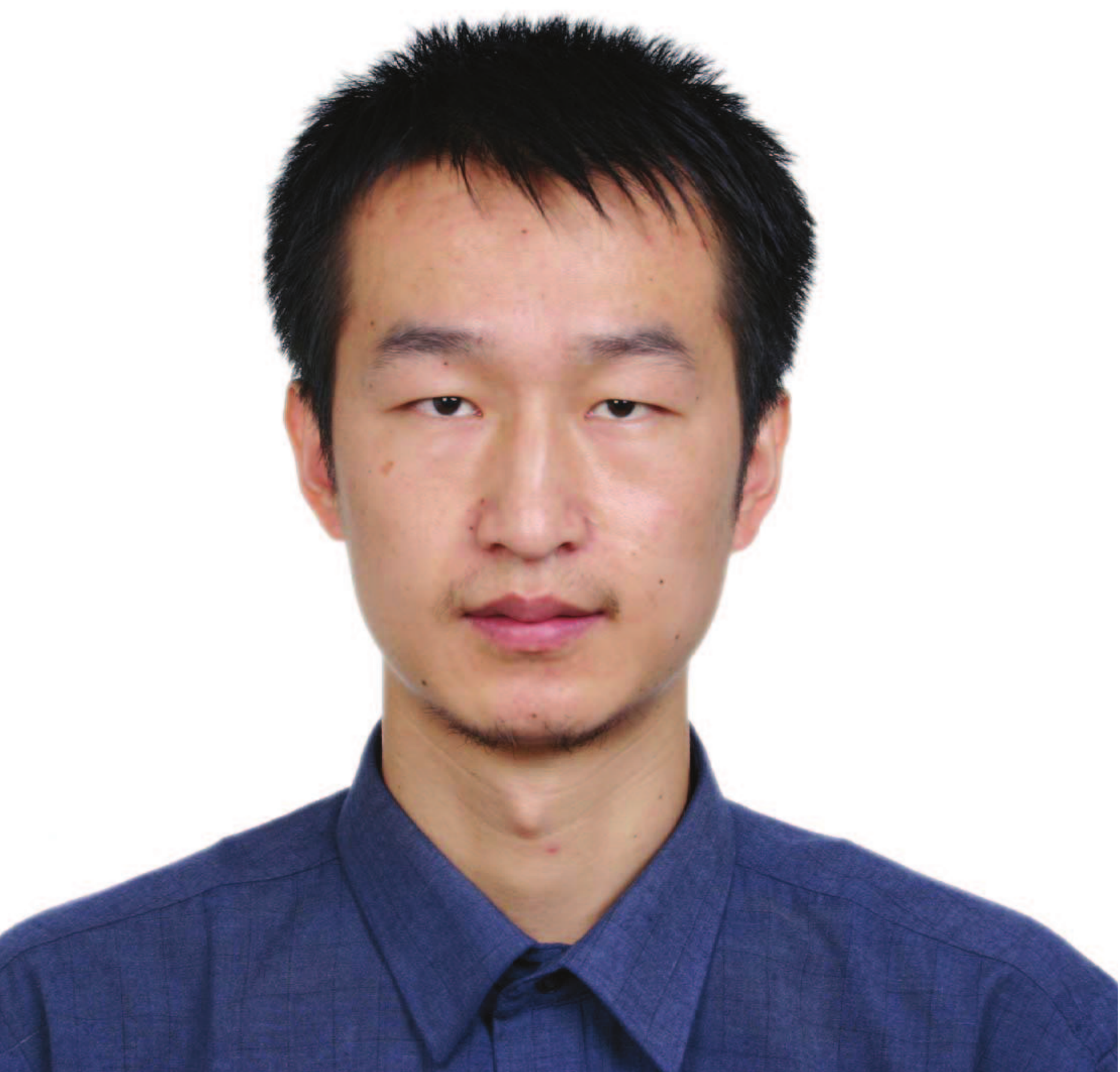}}]{Deke Guo}
received the B.S. degree in industry engineering from the Beijing University of Aeronautics and Astronautics, Beijing, China, in 2001, and the Ph.D. degree in management science and engineering from the National University of Defense Technology, Changsha, China, in 2008. He is currently a Professor with the College of System Engineering, National University of Defense Technology, and is also with the College of Intelligence and Computing, Tianjin University. His research interests include distributed systems, software-defined networking, data center networking, wireless and mobile systems, and interconnection networks. He is a senior member of the IEEE and a member of the ACM.
\end{IEEEbiography}
\vspace{-0.6in}
\begin{IEEEbiography}[{\includegraphics[width=1in,height=1.25in,clip,keepaspectratio]{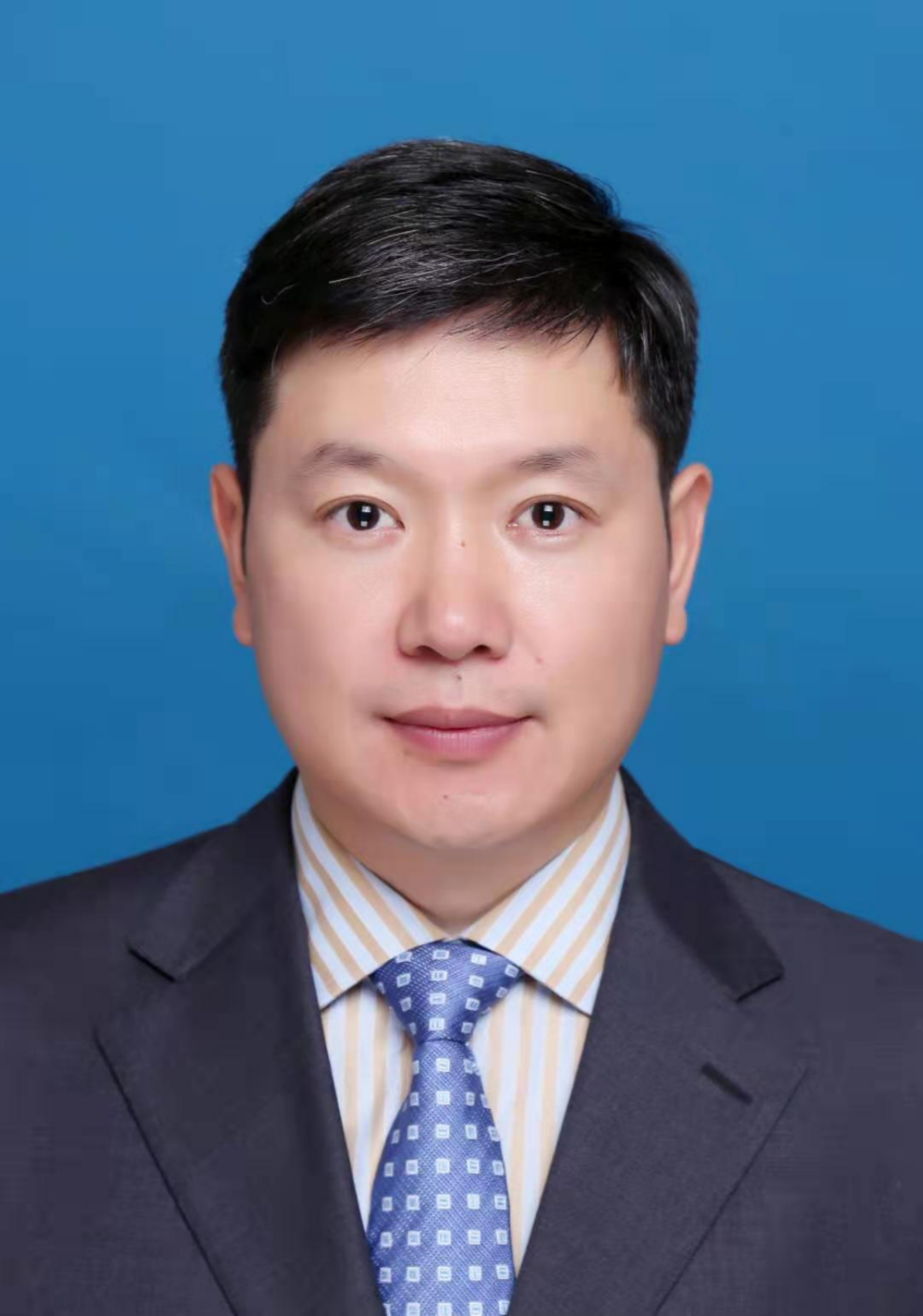}}]{Yongxiang Liu} received the B.S. and Ph.D. degrees from the National University of Defense Technology (NUDT) in 1999 and 2004,
respectively. In 2008, he worked at Imperial College London, as an Academic Visitor. Since 2004, he has been
with NUDT, where he is currently a Professor at the College of Electrical Science and Engineering, conducting
research on radar target recognition, time-frequency analysis, and micromotions.
\end{IEEEbiography}
\vspace{-0.6in}
\begin{IEEEbiography}[{\includegraphics[width=1in,height=1.25in,clip,keepaspectratio]{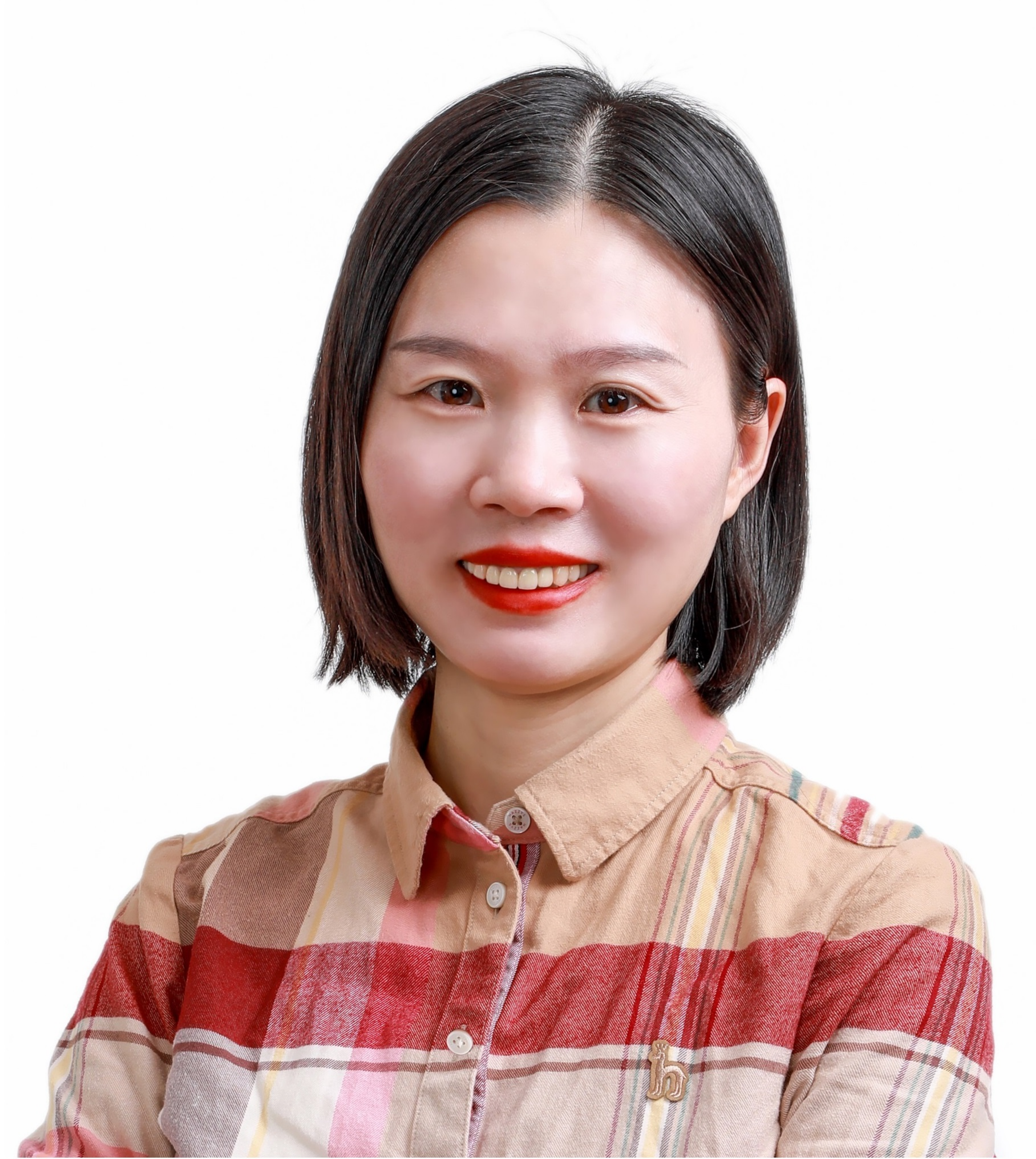}}]{Li Liu} received the Ph.D. degree in information and communication engineering from the National University of Defense Technology (NUDT), China, in 2012. She is currently a Full Professor with NUDT. During her Ph.D. study, she spent more than two years as a Visiting Student at the University of Waterloo, Canada, from 2008 to 2010. From 2015 to 2016, she spent ten months visiting the Multimedia Laboratory at the Chinese University of Hong Kong. From 2016.12 to 2018.11, she worked as a senior researcher at the Machine Vision Group at the University of Oulu, Finland. Her current research interests include Computer Vision, Machine Learning, Artificial Intelligence, Trustworthy AI, Synthetic Aperture Radar. Her papers currently have over 10200 citations in Google Scholar. 
\end{IEEEbiography}

\end{document}